\documentclass{article} 
\usepackage{configs/iclr2022_conference}
\usepackage{times}

\usepackage{amsmath,amsfonts,bm}

\def\eqref#1{equation~\ref{#1}}

\def\1{\bm{1}}

\DeclareMathAlphabet{\mathsfit}{\encodingdefault}{\sfdefault}{m}{sl}
\SetMathAlphabet{\mathsfit}{bold}{\encodingdefault}{\sfdefault}{bx}{n}

\usepackage{wrapfig}

\usepackage[utf8]{inputenc} 
\usepackage[T1]{fontenc}    

\usepackage{hyperref}       
\usepackage{url}            
\usepackage{booktabs}       

\usepackage{amsfonts}       
\usepackage{amsmath}
\usepackage{bm}

\usepackage{nicefrac}       

\usepackage{microtype}      
\usepackage{xcolor}
\usepackage{soul}

\usepackage{graphicx}
\usepackage{wrapfig}
\usepackage[labelformat=simple]{subcaption}

\usepackage{multirow}
\usepackage{multicol}
\usepackage{tabularx}
\usepackage{makecell}

\newcolumntype{s}{>{\hsize=.15\hsize}X}

\usepackage{algorithm}
\usepackage{algpseudocode}

\algnewcommand\algorithmicto{\textbf{to}}
\algrenewtext{For}[3]{\algorithmicfor\ #1 $\gets$ #2 \algorithmicto\ #3 \algorithmicdo}

\algnewcommand\algorithmicforp{\textbf{for}}
\algnewcommand\algorithmicdop{\textbf{do in parallel}}
\algdef{S}[FOR]{ForP}[3]{\algorithmicforp\ #1 $\gets$ #2 \algorithmicto\ #3 \algorithmicdop}

\usepackage{tikz}
\usetikzlibrary{fit,calc}
\newcommand*{\tikzmk}[1]{\tikz[remember picture,overlay,] \node (#1) {};\ignorespaces}
\newcommand{\boxitf}[1]{\tikz[remember picture,overlay]{\node[yshift=1pt,fill=#1,opacity=.25,fit={($(A)+(-0.17\linewidth,-8pt)$)($(B)+(.88\linewidth,2\baselineskip)$)}] {};}\ignorespaces}

\newcommand{\boxits}[1]{\tikz[remember picture,overlay]{\node[yshift=1pt,fill=#1,opacity=.15,fit={($(A)+(-0.96\linewidth,-8pt)$)($(B)+(.88\linewidth,.05\baselineskip)$)}] {};}\ignorespaces}

\definecolor{codell}{HTML}{76b5c5} 
\definecolor{codecg}{HTML}{e27743} 

\newtheorem{theorem}{Theorem}

\newtheorem{lemma}{Lemma}

\newtheorem{assumption}{Assumption}

\graphicspath{ {../figures/} }

\definecolor{links}{HTML}{0078b0} 
\definecolor{files}{HTML}{fc6160}
\hypersetup{
    colorlinks=true,
    linkcolor=files,
    filecolor=files,      
    urlcolor=links,
    citecolor=links,
}

\newcommand{\CG}[0]{\textsc{LLCG}~}
\newcommand*{\CGE}{\textsc{LLCG}}

\newcommand{\makeappendixtitle}{%
    \vbox{\hsize\textwidth
        {\LARGE\sc \textbf{Supplementary Material}\\
        Learn Locally, Correct Globally: A Distributed Algorithm for Training Graph Neural Networks \par}
        \vskip 0.3in minus 0.1in
    }
}

\usepackage{hyperref}
\usepackage{url}

\usepackage[toc,page,header]{appendix}
\usepackage{minitoc}

\title{Learn Locally, Correct Globally:\\ A Distributed Algorithm for Training Graph Neural Networks}

\author{Morteza Ramezani\thanks{Equal Contribution.} ,~ Weilin Cong\footnotemark[1] ,~ Mehrdad Mahdavi \\
\textbf{Mahmut T. Kandemir, ~Anand Sivasubramaniam} \\
Pennsylvania State University, University Park, PA 16802, USA \\
\texttt{morteza@cse.psu.edu, wxc272@psu.edu, mzm616@psu.edu}\\
\texttt{anand@cse.psu.edu, kandemir@cse.psu.edu} \\
}

\iclrfinalcopy 
\begin{document}

\renewcommand \thepart{}
\renewcommand \partname{}
\doparttoc 
\faketableofcontents 

\maketitle

\begin{abstract}
Despite the recent success of Graph Neural Networks (GNNs), training GNNs on large graphs remains challenging. The limited resource capacities of the existing servers, the dependency between nodes in a graph, and the privacy concern due to the centralized storage and model learning have spurred the need to design an effective distributed algorithm for GNN training. However, existing distributed GNN training methods impose either excessive communication costs or large memory overheads that hinders their scalability. To overcome these issues, we propose a communication-efficient distributed GNN training technique named $\text{\textit{Learn Locally, Correct Globally}}$ (LLCG). To reduce the communication and memory overhead, each local machine in LLCG first trains a GNN on its local data by ignoring the dependency between nodes among different machines, then sends the locally trained model to the server for periodic model averaging. However, ignoring node dependency could result in significant performance degradation. To solve the performance degradation, we propose to apply $\text{\textit{Global Server Corrections}}$ on the server to refine the locally learned models. We rigorously analyze the convergence of distributed methods  with periodic model averaging for training GNNs and show that naively applying periodic model averaging but ignoring the dependency between nodes will suffer from an irreducible residual error. However, this residual error can be eliminated  by utilizing the proposed global corrections to entail fast convergence rate. Extensive experiments on real-world datasets show that LLCG can significantly improve the efficiency without hurting the performance.
\end{abstract}


\section{Introduction}
\label{section:intro}

In recent years, Graph Neural Networks (GNNs) have achieved impressive results across numerous graph-based applications, including social networks~\citep{hamilton2017inductive,deng2019learning}, 
\begin{wrapfigure}[13]{r}{0.3\textwidth}
    \centering
    \vspace{-8pt}
    \includegraphics[width=1\linewidth]{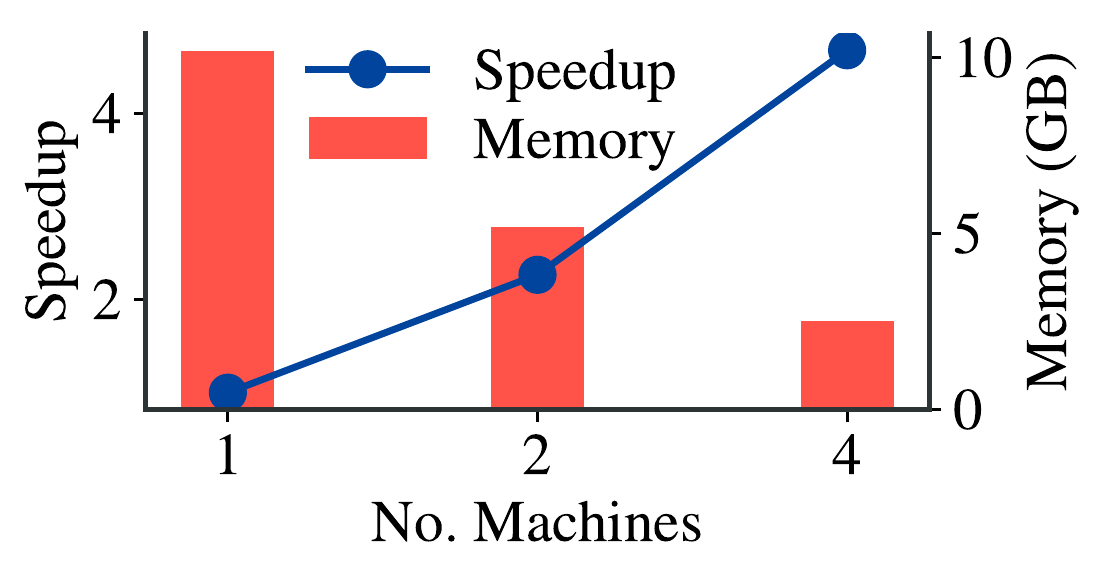}
    \vspace{-15pt}
    \caption{
    Comparison of the speedup and the memory consumption of distributed multi-machine training and centralized single machine training on the \texttt{Reddit} dataset.
    }
    \label{figure:problems2}
\end{wrapfigure}
recommendation systems~\citep{ying2018graph,wang2018billion}, and drug discovery~\citep{fout2017protein,do2019graph, ghorbani2022ra, faez2021deep}.
Despite their recent success, effective training of GNNs on large-scale real-world graphs, such as Facebook social network~\citep{boldi2004webgraph}, remains challenging.
Although several attempts have been made to scale GNN training by sampling techniques~\citep{hamilton2017inductive, zou2019layer, zeng2019graphsaint, chiang2019cluster, chen2017stochastic,zhang2021biased, ramezani2020gcn}, they are still inefficient for training on extremely large graphs, due to the unique structure of GNNs and the limited memory capacity/bandwidth of current servers.
One potential solution to tackle these limitations is employing distributed training with data parallelism, which have become almost a \textit{de facto} standard for fast and accurate training for natural language processing~\citep{lin2021fednlp,hard2018federated} and computer vision~\citep{bonawitztowards,konevcny2018federated}.
For example, as shown in Figure~\ref{figure:problems2}, moving from single machine to multiple machines reduces the training time and alleviates the memory burden on each machine.
Besides, scaling the training of GNNs with sampling techniques can result in privacy concerns:
existing sampling-based methods require centralized data storage and model learning, which could result in privacy concerns in real-world scenarios~\citep{shin2018privacy,wu2021fedgnn}.
Fortunately, the privacy in distributed learning can be preserved by avoiding mutual access to data between different local machines, and using only a trusted third party server to access the entire data.

Nonetheless, generalizing the existing data parallelism techniques of classical distributed training to the graph domain is non-trivial, which is mainly due to the dependency between nodes in a graph. 
For example, unlike solving image classification problems where images are mutually independent, such that we can divide the image dataset into several partitions without worrying about the dependency between images; 
GNNs are heavily relying on the information inherent to a node and its neighboring nodes. 
As a result, partitioning the graph leads to subgraphs with edges spanning subgraphs ({\em cut-edges}), which will cause information loss and hinder the performance of the model~\citep{angerd2020distributed}.
To cope with this problem, \citep{ md2021distgnn, jiang2021communication, angerd2020distributed} propose to transfer node features and~\citep{zheng2020distdgl, tripathy2020reducing, scardapane2020distributed} propose to transfer both the node feature and its hidden embeddings between local machines, both of which can cause significant storage/communication overhead and privacy concerns~\citep{shin2018privacy,wu2021fedgnn}.
\begin{wrapfigure}[18]{r}{0.49\textwidth}
    \vspace{-5pt}
    \centering
    \begin{subfigure}[t]{0.3\textwidth}
        \centering
        \includegraphics[trim=0 0 0 0, width=1\linewidth]{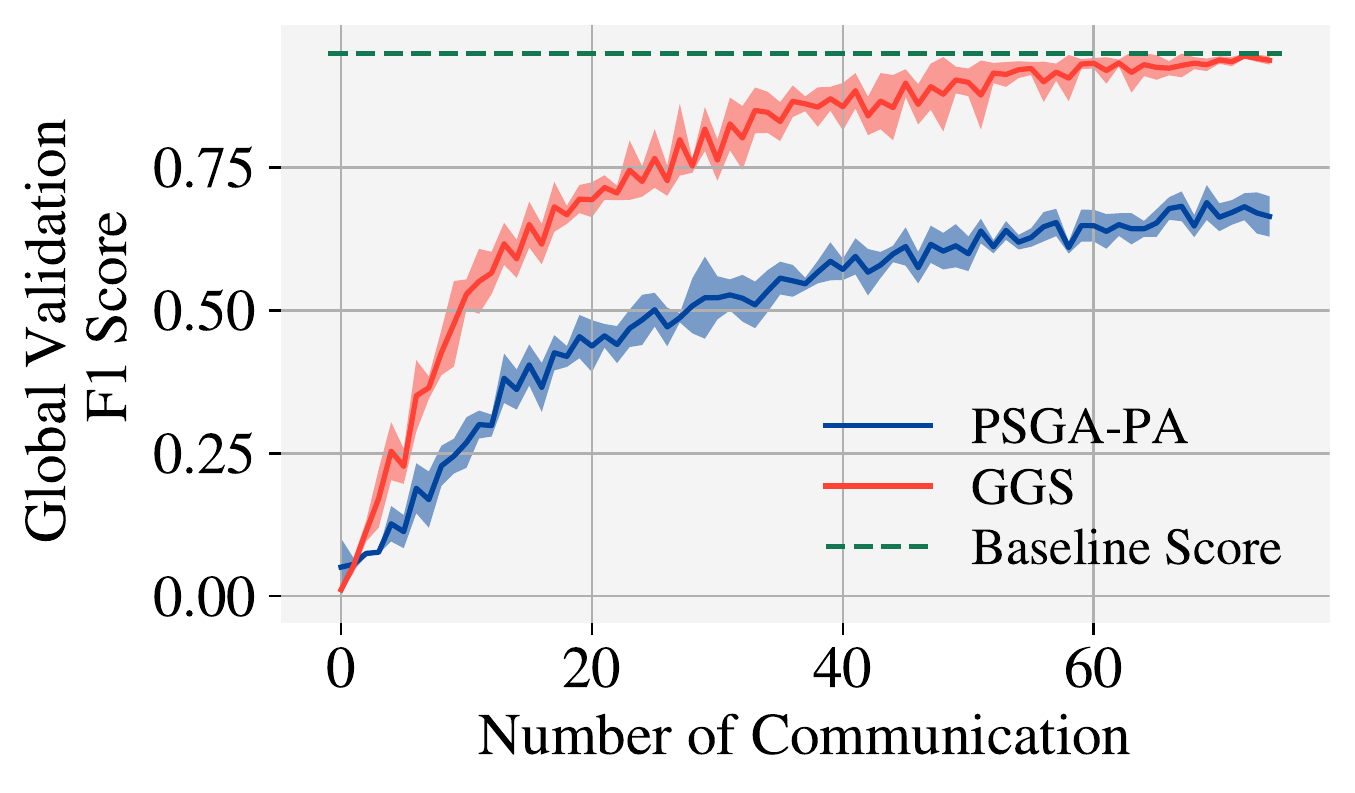}
        \vspace{-18pt}
        \caption{}
        \label{figure:intro-acc}
    \end{subfigure}
    \begin{subfigure}[t]{0.17\textwidth}
        \centering
        \includegraphics[trim=0 0 0 0, width=1\linewidth]{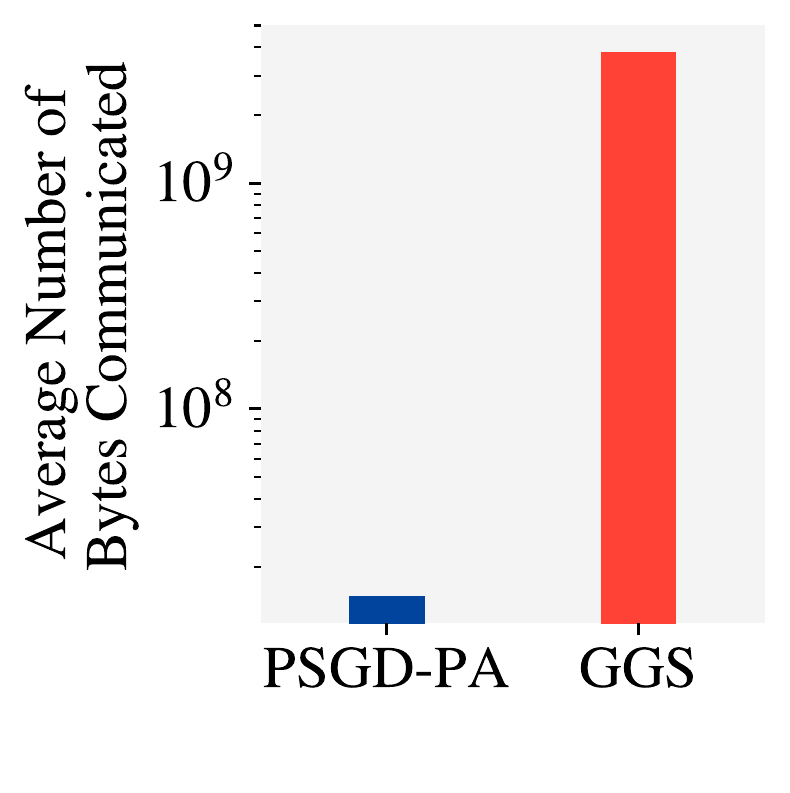}
        \vspace{-18pt}
        \caption{}
        \label{figure:intro-comm}
    \end{subfigure}
    \vspace{-12pt}
    \caption{
    Comparison of (a) the validation F1-score and (b) the average data communicated per round (in bytes and log-scale) for two different distributed GNN training settings, including \textit{Parallel SGD with Periodic Averaging} (PSGD-PA) where the cut-edges are ignored and only the model parameters are transferred and \textit{Global Graph Sampling} (GGS), where the cut-edges are considered and the node features of the cut-edges are transferred to the corresponding local machine, on the \texttt{Reddit} dataset using 8 machines.
    }
    \label{figure:problems}
\end{wrapfigure}

To better understand the challenge of distributed GNN training, we compare the validation F1-score in Figure~\ref{figure:intro-acc} and the average data communicated per round in Figure~\ref{figure:intro-comm} for two different distributed GNN training methods on the \texttt{Reddit} dataset. 
On the one hand, we can observe that when ignoring the cut-edges, \textit{Parallel SGD with Periodic Averaging} (PSGD-PA~\citep{dean2012large,li2019convergence}) suffers from significant accuracy drop and cannot achieve the same accuracy as the single machine training, even by increasing the number of communication.
However, \textit{Global Graph Sampling} (GGS) can successfully reach the baseline by considering the cut-edges and allowing feature transfer, at the cost of significant communication overhead, and potential violation of privacy.

In this paper, we propose a communication-efficient distributed GNN training method, called \textit{Learn Locally, Correct Globally} (\CGE{}).
To reduce the communication overhead, inspired by the recent success of the distributed optimization with periodic averaging~\citep{stich2018local,yu2019parallel}, we propose \textit{Local Training with Periodic Averaging}: where each local machine first locally trains a GNN model by ignoring the cut-edges, then sends the trained model to the server for periodic model averaging, and receive the averaged model from server to continue the training.
By doing so we eliminate the features exchange phase between server and local machines, but it can result in a significant performance degradation due to the lack of the global graph structure and the dependency between nodes among different machines.
To compensate for this error, we propose a {\em Global Server Correction} scheme to take advantage of the available global graph structure on the server and refine the averaged locally learned models before sending it back to each local machine. 
Notice that without {\em Global Server Correction}, \CG is similar to PSGD-PA as introduced in Figure~\ref{figure:problems}.

To get a deeper understanding on the necessity of {\em Global Server Correction}, we provide the first theoretical analysis on the convergence of distributed training for GNNs with periodic averaging. 
In particular, we show that solely averaging the local machine models and ignoring the global graph structure will suffer from an irreducible residual error, which provides sufficient explanation on why \textit{Parallel SGD with Periodic Averaging} can never achieve the same performance as the model trained on a single machine in Figure~\ref{figure:intro-acc}.
Then, we theoretically analyze the convergence of our proposal \CG{}.
We show that by carefully choosing the number of global correction steps, \CG{} can overcome the aforementioned residual error and enjoys $\smash{\mathcal{O}\big({1}/{\sqrt{PT}}\big)}$ convergence rate with $P$ local machines and $T$ iterations of gradient updates, which matches the rate of~\citep{yu2019parallel} on a general (not specific for GNN training) non-convex optimization setting.
Finally, we conduct comprehensive evaluations on real-world graph datasets with ablation study to validate the effectiveness of \CG{} and its improvements over the existing distributed methods.

\noindent\textbf{Related works.~}
\label{section:related}
Recently, several attempts have been made on distributed GNN training.
According to how they deal with the input/hidden feature of nodes that are associated with the cut-edges (i.e., the edges spanning subgraphs of each local machine), existing methods can be classified into two main categories:
($1$) \textit{Input feature only communication-based methods}: In these methods, each local machine receives the input features of all nodes required for the gradient computation from other machines, and trains individually.
However, since the number of required nodes grows exponentially with the number of layers, these methods suffer from a significant communication and storage overhead.
To alleviate these issues, \citep{md2021distgnn} proposes to split the original graph using a min-cut graph partition algorithm that can minimize the number of cut-edges.
\citep{jiang2021communication} proposes to use importance sampling to assign nodes on the local machine with a higher probability. 
\citep{angerd2020distributed} proposes to sample and save a small subgraph from other local machines as an approximation of the original graph structure. 
Nonetheless, these methods are limited to a very shallow GNN structure and suffer from significant performance degradation when the original graph is dense.
($2$) \textit{Input and hidden feature communication-based methods}: These methods propose to communicate hidden features in addition to the input node features.
Although these methods reduce the number of transferred bytes during each communication round (due to the smaller size of hidden embedding and less required nodes features), the number of communication rounds grows linearly as the number of layers, and are prone to more communication delay.
To address these issues, in addition to optimal partitioning of the graph, \citep{zheng2020distdgl} proposes to use sparse embedding to reduce the number of bytes to communicate and \citep{tripathy2020reducing} proposes several graph partitioning techniques to diminish the communication overhead.


\section{Background and problem formulation}
\label{section:background}

In this section, we start by describing Graph Convolutional Network (GCN) and its training algorithm on a single machine, then formulate the problem of distributed GCN training. 
Note that we use GCN with mean aggregation for simplicity, however, our discussion is also applicable to other GNN architectures, such as SAGE~\citep{hamilton2017inductive}, GAT~\citep{velivckovic2017graph}, ResGCN~\citep{li2019deepgcns} and APPNP~\citep{klicpera2018predict}. 


\noindent\textbf{Training GCN on a single machine.~}
Here, we consider the semi-supervised node classification in an undirected graph $\mathcal{G}(\mathcal{V},\mathcal{E})$ with $N = |\mathcal{V}|$ nodes and $|\mathcal{E}|$ edges.
Each node $v_i \in \mathcal{V}$ is associated with a pair $(\mathbf{x}_i, \mathbf{y}_i)$, where $\mathbf{x}_i \in \mathbb{R}^d$ is the input feature vector,  $\mathbf{y}_i\in\mathbb{R}^{|\mathcal{C}|}$ is the ground truth label, and $\mathcal{C}$ is the candidate labels in the multi-class classifications. 
Besides, let $\mathbf{X} = [\mathbf{x}_1,\ldots, \mathbf{x}_N] \in\mathbb{R}^{N\times d}$ denote the input node feature matrix. 
Our goal is to find a set of parameters $\boldsymbol{\theta}=\{\mathbf{W}^{(\ell)}\}_{\ell=1}^L $ by minimizing the empirical loss $\mathcal{L}(\boldsymbol{\theta})$ over all nodes in the training set, i.e.,
\begin{equation} \label{eq:update_rule_convergence_server}
\mathcal{L}(\bm{\theta}) = \frac{1}{N}\sum\nolimits_{i\in\mathcal{V}} \phi (\mathbf{h}_i^{(L)}, \mathbf{y}_i ),\qquad 
    \mathbf{h}_i^{(\ell)} = \sigma\Big( \frac{1}{|\mathcal{N}(v_i)|}\sum\nolimits_{j\in\mathcal{N}(v_i)} \mathbf{h}_j^{(\ell-1)} \mathbf{W}^{(\ell)} \Big),
\end{equation}
where $\phi(\cdot,\cdot)$ is the loss function (e.g., cross entropy loss), $\sigma(\cdot)$ is the activation function (e.g., ReLU), and $\mathcal{N}(v_i)$ is the neighborhood of node $v_i$.
In practice, we can update the model parameters by the stochastic gradient computed on a sampled mini-batch (using full-neighbors) by
\begin{equation} \label{eq:update_rule_convergence_server_sgd}
    \tilde{\nabla}\mathcal{L}(\bm{\theta},\xi) = \frac{1}{B}\sum\nolimits_{i\in\xi} \nabla \phi (\mathbf{h}_i^{(L)}, \mathbf{y}_i ),\qquad
\end{equation}
where $\xi$ denotes an i.i.d. sampled mini-batch of size $B$ and we have $\mathbb{E}[\tilde{\nabla}\mathcal{L}(\bm{\theta},\xi)]  = \nabla \mathcal{L}(\bm{\theta})$.


\begin{figure}[htb]
{\centering
\begin{minipage}{0.85\linewidth}
\vspace{-15pt}
\begin{algorithm}[H]
 \caption{Distributed GCN training with ``\textit{Parallel SGD with Periodic Averaging}''}
 \label{alg:distgnn}

\small
\begin{algorithmic}[1]
    \Require 
    Global parameters $\bar{\boldsymbol{\theta}}^0$, local parameters $\boldsymbol{\theta}^0_p=\bar{\boldsymbol{\theta}}^0$, time-step $t=0$, learning rate $\eta$.
    \For {$r$}{$1$}{$R$}
        \tikzmk{A}
        \ForP {$p$}{$1$}{$P$} \Comment{Parallel training on local machines}
            \State Local machine $p$ receives the global parameters $\boldsymbol{\theta}^{t}_p \gets \bar{\boldsymbol{\theta}}^t$. \Comment{Communication}
            \For {$k$}{$1$}{$K$}
                \State $t \gets t + 1$.
                \State Local machine $p$ constructs the mini-batch $\xi_p^t$ with neighbor sampling.
                \State Local machine $p$ computes 
                the stochastic gradients $\tilde{\nabla} \mathcal{L}_p^\text{local}(\bm{\theta}_p^t, \xi_p^t)$. 
                \State Local machine $p$ updates the local parameter by $\boldsymbol{\theta}^{t+1}_p =  \boldsymbol{\theta}^{t}_p - \eta \tilde{\nabla} \mathcal{L}_p^\text{local}(\bm{\theta}_p^t, \xi_p^t)$.\label{line:model_update_localsgd}
            \EndFor
            \State Local machine $p$ sends the local parameters $\boldsymbol{\theta}^{t+1}_p$ to the server. \label{line:send_to_server_localsgd} \Comment{Communication}
        \EndFor \tikzmk{B}\boxitf{codell}
        \State Server updates the global parameters by parameter averaging $\bar{\boldsymbol{\theta}}^{t+1} = \frac{1}{P}\sum_{p=1}^{P} \boldsymbol{\theta}^{t+1}_p$. \label{line:model_avg_localsgd}
    \EndFor
    \Ensure Server returns trained \texttt{GCN} model with $\min_t \mathbb{E}[ \| \nabla \mathcal{L}(\bar{\boldsymbol{\theta}}^{t}) \|^2 ]$.
\end{algorithmic}
\end{algorithm} 
\end{minipage}
\par
}
\end{figure}

\noindent\textbf{Distributed GCN training with periodic averaging.~}
In this paper, we consider the distributed learning setting with $P$ local machines and a single parameter server. 
The original input graph $\mathcal{G}$ is partitioned into $P$ subgraphs, where $\mathcal{G}_p(\mathcal{V}_p, \mathcal{E}_p)$ denotes the subgraph on the $p$-th local machine with $N_p = |\mathcal{V}_p|$ nodes, and $\mathbf{X}_p \in \mathbb{R}^{N_p \times d}$ as the input feature of all nodes in $\mathcal{V}_p$ located on the $p$-th machine.
Then, the full-batch local gradient $\nabla \mathcal{L}_p^\text{local}(\boldsymbol{\theta}_p)$ is computed as
\begin{equation} \label{eq:update_rule_convergence}
    \nabla \mathcal{L}_p^\text{local}(\bm{\theta}_p) = \frac{1}{N_p}\sum\nolimits_{i\in\mathcal{V}_p} \nabla \phi( \mathbf{h}_i^{(L)}, \mathbf{y}_i ),~~
    \mathbf{h}_i^{(\ell)} = \sigma\Big( \frac{1}{|\mathcal{N}_p(v_i)|} \sum\nolimits_{j\in\mathcal{N}_p(v_i)} \mathbf{h}_j^{(\ell-1)} \mathbf{W}_p^{(\ell)} \Big),
\end{equation}
where $\bm{\theta}_p = \{\mathbf{W}_p^{(\ell)}\}_{\ell=1}^L$ is the model parameters on the $p$-th local machine, $\mathcal{N}_p(v_i) = \{v_j | (v_i, v_j) \in \mathcal{E}_p\}$ is the local neighbors of node $v_i$ on the $p$-th local machine.
When the graph is large, the computational complexity of forward and backward propagation could be very high. One practical solution is to compute the stochastic gradient on a sampled mini-batch with neighbor sampling, i.e.,
\begin{equation} \label{eq:update_rule_convergence_neighbor_sampling}
    \tilde{\nabla} \mathcal{L}_p^\text{local}(\bm{\theta}_p,\xi_p) = \frac{1}{B_p}\sum\nolimits_{i\in\xi_p} \nabla \phi( \tilde{\mathbf{h}}_i^{(L)}, \mathbf{y}_i ),~~
    \tilde{\mathbf{h}}_i^{(\ell)} = \sigma\Big( \frac{1}{|\tilde{\mathcal{N}}_p(v_i)|} \sum\nolimits_{j\in\tilde{\mathcal{N}}_p(v_i)} \tilde{\mathbf{h}}_j^{(\ell-1)} \mathbf{W}_p^{(\ell)} \Big),
\end{equation}
where $\xi_p$ is an i.i.d. sampled mini-batch of $B_p$ nodes, $\tilde{\mathcal{N}}_p(v_i) \subset \mathcal{N}(v_i)$ is the sampled neighbors.

An illustration of distributed GCN training with \textit{Parallel SGD with Periodic Averaging} (PSGD-PA) is summarized in Algorithm~\ref{alg:distgnn}.
Before training, the server maintains a global model $\bar{\bm{\theta}}^0$ and each local machine keeps a local copy of the same model $\bm{\theta}_p^0$. 
During training, the local machine first updates the local model $\bm{\theta}_p^t$ using the stochastic gradient $\smash{\tilde{\nabla} \mathcal{L}_p^\text{local}(\bm{\theta}_p^t, \xi_p^t)}$ computed by Eq.~\ref{eq:update_rule_convergence_neighbor_sampling} for $K$ iterations (line~\ref{line:model_update_localsgd}), then sends the local model $\bm{\theta}_p^t$ to the server (line~\ref{line:send_to_server_localsgd}). 
At each communication step, the server collects and averages the model parameters from the local machines (line~\ref{line:model_avg_localsgd}) and send the averaged model $\boldsymbol{\theta}^{t+1}_p$ back to each local machine. 

\noindent\textbf{Limitations.~}
Although PSGD-PA can significantly reduce the communication overhead by transferring the locally trained models instead of node feature/embeddings (refer to Figure~\ref{figure:intro-comm}), it suffers from performance degeneration due to ignorance of the cut-edges (refer to Figure~\ref{figure:intro-acc}).
In the next section, we introduce a communication-efficient algorithm \CG{} that does not suffer from this issue, and can achieve almost the same performance as training the model on a single machine. 

\section{Proposed Algorithm: Learn Locally Correct Globally}
\label{section:method}

In this section, we describe \textit{Learn Locally, Correct Globally} (\textsc{LLCG}) for distributed  GNN training.
LLCG includes two main phases, \textit{local training with periodic model averaging} and \textit{global server correction}, to help reduce both the number of required communications and size of transferred data,  without compromising the predictive accuracy. 
We summarize the details of \textsc{LLCG} in Algorithm~\ref{alg:distgnn-corr}.

\setlength{\textfloatsep}{5pt}
\begin{figure}[htb]
{\centering
\begin{minipage}{0.86\linewidth}
\vspace{-15pt}
\begin{algorithm}[H]
 \caption{Distributed GCN training by ``\textit{Learn Locally, Correct Globally}'' }
 \label{alg:distgnn-corr}
 \small
 \begin{algorithmic}[1]
    \Require 
    Global parameters $\bar{\boldsymbol{\theta}}^0$,
    local parameters $\boldsymbol{\theta}^0_p$, time-step $t=0$, local step size hyper-parameters $K,\rho$, and learning rate $\gamma, \eta$
    \For {$r$}{$1$}{$R$}
        \tikzmk{A}
        \ForP {$p$}{$1$}{$P$} \Comment{Parallel training on local machine}
            \State Local machine $p$ receives the global parameters $\boldsymbol{\theta}^{t}_p \gets \bar{\boldsymbol{\theta}}^t$ \Comment{Communication}\label{line:local_receive_model}
            \For {$k$}{$1$}{$K\rho^r$} \label{line:local_update_start}
                \State $t \gets t + 1$
                \State Local machine $p$ constructs the mini-batch $\xi_p^t$ with neighbor sampling
                \State Local machine $p$ computes 
                stochastic gradients $\tilde{\nabla} \mathcal{L}_p^\text{local}(\bm{\theta}_p^t, \xi_p^t)$ 
                \State Local machine $p$ updates model parameter by $\boldsymbol{\theta}^{t+1}_p =  \boldsymbol{\theta}^{t}_p - \eta \tilde{\nabla} \mathcal{L}_p^\text{local}(\bm{\theta}_p^t, \xi_p^t)$
            \EndFor \label{line:local_update_end}
            \State Local machine $p$ sends the local parameters $\boldsymbol{\theta}^{t+1}_p$ to the server \Comment{Communication} \label{line:server_get_local}
        \EndFor \tikzmk{B}\boxitf{codell}
        \State Server updates the global parameters using parameter averaging $\bar{\boldsymbol{\theta}}^{t+1} = \frac{1}{P}\sum_{p=1}^{P} \boldsymbol{\theta}^{t+1}_p$ \label{line:model_avg}
        \tikzmk{A}
        \For {$s$}{$1$}{$S$} \Comment{Server Correction} \label{line:server_correction_start}
            \State $t \gets t + 1$
            \State Server constructs a mini-batch $\xi^t$ with full-neighbors \label{line:server_mini_batch_full_neighbors}
            \State Server computes 
            the stochastic gradient $\tilde{\nabla} \mathcal{L}(\bar{\bm{\theta}}^t, \xi^t)$ \label{line:server_compute_gradient}
            \State Server updates the global parameters by $\bar{\boldsymbol{\theta}}^{t+1} =  \bar{\boldsymbol{\theta}}^{t} - \gamma \tilde{\nabla} \mathcal{L}(\bar{\bm{\theta}}^t, \xi^t)$ \label{line:server_update_model}
        \EndFor \tikzmk{B} \boxits{codecg}  \label{line:server_correction_end}
    \EndFor
    \Ensure Server return \texttt{GCN} model with trained $\min_t \mathbb{E}[ \| \nabla \mathcal{L}(\bar{\boldsymbol{\theta}}^{t}) \|^2 ]$
\end{algorithmic}
\end{algorithm}
\end{minipage}
\par
}
\end{figure}

\setlength{\textfloatsep}{10pt}

\subsection{local training with periodic model averaging} \label{section:local_train_periodic_model_avg}

At the beginning of a local epoch, each local machine receives the latest global model parameters from the server (line~\ref{line:local_receive_model}). 
Next, each local machine runs $K \rho^r$ iterations to update the local model (line~\ref{line:local_update_start} to~\ref{line:local_update_end}), where $K$ and $\rho$ are the hyper-parameters that control the local epoch size.
Note that instead of using a fixed local epoch size as Algorithm~\ref{alg:distgnn}, we choose to use exponentially increasing local epoch size in \CGE{} with $\rho>1$.
The reasons are as follows.

At the beginning of the training phase, 
all local models $\bm{\theta}_p^t$ are far from the optimal solution and will receive a gradient $\smash{\tilde{\nabla} \mathcal{L}_p^\text{local}(\bm{\theta}_p^t, \xi_p^t)}$ computed by Eq.~\ref{eq:update_rule_convergence_neighbor_sampling}. 
Using a smaller local update step at the early stage guarantees each local model does not diverge too much from each other before the model averaging step at the server side (line~\ref{line:model_avg}).
However, towards the end of the training, 
all local models $\bm{\theta}_p^t$ will receive relatively smaller gradient $\smash{\tilde{\nabla} \mathcal{L}_p^\text{local}(\bm{\theta}_p^t, \xi_p^t)}$, such that we can chose a larger local epoch size to reduce the number of communications, without worrying about the divergence of local models.
By doing so, after total number of $\smash{T=\sum_{r=1}^R K \rho^r}$ iterations, \CG{} only requires $R=\log_\rho \frac{T}{K}$ rounds of communications.
Therefore, compared to the fully-synchronous method, we can significantly reduce the total number of communications from $\mathcal{O}(T)$ to $\smash{\mathcal{O}(\log_\rho \frac{T}{K})}$.

\subsection{Global server correction}

The design of the global server correction is to ensure that the trained model not only learns from the data on each local machine, but also learns the global structure of the graph, thus reducing the information loss caused by graph partitioning and avoiding cut-edges.
Before the correction, the server receives the locally trained models from all local machines (line~\ref{line:server_get_local}) and applies model parameter averaging (line~\ref{line:model_avg}). 
Next, $S$ server correction steps are applied on top of the averaged model (line~\ref{line:server_correction_start} to~\ref{line:server_correction_end}).
During the correction, the server first constructs a mini-batch $\xi^t$ using full-neighbors\footnote{Note that using full neighbors is required for the server correction but not the local machines}(line~\ref{line:server_mini_batch_full_neighbors}), compute the stochastic gradient $\tilde{\nabla} \mathcal{L}(\bar{\bm{\theta}}^t, \xi^t)$ on the constructed mini-batch by Eq.~\ref{eq:update_rule_convergence_server_sgd} (line~\ref{line:server_compute_gradient}) and update the averaged model $\bar{\bm{\theta}}^t$ for $S$ iterations (line~\ref{line:server_update_model}).
The number of correction steps $S$
\footnote{In practice, we found $S=1$ or $S=2$ works well on most datasets.}
depends on the heterogeneity among the subgraphs on each local machine: the more heterogeneous the subgraphs are, the more correction steps are required to better refine the averaged model and reduce the divergence across the local models.
Note that, the heterogeneity is minimized when employing GGS (Figure~\ref{figure:problems}) with the local machines having access to the full graph, as a result. 
However, GGS requires sampling from the global graph and communication at every iteration, which results in additional overhead and lower efficiency.
Instead, in LLCG we are trading computation on the server for the costly feature communication, and only requires periodic communication.

\vspace{-5pt}
\section{Theoretical Analysis}
\label{section:theory}

\begin{figure}[t]
    \centering
    \includegraphics[width=0.85\textwidth]{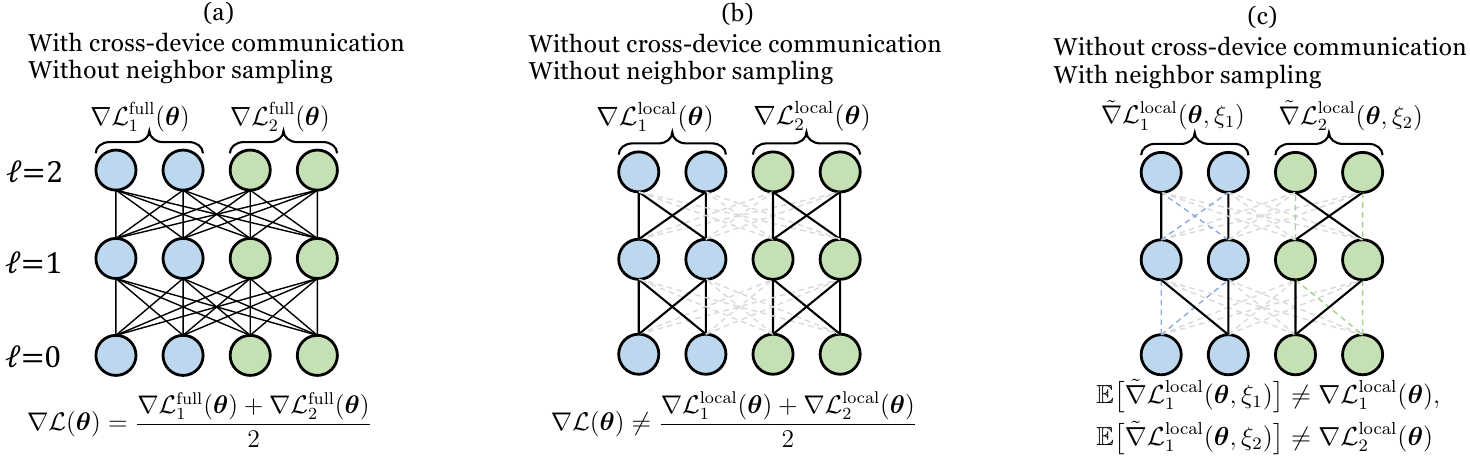}
    \vspace{-8pt}
    \caption{Comparison of notations
    $\nabla \mathcal{L}_p^\text{local}(\bm{\theta})$,  $\tilde{\nabla} \mathcal{L}_p^\text{local}(\bm{\theta},\xi_p)$, and $\nabla \mathcal{L}_p^\text{full}(\bm{\theta})$
    on two local machines, where the blue node and green circles represent nodes on different local machines.
    }
    \vspace{-5pt}
    \label{fig:diff_loss_local_global}
\end{figure}

In this section, we provide the convergence analysis on the distributed training of GCN under two different settings, i.e., with and without server correction.
We first introduce the notations and assumptions for the analysis (Section~\ref{section:notation_and_assumption}). 
Then, we show that periodic averaging of local machine models alone and ignoring the global graph structure will suffer from an irreducible residual error (Section~\ref{section:algorithm_param_avg}). 
Finally, we show that this residual error can be eliminated by running server correction steps after each periodic averaging step on the server (Section~\ref{section:algorithm_correction}). 

\subsection{Notations and assumptions} \label{section:notation_and_assumption}
Let us first recall the notations defined in Section~\ref{section:background}, where $\mathcal{L}(\bm{\theta})$ denotes the global objective function computed using the all node features $\mathbf{X}$ and the original graph $\mathcal{G}$, $\mathcal{L}_p(\bm{\theta})$ denotes the local objective function computed using the local node features $\mathbf{X}_p$ and local graph $\mathcal{G}_p$,
$\bm{\theta}_p^t$ denotes the model parameters on the $p$-th local machine at the $t$-th step, and $\smash{\bar{\bm{\theta}}^t = \frac{1}{P}\sum_{p=1}^P \bm{\theta}_p^t}$ denotes the virtual averaged model at the $t$-th step. 
In the non-convex optimization, our goal is to show the expected gradient of the global objective on the virtual averaged model parameters $\mathbb{E}[\| \nabla \mathcal{L}(\bar{\bm{\theta}}^t) \|^2]$ decreases as the number of local machines $P$ and the number of training steps $T$ increase. 
Besides, we introduce $\nabla \mathcal{L}_p^\text{full}(\bm{\theta})$ as the gradient computed on the  $p$-th local machine but have access the full node features $\mathbf{X}$ and the original graph structure $\mathcal{G}$ as
\begin{equation}
    \nabla \mathcal{L}_p^\text{full}(\bm{\theta}) = \frac{1}{|\mathcal{V}_p|}\sum\nolimits_{i\in\mathcal{V}_p} \nabla \phi (\mathbf{h}_i^{(L)}, y_i ),\qquad
    \mathbf{h}_i^{(\ell)} = \sigma\Big(  \frac{1}{|\mathcal{N}(v_i)|}\sum\nolimits_{j\in\mathcal{N}(v_i)}  \mathbf{h}_j^{(\ell-1)} \mathbf{W}_p^{(\ell)} \Big).
\end{equation}
Please refer to Figure~\ref{fig:diff_loss_local_global} for an illustration of different gradient computations.
Besides, we introduce \textit{local-global gradient discrepancy} as $\kappa^2 = \kappa^2_{\mathbf{A}} + \kappa^2_{\mathbf{X}} $, where
$\kappa^2_{\mathbf{A}} = \max_{p \in [P]} \{ \| \nabla \mathcal{L}_p^\text{local}(\bm{\theta}) - \nabla \mathcal{L}_p^\text{full}(\bm{\theta}) \|^2 \}$ is  the maximum difference between the gradient computed on the local machine with and without having access to the global graph structure, which is mainly due to fact that the local machines are oblivious to the full graph information; and $\kappa^2_{\mathbf{X}} = \max_{p \in [P]} \{ \| \nabla \mathcal{L}_p^\text{full}(\bm{\theta}) - \nabla \mathcal{L}(\bm{\theta}) \|^2 \}$ is the maximum difference between the gradient computed using the local node and all nodes, which is mainly due to the heterogeneity of the node features on each local machine, and we have $\kappa^2_{\mathbf{X}}=0$ if the nodes are i.i.d. sampled to each local machine. 
Notice that \textit{local-global gradient discrepancy} $\kappa^2$ plays an important role in our theoretical results.

For the convergence analysis, we make the following standard assumptions.
\vspace{-5pt}
\begin{assumption} \label{assumption:bound_local_variance} 
The stochastic gradient on the $p$-th local machine (with neighbor sampling) has stochastic gradient variance bounded by $\sigma_\text{var}^2$ and stochastic gradient bias bounded by $\sigma_\text{bias}^2$, i.e., 
$ \mathbb{E}[\| \tilde{\nabla} \mathcal{L}_p^\text{local}(\boldsymbol{\theta};\xi) - \mathbb{E}[\tilde{\nabla} \mathcal{L}_p^\text{local}(\boldsymbol{\theta};\xi)]\|^2 ] \leq \sigma^2_\text{var},~
    \mathbb{E}[\| \mathbb{E}[\tilde{\nabla} \mathcal{L}_p^\text{local}(\boldsymbol{\theta};\xi)] - \nabla \mathcal{L}_p^\text{local}(\boldsymbol{\theta})\|^2 ] \leq \sigma^2_\text{bias}.$
\end{assumption}
\vspace{-5pt}
\begin{assumption} \label{assumption:bound_local_variance_2} 
The stochastic gradient for global server correction (with full neighbors) has stochastic gradient variance bounded by $\sigma_\text{global}^2$, i.e., $\mathbb{E}[\| \tilde{\nabla} \mathcal{L}^\text{full}_p(\boldsymbol{\theta};\xi) - \nabla \mathcal{L}_p^\text{full}(\boldsymbol{\theta})]\|^2 ] \leq \sigma^2_\text{global}$.
\end{assumption}
\vspace{-5pt}
The existence of stochastic gradient bias and variance in sampling-based GNN training have been studied in~\citep{cong2020minimal,cong2021importance}, where~\citep{cong2021importance} further quantify the stochastic gradient bias and variance as a function of the number of GCN layers.
In particular, they show that the existence of $\sigma^2_\text{bias}$ is due to neighbor sampling and non-linear activation, and we have $\sigma^2_\text{bias}=0$ if all neighbors are used or the non-linear activation is removed.
The existence of $\sigma^2_\text{var}$ is because we are sampling mini-batches to compute the stochastic gradient on each local machine during training. As the mini-batch size increases, $\sigma^2_\text{var}$ will be decreasing, and we have $\sigma^2_\text{var}=0$ when using full-batch.

\subsection{Distributed GNN via Parameter Averaging}\label{section:algorithm_param_avg}

In the following, we provide the first convergence analysis on distributed training of GCN. 
We show that solely periodic averaging of the local machine models and ignoring the global graph structure suffers from an upper bound that is irreducible with the number of training steps. 
Comparing to the traditional distributed training (e.g., distributed training Convolutional Neural Network for image classification~\citep{dean2012large,li2019convergence}),
the key challenges in the distributed GCN training is the two different types of gradient bias:
($1$) The expectation of the local full-batch gradient is a biased estimation of the global full-batch gradient, i.e., $\smash{\frac{1}{P} \sum_{p=1}^P \nabla \mathcal{L}_p^\text{local}(\bm{\theta}) \neq \nabla \mathcal{L}(\bm{\theta})}$. This is because each local machine does not have access to the original input graph and full node feature matrix. Note that the aforementioned equivalence is important for the classifcal distributed training analysis~\cite{dean2012large,yu2019parallel}. ($2$) The expectation of the local stochastic gradient is a biased estimation of the local full-batch gradient i.e., $\mathbb{E}[\tilde{\nabla} \mathcal{L}_p^\text{local}(\bm{\theta}, \xi_p)] \neq \nabla \mathcal{L}_p^\text{local}(\bm{\theta})$. This is because the stochastic gradient on each local machine is computed by using neighbor sampling, which has been studied in~\citep{cong2021importance}.

\begin{theorem} [Distributed GCN via Parameter Averaging] \label{theorem:distgnn_param_avg}
Consider applying model averaging for GNN training under Assumption~\ref{assumption:bound_local_variance} and~\ref{assumption:bound_local_variance_2}.
If we choose learning rate $\eta = \frac{\sqrt{P}}{\sqrt{T}}$ and the local step size $K \leq  \smash{\frac{\sqrt{2}T^{1/4}}{8L P^{3/4}}}$, then for any $T \geq L^2 P$ steps of gradient updates we have
\begin{equation*}
    \frac{1}{T}\sum\nolimits_{t=0}^{T-1} \mathbb{E}[ \| \nabla \mathcal{L}(\bar{\bm{\theta}}^t) \|^2 ]  = \mathcal{O}\left(\frac{1}{\sqrt{PT}}\right) + \mathcal{O}(\kappa^2 + \sigma_\text{bias}^2) .
\end{equation*}
\end{theorem}
\vspace{-5pt}
Theorem~\ref{theorem:distgnn_param_avg} implies that, by carefully choosing the learning rate $\eta$ and the local step size $K$, the gradient norm computed on the virtual averaged model is bounded by $\smash{\mathcal{O}({1}/{\sqrt{PT}})}$ after $R = T/K = \smash{\mathcal{O}(\frac{P^{3/4}}{T^{3/4}})}$ communication rounds, but suffers from an irreducible residual error upper bound $\mathcal{O}(\kappa^2 + \sigma_\text{bias}^2)$.
In the next section, we show that this residual error can be eliminated by applying server correction.

\subsection{Distributed GCN via Server Correction}\label{section:algorithm_correction}

Before proceeding to our result, in order to simplify the presentation, let us first define the notation $\smash{G_\text{global}^r = \min_{t \in \mathcal{T}_\text{global}(r)} \mathbb{E}[ \| \nabla \mathcal{L}(\bar{\bm{\theta}}^t) \|^2]}$ and $\smash{G_\text{local}^r=\min_{t \in \mathcal{T}_\text{local}(r)} \mathbb{E}\big[ \big\| \frac{1}{P}\sum_{p=1}^P \nabla \mathcal{L}_p^\text{local}(\bm{\theta}_p^t) \big\|^2\big]}$ as the minimum gradient computed at the $r$-th round global and local step, where $\mathcal{T}_\text{global}(r)$ and $\mathcal{T}_\text{local}(r)$ are the number of iteration run after the $r$-th communication round on server and local machine, respectively. Please refer to Eq.~\ref{eq:T_local_server} in Appendix~\ref{section:thm2_main} for a formal definition.
\vspace{-5pt}
\begin{theorem} \label{theorem:distgnn_global_correction}
Consider applying model averaging for GCN training under Assumption~\ref{assumption:bound_local_variance} and~\ref{assumption:bound_local_variance_2}.
If we choose learning rate $\gamma=\eta = \frac{\sqrt{P}}{\sqrt{T}}$, the local step size $K,\rho$ such that $\sum_{r=1}^R K^2 \rho^{2r} \leq \frac{R T^{1/2}}{32 L^2 P^{3/2}}$, and server correction step size $S = \max_{r\in[R]} \big( \frac{\kappa^2 + 2\sigma_\text{bias}^2}{1-L(\sqrt{P/T})} - G_\text{local}^r\big) \frac{K\rho^r}{G_\text{local}^r}$, then for any 
$T \geq L^2 P$ steps of gradient updates we have: $\frac{1}{T}\sum\nolimits_{t=1}^T \mathbb{E}[ \| \nabla \mathcal{L}(\bar{\bm{\theta}}^t) \|^2 ] = \mathcal{O}\big(\frac{1}{\sqrt{PT}}\big).$
\end{theorem}

\vspace{-5pt}
Theorem~\ref{theorem:distgnn_global_correction} implies that, by carefully choosing the learning rates $\gamma$ and $\eta$, the local step size hyper-parameters $K,\rho$, and the number of global correction steps $S$, 
after $T$ steps ($R$ rounds of communication),  employing parameter averaging with \textit{Global Server Correction}, we have the norm of gradient bounded by $\mathcal{O}({1}/{\sqrt{PT}})$, without suffering the residual error that exists in the naive parameter averaging (in Theorem~\ref{theorem:distgnn_param_avg}). 
Besides, the server correction step size is proportional to the scale of $\kappa^2$ and local stochastic gradient bias $\sigma_\text{bias}^2$.
The larger $\kappa^2$ and $\sigma_\text{bias}^2$, the more corrections are required to eliminate the residual error.
However, in practice, we observe that a very small number of correction steps (e.g., $S=1$) performs well, which minimizes the computation overhead on the server.

\section{Experiments}
\label{section:experiment}

\noindent\textbf{Real-world simulation.~}
In a real-world distributed setting, the server and local machines are located on different machines, connected through the network~\citep{li2020pytorch}. However, for our experiments, we only have access to a single machine with multiple GPUs. 
As a result, we simulate a real-world distributed learning scenario, such that each GPU is responsible for the computation of two local machines (8 in total) and the CPU acts as the server. 
For these reasons, in our evaluations, we opted to report the communication size and number of communication rounds, instead of the wall-clock time, which can show the benefit of distributed training. 
We argue that these are acceptable measures in real-world scenarios as well since the two main factors in distributed training are initializing connection overhead and bandwidth~\citep{tripathy2020reducing}. \vspace{1mm} \\
\noindent\textbf{Baselines.~} 
To illustrate the effectiveness of \CGE{}, we setup two general synchronized distributed training techniques as the our baseline methods, namely ``\textit{Parallel SGD with Parameter Averaging}'' (PSGD-PA) and ``\textit{Global Graph Sampling}'' (GGS), as introduced in Figure~\ref{figure:problems}, where the cut-edges in PSGD-PA are ignored and only the model parameters are transferred, but the cut-edges in GGS are considered and the node features of the cut-edges are transferred to the corresponding machine.
Note that we choose GGS as a reasonable representation for most existing proposals~\citep{md2021distgnn, zheng2020distdgl, tripathy2020reducing} for distributed GNN training, since these methods have very close communication cost and also require a large cluster of machines to truly show their performance improvement.
We also use PSGD-PA as a lower bound for communication size, which is widely used in traditional distributed training and similar to the one used in~\citep{angerd2020distributed,jiang2021communication}.
However, we did not specifically include these methods in our results since we could not reproduce their results in our settings.
Please refer to Appendix~\ref{appendix:experiments} for a detailed description of implementation, hardware specification and link to our source code. \vspace{1mm}\\
\noindent\textbf{Datasets and evaluation metric.~}
We compare \CG{} and other baselines on real-world semi-supervised node classification datasets, details of which are summarized in Table~\ref{table:datasets-detailed} in the Appendix.
The input graphs are splitted into multiple subgraphs using \texttt{METIS} before training, then the same set of subgraphs are used for all baselines.
For training, we use neighborhood sampling~\citep{hamilton2017inductive} with $10$ neighbors sampled per node and $\rho=1.1$ for \CGE{}. 
For a fair comparison, we chose the base local update step $K$ such that \CG{} has the same number of local update steps as \textsc{PSGD-PA}.
During evaluation, we use full-batch without sampling, and report the performance on the full graph using $\mathsf{AUC~ROC}$ and $\mathsf{F1~Micro}$ as the evaluation metric.
Unless otherwise stated, we conduct each experiment five times and report the mean and standard deviation.

\subsection{Primary Results}

In this section, we compare our proposed \CG{} algorithm with baselines on four datasets.
Due to space limitations we defer the detailed discussion on additional datasets to the Appendix~\ref{appendix:largedata}.

\begin{figure}[t]
    \vspace{-15pt}
    \centering
    \begin{subfigure}[t]{0.24\textwidth}
        \centering
        \includegraphics[clip, trim=25 0 15 5, width=1\linewidth]{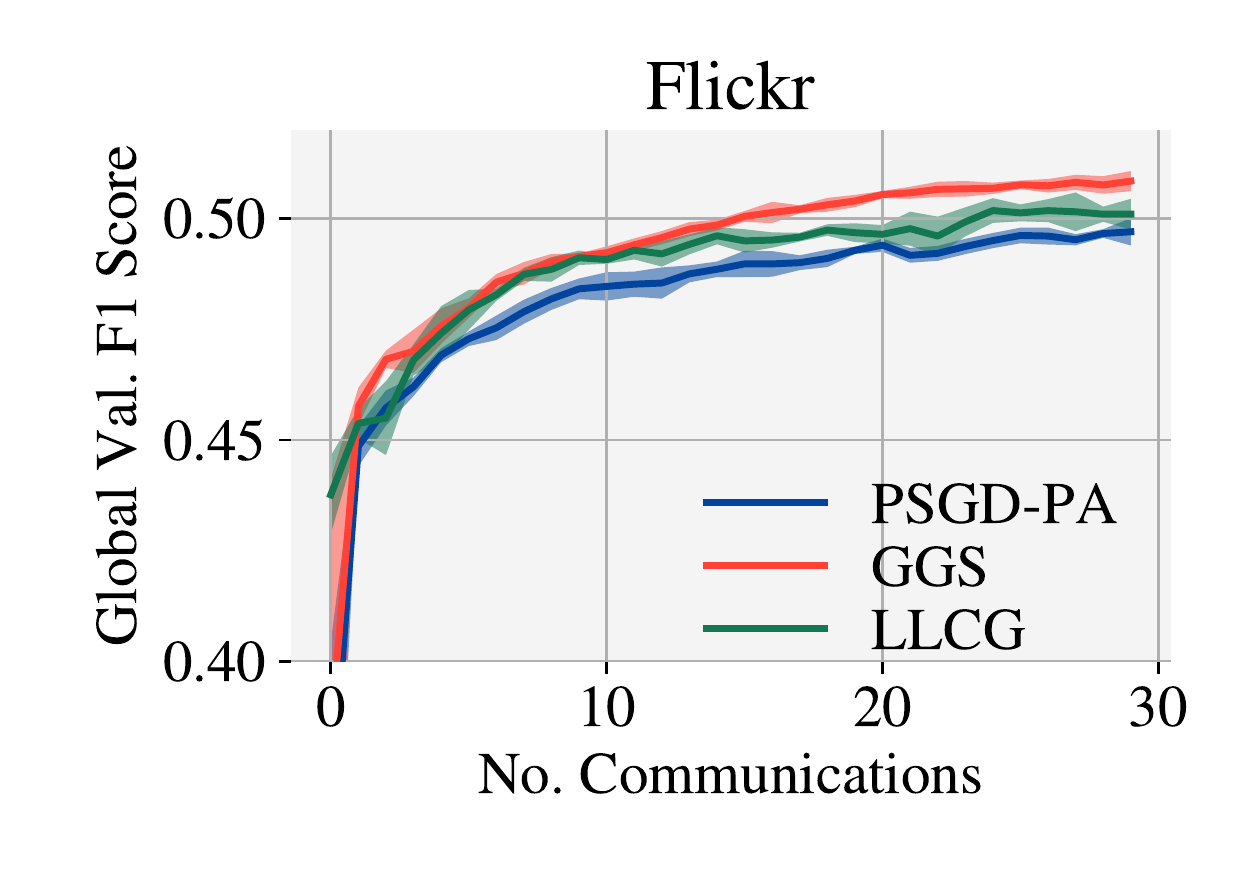}
        \vspace{-22pt}\caption{}\vspace{-3pt}
        \label{subfig:acc-flickr}
    \end{subfigure}
    \hfill
    \begin{subfigure}[t]{0.24\textwidth}
        \centering
        \includegraphics[clip, trim=25 0 15 5, width=1\linewidth]{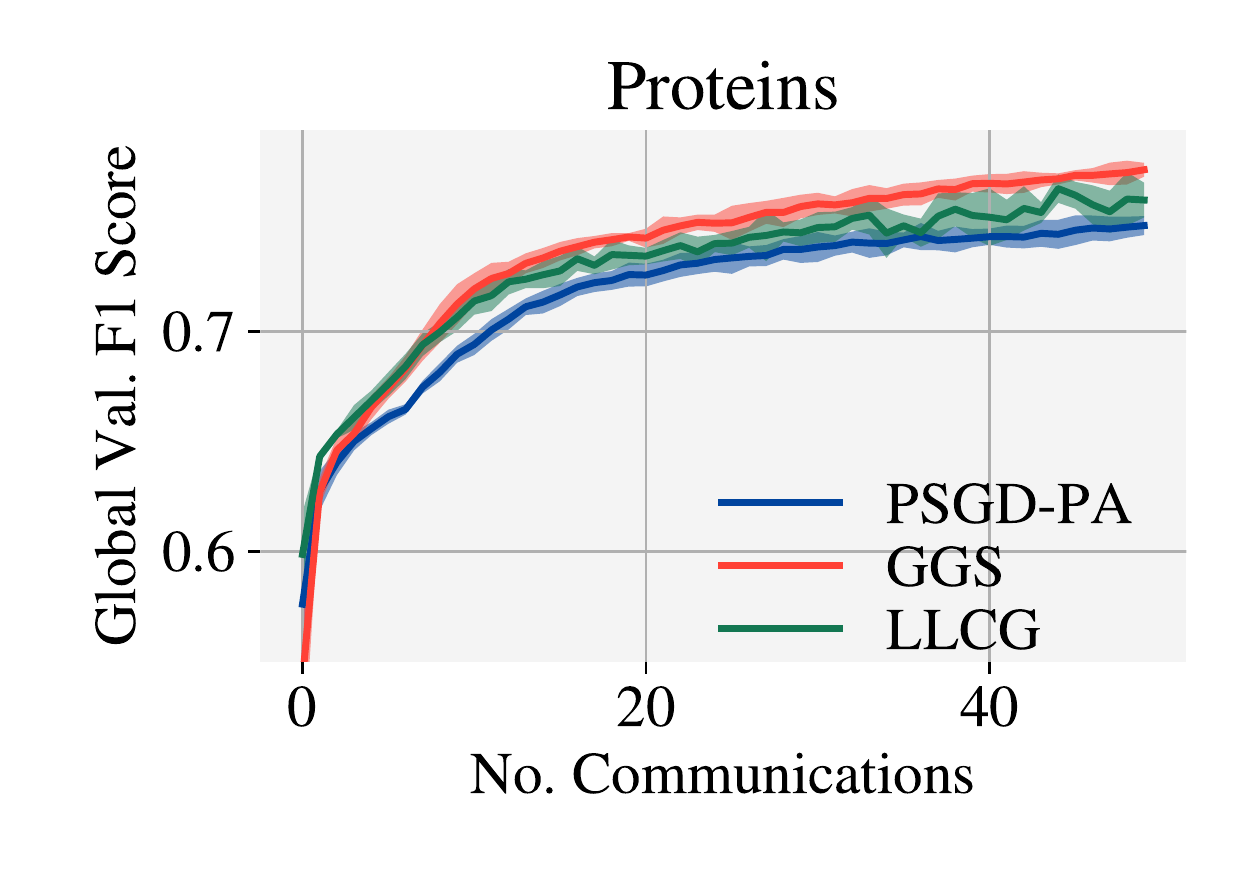}
        \vspace{-22pt}\caption{}\vspace{-3pt}
        \label{subfig:acc-proteins}
    \end{subfigure}
    \hfill
    \begin{subfigure}[t]{0.24\textwidth}
        \centering
        \includegraphics[clip, trim=25 0 15 5, width=1\linewidth]{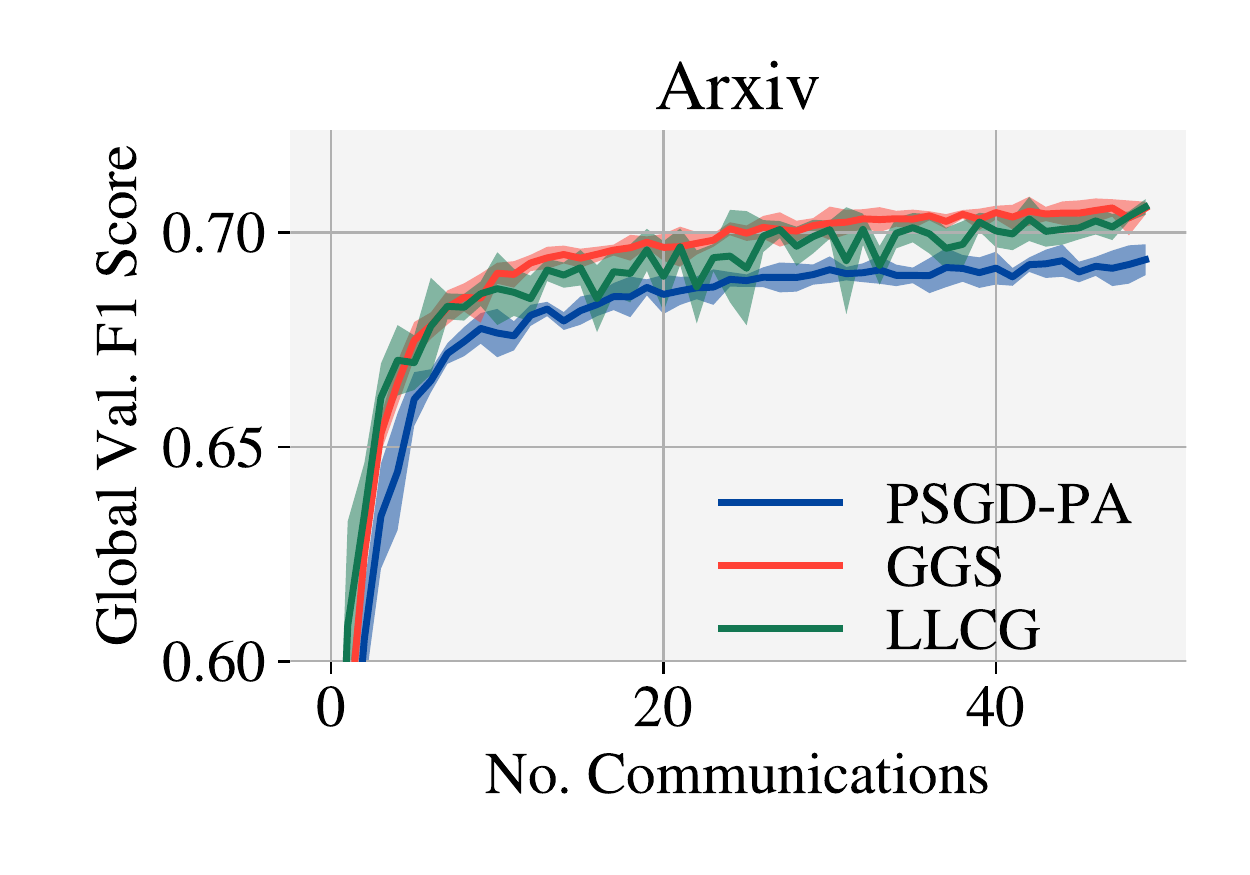}
        \vspace{-22pt}\caption{}\vspace{-3pt}
        \label{subfig:acc-arxiv}
    \end{subfigure}
    \hfill
    \begin{subfigure}[t]{0.24\textwidth}
        \centering
        \includegraphics[clip, trim=15 0 15 5, width=1\linewidth]{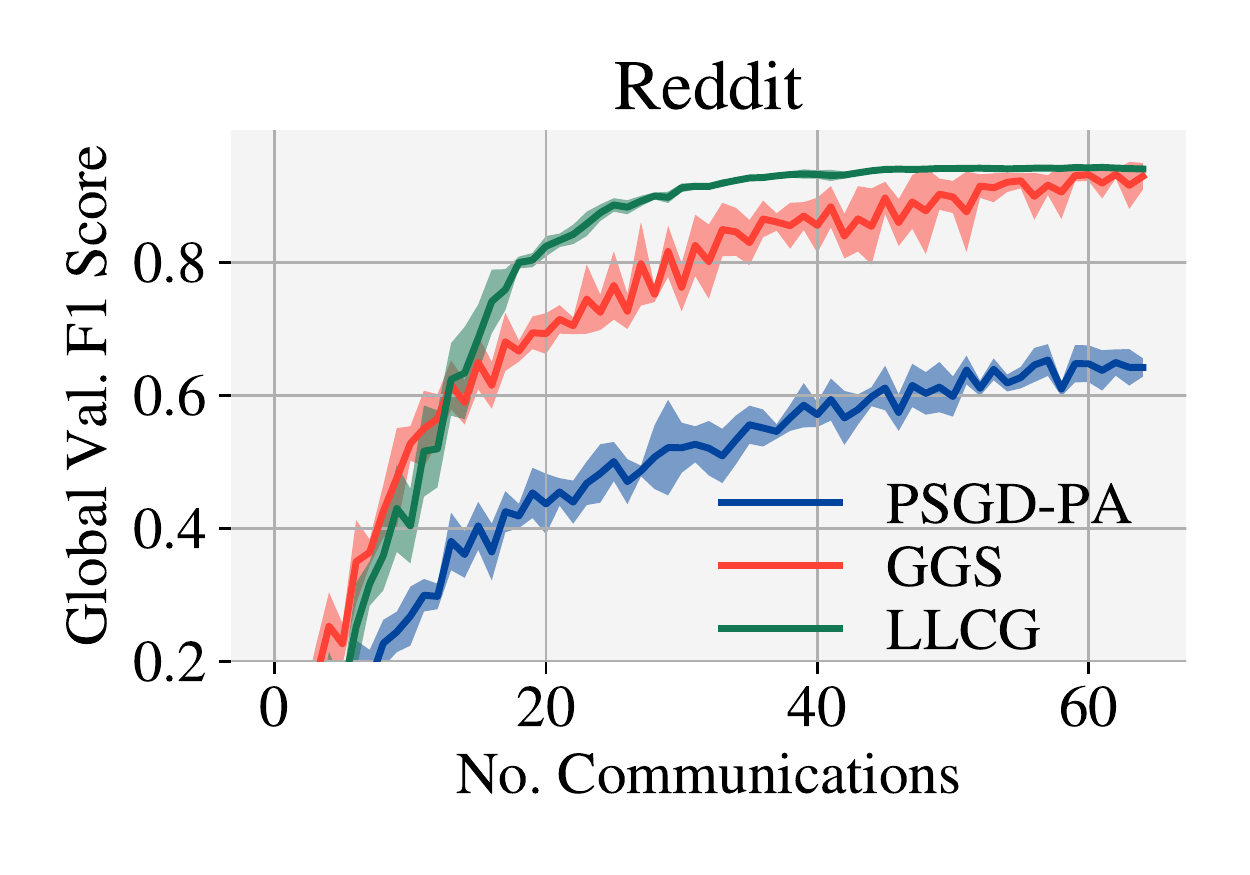}
        \vspace{-22pt}\caption{}\vspace{-3pt}
        \label{subfig:acc-reddit}
    \end{subfigure}
    \begin{subfigure}[t]{0.24\textwidth}
        \centering
        \includegraphics[clip, trim=17 0 15 10, width=1\linewidth]{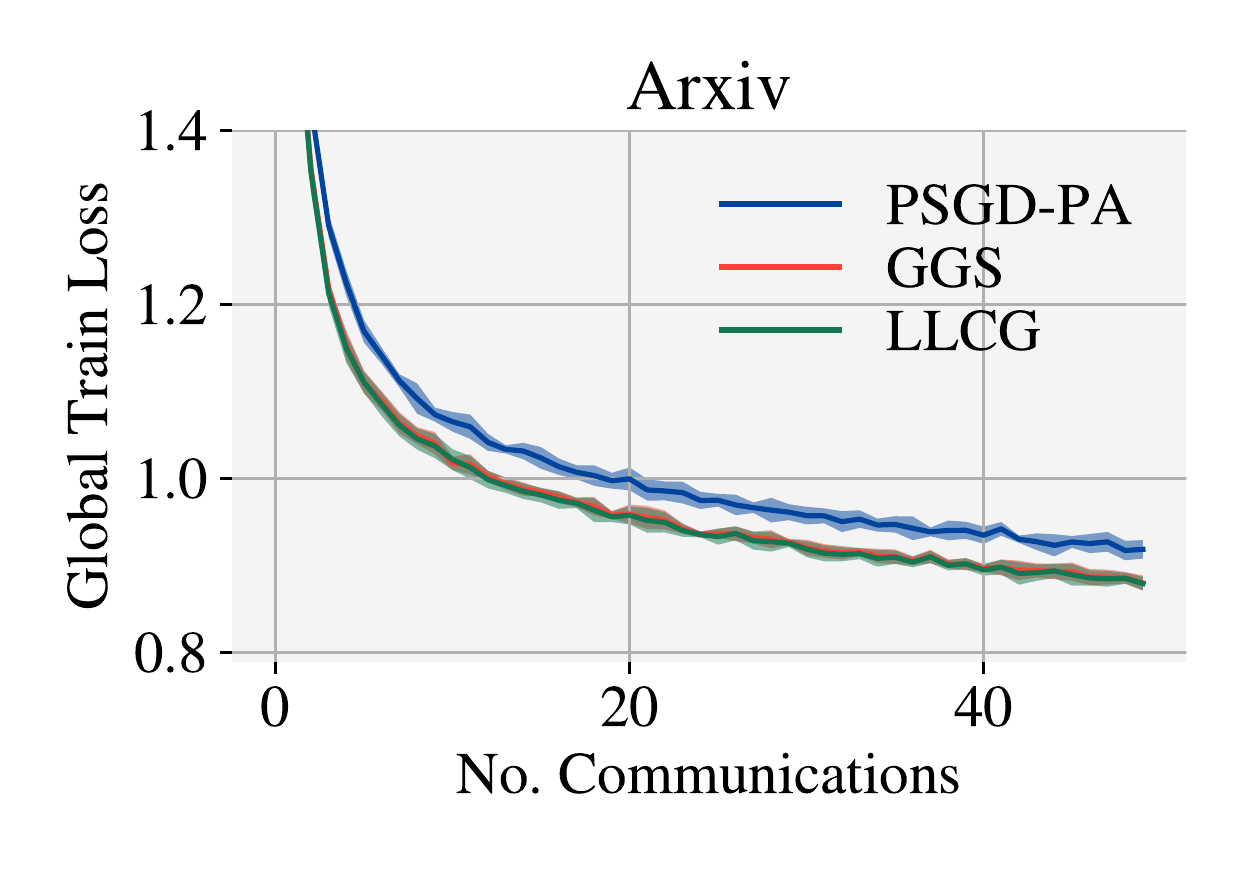}
        \vspace{-22pt}\caption{}\vspace{-3pt}
        \label{subfig:loss-arxiv}
    \end{subfigure}
    \hfill
    \begin{subfigure}[t]{0.24\textwidth}
        \centering
        \includegraphics[clip, trim=15 0 15 10, width=1\linewidth]{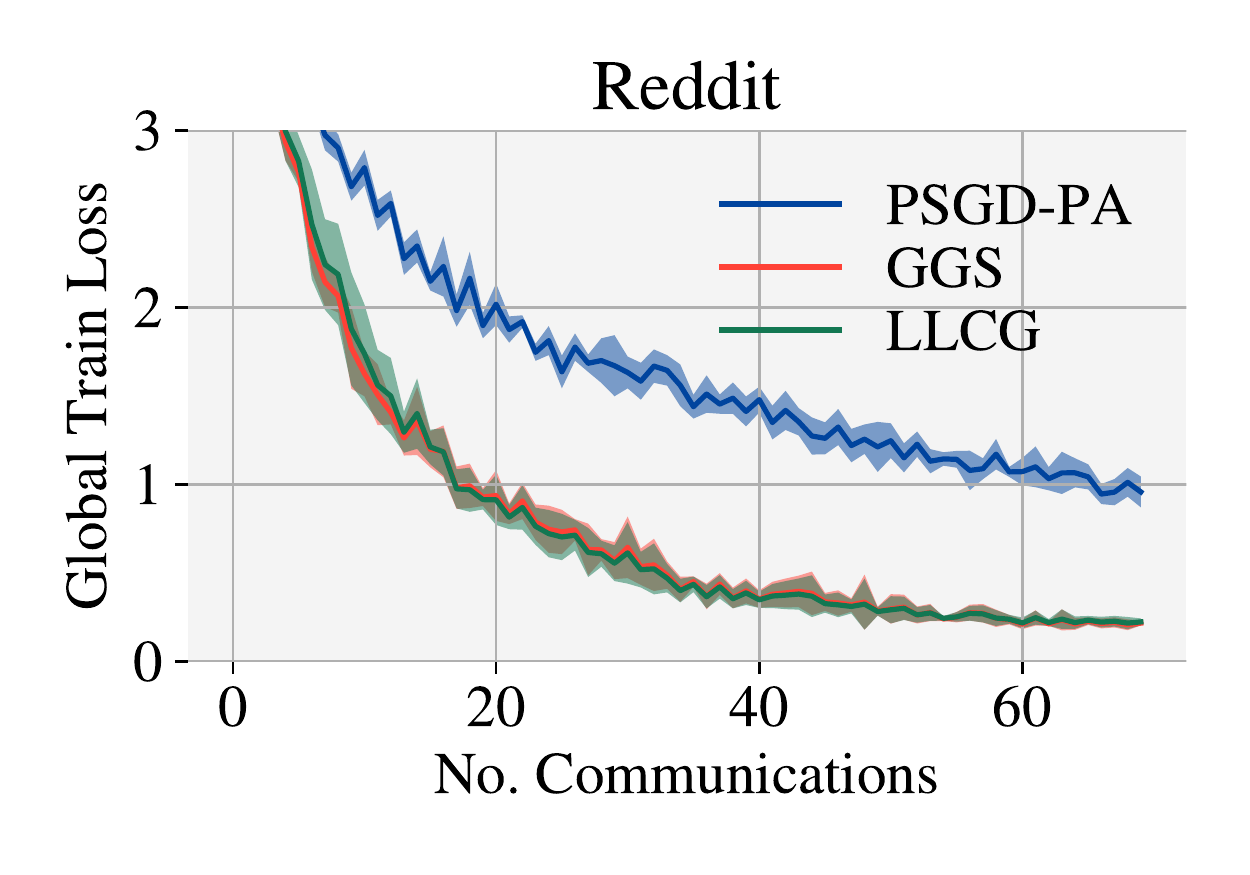}
        \vspace{-22pt}\caption{}\vspace{-3pt}
        \label{subfig:loss-reddit}
    \end{subfigure}
    \hfill
    \begin{subfigure}[t]{0.24\textwidth}
        \centering
        \includegraphics[clip, trim=15 0 15 10, width=1\linewidth]{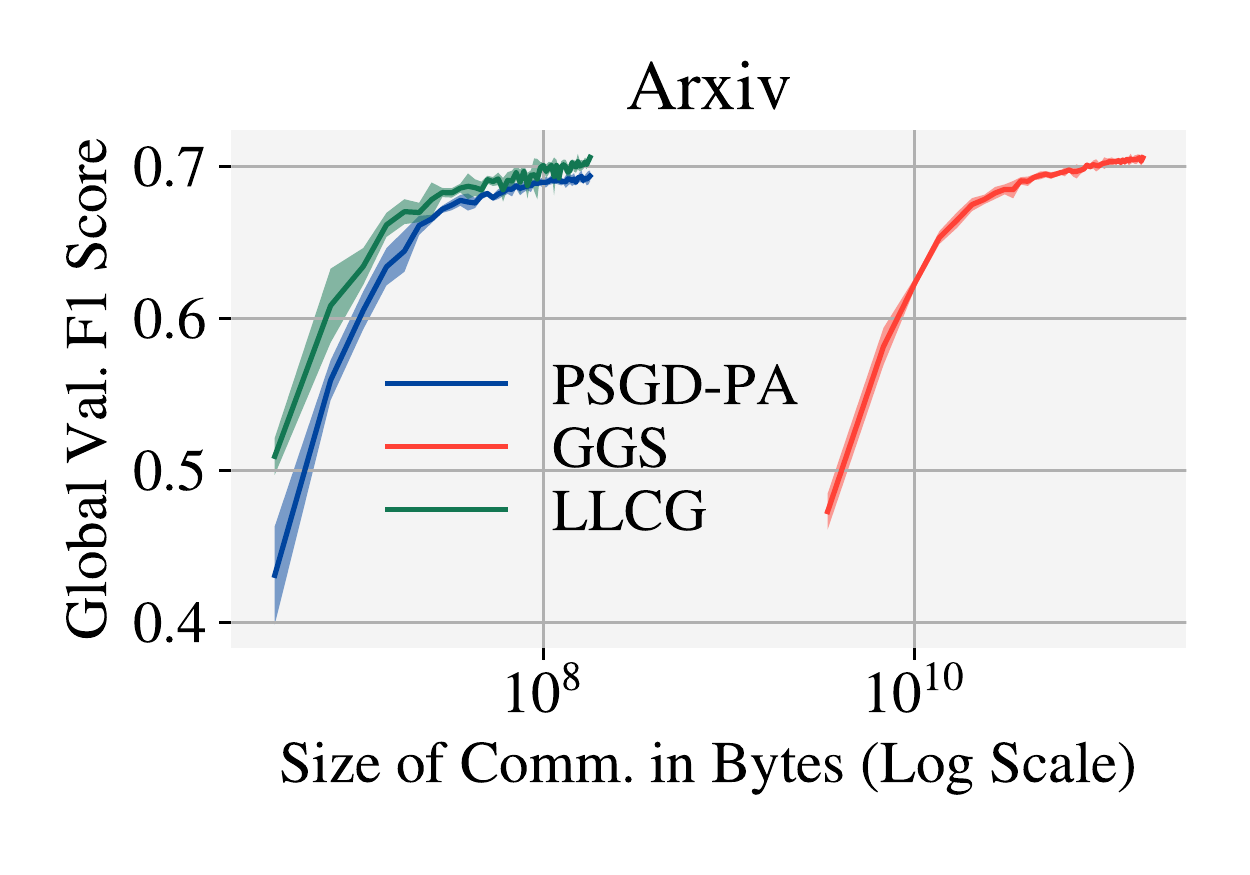}
        \vspace{-22pt}\caption{}\vspace{-3pt}
        \label{subfig:size-arxiv}
    \end{subfigure}
    \hfill
    \begin{subfigure}[t]{0.24\textwidth}
        \centering
        \includegraphics[clip, trim=20 0 20 10, width=1\linewidth]{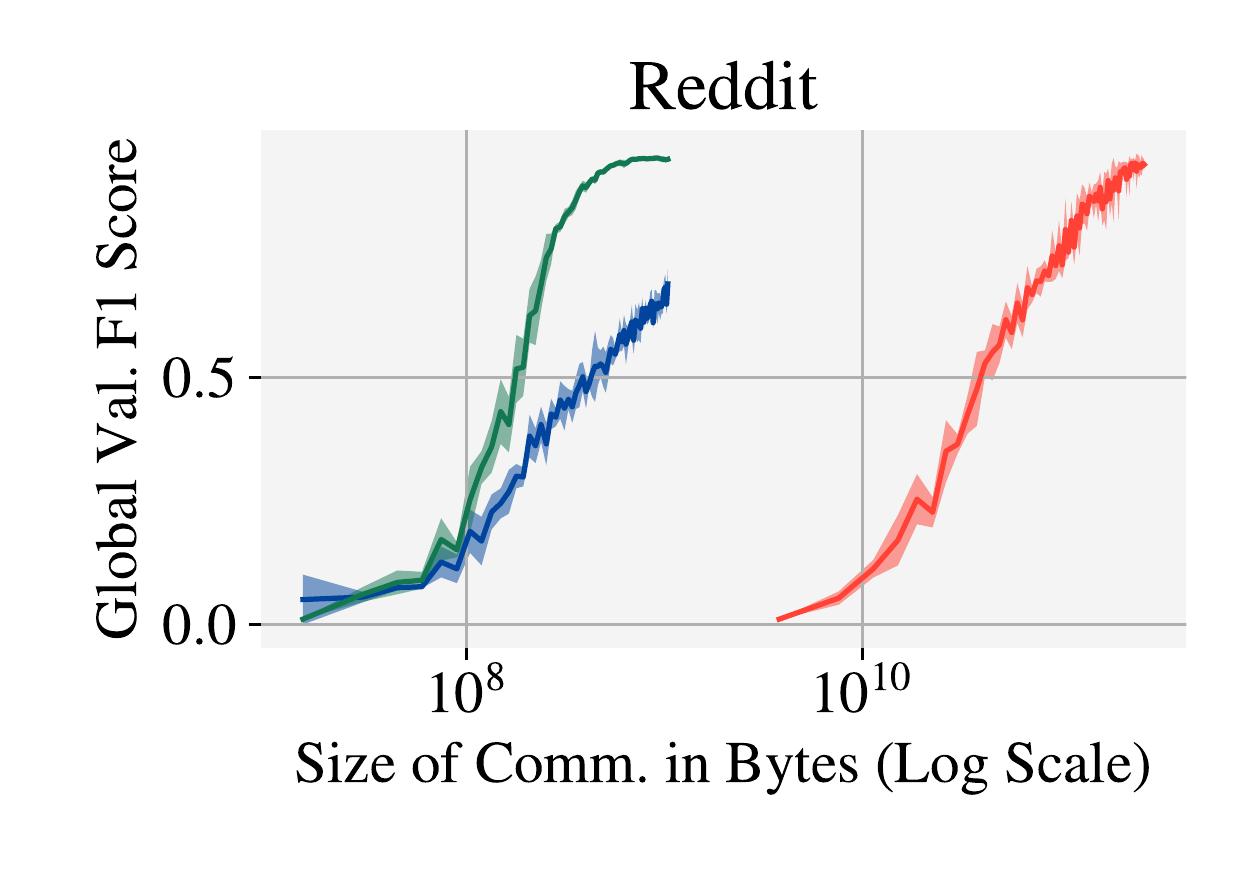}
        \vspace{-22pt}\caption{}\vspace{-3pt}
        \label{subfig:size-reddit}
    \end{subfigure}
    \vspace{-5pt}
    \caption{
    Comparing \CG{} against PSGD-PA and GGS on real-world datasets.
    We show the global validation score in terms of the number of communications in (\textit{a,b,c,d}), the training loss per round of communications in (\textit{e,f}), and the global validation score per bytes of exchanged data in (\textit{g,h}). 
    }
    \vspace{-2pt}
    \label{figure:main-results}
\end{figure}
\setlength{\textfloatsep}{1pt}
\begin{table}[t]
\centering
\vspace{-5pt}
\caption{Comparison of performance and the average Megabytes of node representation/feature communicated per round on various datasets. 
}
\label{table:new-results}
\vspace{-10pt}
\resizebox{0.925\linewidth}{!}{
\begin{tabular}{llcrlrlrl} \toprule
                            & Method  &\multirow{2}{*}{\shortstack{No.\\ Comm.}} & \multicolumn{2}{c}{GCN~/~SAGE}   & \multicolumn{2}{c}{GAT} & \multicolumn{2}{c}{APPNP} \\
                            &         &                       & Performance         & Avg. MB   & Performance & Avg. MB           & Performance  & Avg. MB           \\ \midrule\midrule
\multirow{3}{*}{\begin{tabular}[c]{@{}l@{}} \texttt{Flickr}\\ (F1-score) \end{tabular}}       
                            & PSGD-PA & \multirow{3}{*}{50}   & $49.08{\scriptstyle\pm0.27}$   & $12.57$   & $51.56{\scriptstyle\pm0.28}$  & $4.24$     & $50.81{\scriptstyle\pm0.48}$   & $8.40$     \\
                            & GGS     &                       & $51.22{\scriptstyle\pm0.13}$   & $1849.32$ & $52.41{\scriptstyle\pm0.29}$  & $1895.61$  & $51.33{\scriptstyle\pm0.33}$   & $1897.82$  \\
                            & \textbf{\CG{}}   &              & $50.38{\scriptstyle\pm0.20}$   & $12.57$   & $52.01{\scriptstyle\pm0.33}$  & $4.24$     & $51.15{\scriptstyle\pm0.25}$   & $8.40$     \\ \midrule
\multirow{3}{*}{\begin{tabular}[c]{@{}l@{}} \texttt{OGB-Proteins}\\ (ROC-AUC) \end{tabular}} 
                            & PSGD-PA & \multirow{3}{*}{100}  & $72.85{\scriptstyle\pm0.70}$   & $6.20$    & $64.95{\scriptstyle\pm1.01}$  & $3.14$     & $71.10{\scriptstyle\pm0.79}$   & $7.31$      \\
                            & GGS     &                       & $74.78{\scriptstyle\pm0.36}$   & $922.42$  & $68.11{\scriptstyle\pm0.60}$  & $912.79$   & $71.29{\scriptstyle\pm0.31}$   & $917.20$    \\
                            & \textbf{\CG{}}   &              & $73.92{\scriptstyle\pm0.45}$   & $6.20$    & $67.62{\scriptstyle\pm0.58}$  & $3.14$     & $71.18{\scriptstyle\pm0.43}$   & $7.31$      \\ \midrule
\multirow{3}{*}{\begin{tabular}[c]{@{}l@{}} \texttt{OGB-Arxiv}\\ (F1-score) \end{tabular}}    
                            & PSGD-PA & \multirow{3}{*}{100}  & $69.43{\scriptstyle\pm0.21}$   & $3.55$    & $69.88{\scriptstyle\pm0.18}$  & $3.59$     & $68.48{\scriptstyle\pm0.17}$   & $7.71$      \\
                            & GGS     &                       & $70.51{\scriptstyle\pm0.26}$   & $3391.03$ & $70.82{\scriptstyle\pm0.23}$  & $3396.79$  & $69.01{\scriptstyle\pm0.10}$   & $3394.33$   \\
                            & \textbf{\CG{}}   &              & $70.21{\scriptstyle\pm0.13}$   & $3.55$    & $70.58{\scriptstyle\pm0.37}$  & $3.59$     & $68.73{\scriptstyle\pm0.29}$   & $7.71$      \\ \midrule
\multirow{3}{*}{\begin{tabular}[c]{@{}l@{}} \texttt{Reddit}\\ (F1-score) \end{tabular}}    
                            & PSGD-PA & \multirow{3}{*}{75}   & $71.17{\scriptstyle\pm1.06}$   & $14.83$   & $70.57{\scriptstyle\pm1.24}$  & $7.48$     & $83.48{\scriptstyle\pm0.81}$   & $11.63$     \\
                            & GGS     &                       & $94.77{\scriptstyle\pm0.20}$   & $3798.81$ & $95.03{\scriptstyle\pm0.48}$  & $3805.28$  & $95.23{\scriptstyle\pm0.22}$   & $3770.46$   \\ 
                            & \textbf{\CG{}}   &              & $94.67{\scriptstyle\pm0.15}$   & $14.83$   & $94.73{\scriptstyle\pm0.23}$  & $7.48$     & $94.64{\scriptstyle\pm0.17}$   & $11.63$     \\
\bottomrule
\end{tabular}
}
\end{table}

\begin{figure}[t]
    \vspace{-15pt}
    \hfill
    \centering
    \begin{minipage}{0.28\textwidth}
        \centering
        \includegraphics[trim=20 10 20 0, width=1\linewidth]{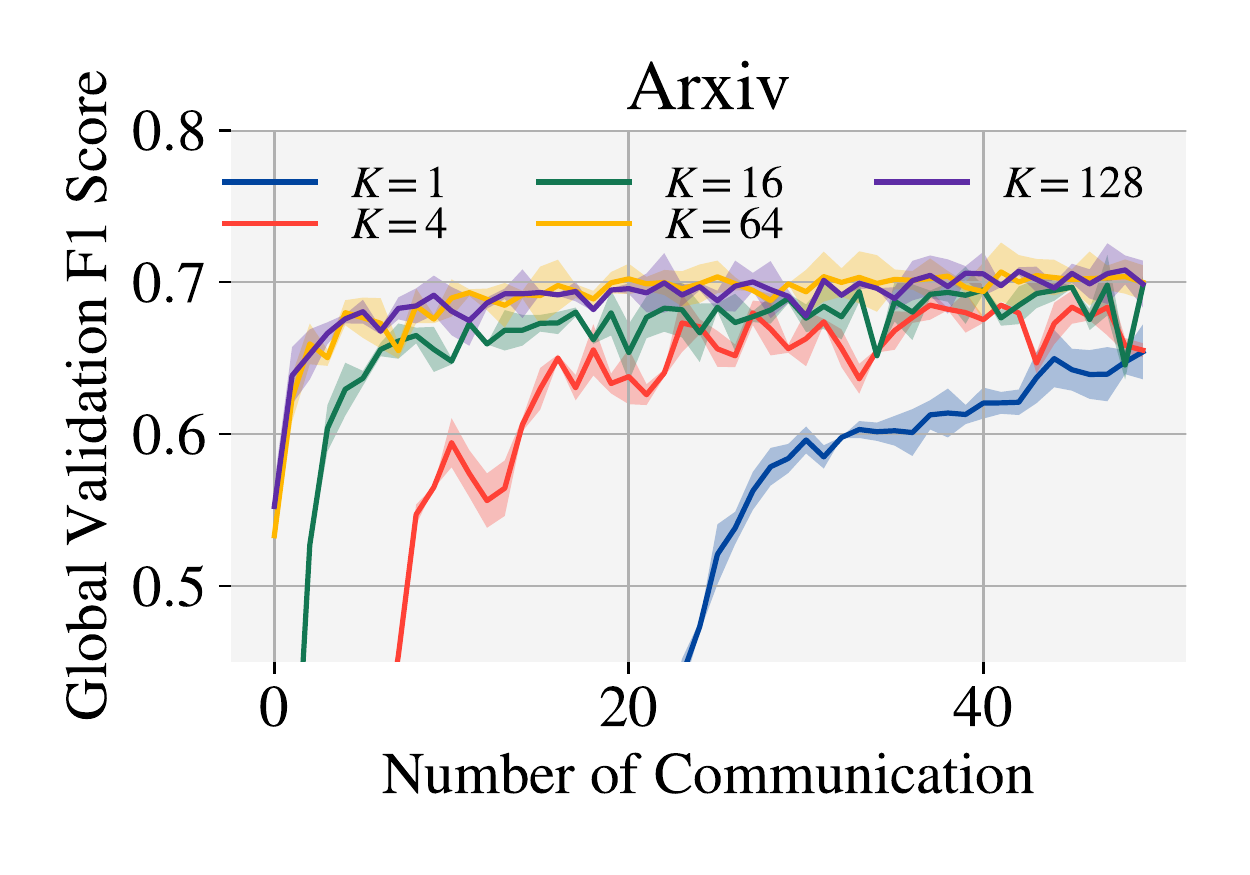}
        \vspace{-20pt}
        \caption{Effect of local epoch size ($K$)}
        \label{figure:k-arxiv}
    \end{minipage}
    \hfill
    \begin{minipage}{0.56\textwidth}
        \centering
        \begin{subfigure}[t]{0.49\linewidth}
            \centering
            \includegraphics[trim=20 10 20 0, width=1\linewidth]{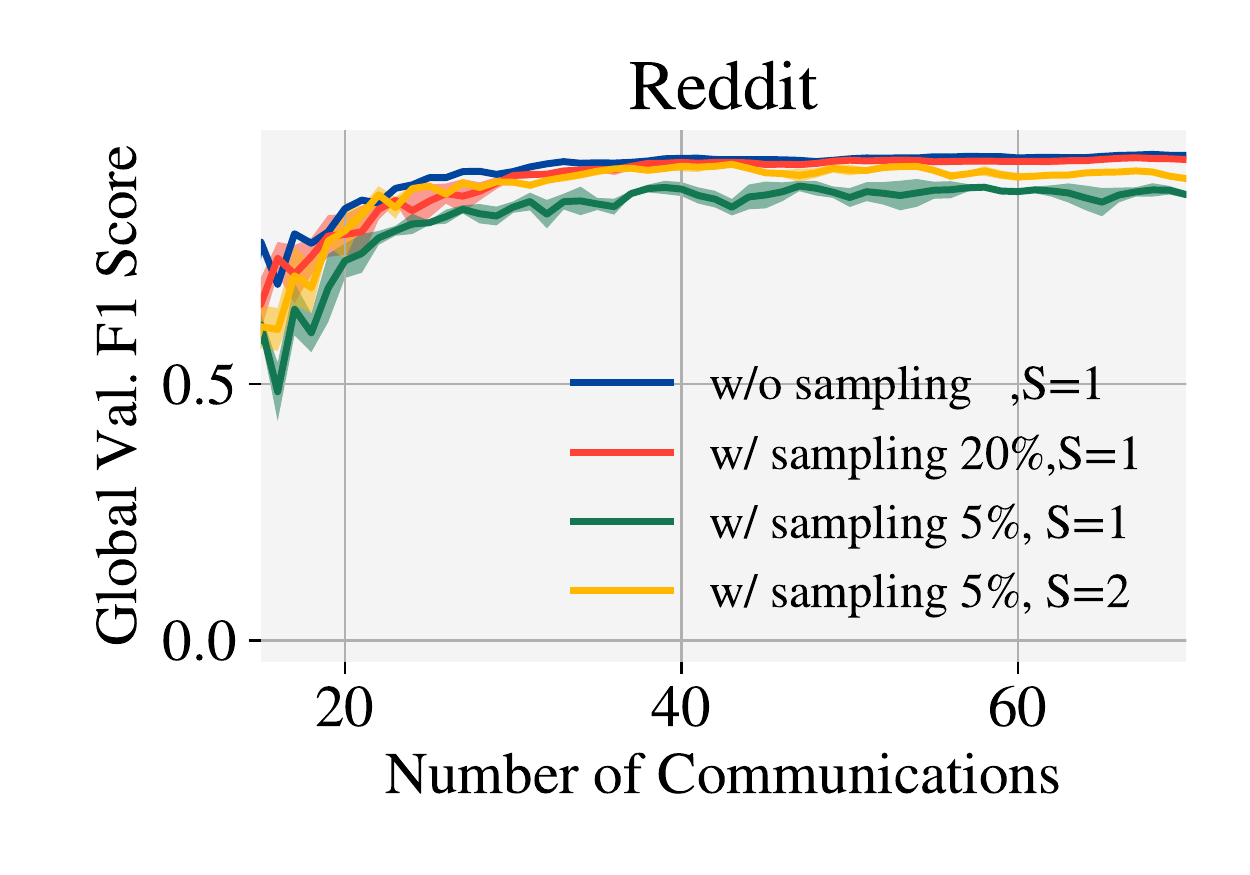}
            \label{figure:local-sampling-reddit}
        \end{subfigure}
        \hfill
        \begin{subfigure}[t]{0.49\linewidth}
            \centering
            \includegraphics[trim=20 10 20 0, width=1\linewidth]{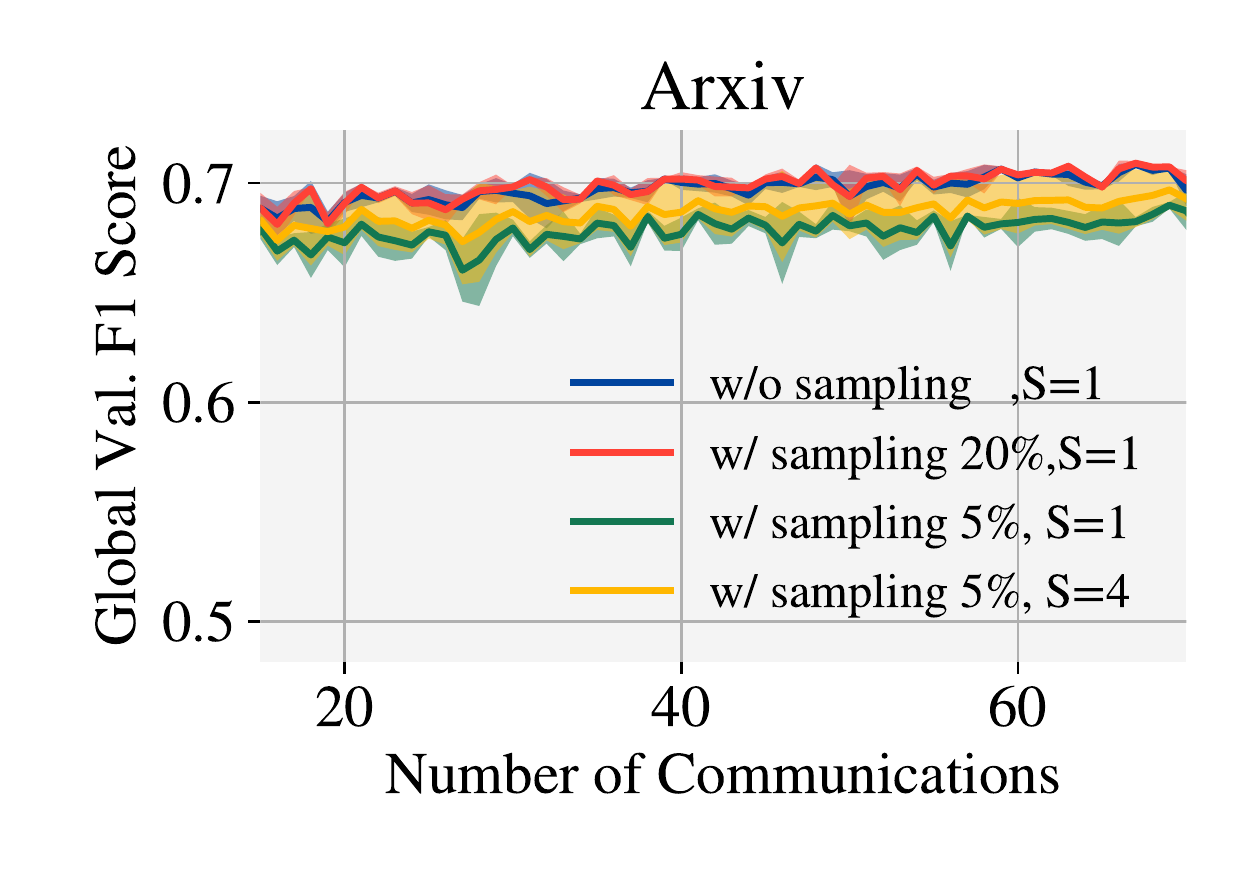}
            \label{figure:local-sampling-arxiv}
        \end{subfigure}
        \vspace{-20pt}
        \caption{Effect of sampling on local machine and number of correction steps on the server}
        \label{figure:local-sampling}
    \end{minipage}
    \hfill
    \vspace{5pt}
\end{figure}

\textbf{LLCG requires same number of communications.}~
Figure~\ref{subfig:acc-flickr} through \ref{subfig:acc-reddit} illustrate the validation accuracy per communication rounds on four different datasets. 
We run a fixed number of communication rounds and plot the global validation score (the validation score computed using the full-graph on the server) at the end of each communication step. 
For PSGD-PA and GGS, the score is calculated on the averaged model, whereas for \CG{} the validation is calculated after the correction step. 
It can be seen that PSGD-PA suffers from performance drop compared to other two methods, due to the residual error we discussed in Section~\ref{section:theory}, while both GGS and \CG{} perform well and can achieve the expected accuracy. 
Note that the performance drop of PSGD-PA can vary across different datasets; 
in some cases such as \texttt{Reddit}, PSGD-PA can significantly hurt the accuracy, while on other datasets the gap is smaller. 
Nevertheless, \CG{} can always close the gap between PSGD-PA and GGS with minimal overhead.
\vspace{0.8mm}\\
\noindent\textbf{LLCG convergences as fast as GGS.}~
To represent the effect of communication on the real-time convergence, in Figure~\ref{subfig:loss-arxiv} and ~\ref{subfig:loss-reddit}, we plot the global training loss (training loss computed on the full-graph on the server) after each communication round.
Similar to the accuracy score, the training loss is also computed on the server averaged (and corrected, in case of LLCG) global model. 
These results clearly indicate that \CG{} can improve the convergence over PSGD-PA, while it shows a similar convergence speed to GGS. \vspace{0.8mm}\\
\noindent\textbf{LLCG exchanges data as little as PSGD-PA.}~
Figure~\ref{subfig:size-arxiv} and ~\ref{subfig:size-reddit} show the relation between global validation accuracy with the average size (volume) of communication in bytes. 
As expected, this figure clearly shows the effectiveness of \CG{}.
On the one hand, \CG{} has a similar amount of communication volume as PSGD-PA but can achieve a higher accuracy. On the other hand, \CG{} requires significantly less amount of communication volume than GGS to achieve the same accuracy, which leads to slower training time in real world settings. \vspace{0.8mm} \\
\noindent\textbf{LLCG works with various GNN models and aggregations.}~
We evaluate four popular GNN models, used in recent graph learning literature: GCN~\cite{kipf2016semi}, SAGE~\cite{hamilton2017inductive}, GAT~\cite{velivckovic2017graph} and APPNP~\cite{klicpera2018predict}.
In Table~\ref{table:new-results}, we summarize the test score and average communication size (in MB) on different datasets for a fixed number of communication rounds.
Note that we only include the results for the aggregation methods (GCN or SAGE) that have higher accuracy for the specific datasets, details of which can be found in Appendix~\ref{section:dataset_and_model_details}.
As shown here, \CG{} can consistently improve the test accuracy for all different models compared to PSGD-PA, while the communication size is significantly lower than GGS, since \CG{} only needs to exchange the model parameters. \vspace{0.8mm}\\
\noindent\textbf{Effect of local epoch size.}~
Figure~\ref{figure:k-arxiv} compares the effect of various values of local epoch size $K \in \{1,4,16,64,128\}$ for fixed $\rho$ and $S$ on the \texttt{OGB-Arxiv} dataset. 
When using {\em fully synchronous} with $K=1$, the model suffers from very slow convergence and needs more communications. 
Further increasing the $K$ to larger values can speed up the training; however, we found a diminishing return point for $K>128$ in this dataset and extremely large $K$ in general. \vspace{0.8mm}\\
\noindent\textbf{Effect of sampling in local machines.~}
In Figure~\ref{figure:local-sampling}, we report the validation scores per round of communication to compare the effect of neighborhood sampling at local machines.
We can observe that when the neighborhood sampling size is reasonably large (i.e., $20\%$), the performance is very similar to full neighborhood training.
However, reducing the neighbor sampling ratio to $5\%$ could result in a larger local stochastic gradient bias $\sigma^2_\text{bias}$, which requires using more correction steps ($S$).

\vspace{-5pt}
\section{Concluding Remarks}
\label{section:conclusion}

In this paper, we propose a novel distributed algorithm for training Graph Neural Networks (GNNs). 
We theoretically analyze various GNN models and discover that, unlike the traditional deep neural networks, due to inherent data samples dependency in GNNs, naively applying periodic parameter averaging leads to a residual error and current solutions to this issue impose huge communication overheads. 
Instead, our proposal tackles these problems by applying correction on top of locally learned models, to infuse the global structure of the graph back into the network and avoid any costly communication. In addition, through extensive empirical analysis, we support our theoretical findings and demonstrate that \CG{} can achieve high accuracy without additional communication costs.


\section*{Acknowledgements}
This work was supported in part by CRISP, one of six centers in JUMP, a Semiconductor Research
Corporation (SRC) program sponsored by DARPA and NSF grants 1909004, 1714389, 1912495,
1629915, 1629129, 1763681, 2008398.

\section*{Reproducibility Statement}
We provide a GitHub repository in Appendix~\ref{appendix:experiments} including all code and scripts used in our experimental studies.
This repository includes a \texttt{README.md} file, explaining how to install and prepare the code and required packages.
Detailed instruction on how to use the partitioning scripts is provided for various datasets.
In addition, we provide several configuration files (under \texttt{scripts/configs}) folder for different hyper-parameters on each individual dataset, and a general script (\texttt{scripts/run-config.py}) to run and reproduce the results with these configurations.
Details of various models and parameters used in our evaluation studies can also be found in Appendix~\ref{appendix:experiments}.

\bibliography{references}

\begin{thebibliography}{43}
\providecommand{\natexlab}[1]{#1}
\providecommand{\url}[1]{\texttt{#1}}
\expandafter\ifx\csname urlstyle\endcsname\relax
  \providecommand{\doi}[1]{doi: #1}\else
  \providecommand{\doi}{doi: \begingroup \urlstyle{rm}\Url}\fi

\bibitem[Angerd et~al.(2020)Angerd, Balasubramanian, and
  Annavaram]{angerd2020distributed}
Alexandra Angerd, Keshav Balasubramanian, and Murali Annavaram.
\newblock Distributed training of graph convolutional networks using subgraph
  approximation.
\newblock \emph{arXiv preprint arXiv:2012.04930}, 2020.

\bibitem[Boldi \& Vigna(2004)Boldi and Vigna]{boldi2004webgraph}
Paolo Boldi and Sebastiano Vigna.
\newblock The webgraph framework {I:} compression techniques.
\newblock In \emph{Proceedings of the 13th international conference on World
  Wide Web, {WWW} 2004, New York, NY, USA, May 17-20, 2004}, pp.\  595--602.
  {ACM}, 2004.
\newblock \doi{10.1145/988672.988752}.

\bibitem[Bonawitz et~al.(2019)Bonawitz, Eichner, Grieskamp, Huba, Ingerman,
  Ivanov, Kiddon, Konecn{\'{y}}, Mazzocchi, McMahan, Overveldt, Petrou, Ramage,
  and Roselander]{bonawitztowards}
Keith Bonawitz, Hubert Eichner, Wolfgang Grieskamp, Dzmitry Huba, Alex
  Ingerman, Vladimir Ivanov, Chlo{\'{e}} Kiddon, Jakub Konecn{\'{y}}, Stefano
  Mazzocchi, Brendan McMahan, Timon~Van Overveldt, David Petrou, Daniel Ramage,
  and Jason Roselander.
\newblock Towards federated learning at scale: System design.
\newblock In \emph{Proceedings of Machine Learning and Systems 2019, MLSys
  2019, Stanford, CA, USA, March 31 - April 2, 2019}. mlsys.org, 2019.

\bibitem[Chen et~al.(2018)Chen, Zhu, and Song]{chen2017stochastic}
Jianfei Chen, Jun Zhu, and Le~Song.
\newblock Stochastic training of graph convolutional networks with variance
  reduction.
\newblock In \emph{Proceedings of the 35th International Conference on Machine
  Learning, {ICML} 2018, Stockholmsm{\"{a}}ssan, Stockholm, Sweden, July 10-15,
  2018}, volume~80 of \emph{Proceedings of Machine Learning Research}, pp.\
  941--949. {PMLR}, 2018.

\bibitem[Chiang et~al.(2019)Chiang, Liu, Si, Li, Bengio, and
  Hsieh]{chiang2019cluster}
Wei{-}Lin Chiang, Xuanqing Liu, Si~Si, Yang Li, Samy Bengio, and Cho{-}Jui
  Hsieh.
\newblock Cluster-gcn: An efficient algorithm for training deep and large graph
  convolutional networks.
\newblock In \emph{Proceedings of the 25th {ACM} {SIGKDD} International
  Conference on Knowledge Discovery {\&} Data Mining, {KDD} 2019, Anchorage,
  AK, USA, August 4-8, 2019}, pp.\  257--266. {ACM}, 2019.
\newblock \doi{10.1145/3292500.3330925}.

\bibitem[Cong et~al.(2020)Cong, Forsati, Kandemir, and
  Mahdavi]{cong2020minimal}
Weilin Cong, Rana Forsati, Mahmut~T. Kandemir, and Mehrdad Mahdavi.
\newblock Minimal variance sampling with provable guarantees for fast training
  of graph neural networks.
\newblock In \emph{{KDD} '20: The 26th {ACM} {SIGKDD} Conference on Knowledge
  Discovery and Data Mining, Virtual Event, CA, USA, August 23-27, 2020}, pp.\
  1393--1403. {ACM}, 2020.

\bibitem[Cong et~al.(2021)Cong, Ramezani, and Mahdavi]{cong2021importance}
Weilin Cong, Morteza Ramezani, and Mehrdad Mahdavi.
\newblock On the importance of sampling in learning graph convolutional
  networks.
\newblock \emph{arXiv preprint arXiv:2103.02696}, 2021.

\bibitem[Dean et~al.(2012)Dean, Corrado, Monga, Chen, Devin, Le, Mao, Ranzato,
  Senior, Tucker, Yang, and Ng]{dean2012large}
Jeffrey Dean, Greg Corrado, Rajat Monga, Kai Chen, Matthieu Devin, Quoc~V. Le,
  Mark~Z. Mao, Marc'Aurelio Ranzato, Andrew~W. Senior, Paul~A. Tucker, Ke~Yang,
  and Andrew~Y. Ng.
\newblock Large scale distributed deep networks.
\newblock In \emph{Advances in Neural Information Processing Systems 25: 26th
  Annual Conference on Neural Information Processing Systems 2012. Proceedings
  of a meeting held December 3-6, 2012, Lake Tahoe, Nevada, United States},
  pp.\  1232--1240, 2012.

\bibitem[Deng et~al.(2019)Deng, Rangwala, and Ning]{deng2019learning}
Songgaojun Deng, Huzefa Rangwala, and Yue Ning.
\newblock Learning dynamic context graphs for predicting social events.
\newblock In \emph{Proceedings of the 25th {ACM} {SIGKDD} International
  Conference on Knowledge Discovery {\&} Data Mining, {KDD} 2019, Anchorage,
  AK, USA, August 4-8, 2019}, pp.\  1007--1016. {ACM}, 2019.
\newblock \doi{10.1145/3292500.3330919}.

\bibitem[Do et~al.(2019)Do, Tran, and Venkatesh]{do2019graph}
Kien Do, Truyen Tran, and Svetha Venkatesh.
\newblock Graph transformation policy network for chemical reaction prediction.
\newblock In \emph{Proceedings of the 25th {ACM} {SIGKDD} International
  Conference on Knowledge Discovery {\&} Data Mining, {KDD} 2019, Anchorage,
  AK, USA, August 4-8, 2019}, pp.\  750--760. {ACM}, 2019.
\newblock \doi{10.1145/3292500.3330958}.

\bibitem[Faez et~al.(2021)Faez, Ommi, Baghshah, and Rabiee]{faez2021deep}
Faezeh Faez, Yassaman Ommi, Mahdieh~Soleymani Baghshah, and Hamid~R Rabiee.
\newblock Deep graph generators: A survey.
\newblock \emph{IEEE Access}, 9:\penalty0 106675--106702, 2021.

\bibitem[Fout et~al.(2017)Fout, Byrd, Shariat, and Ben{-}Hur]{fout2017protein}
Alex Fout, Jonathon Byrd, Basir Shariat, and Asa Ben{-}Hur.
\newblock Protein interface prediction using graph convolutional networks.
\newblock In \emph{Advances in Neural Information Processing Systems 30: Annual
  Conference on Neural Information Processing Systems 2017, December 4-9, 2017,
  Long Beach, CA, {USA}}, pp.\  6530--6539, 2017.

\bibitem[Ghorbani et~al.(2022)Ghorbani, Kazi, Baghshah, Rabiee, and
  Navab]{ghorbani2022ra}
Mahsa Ghorbani, Anees Kazi, Mahdieh~Soleymani Baghshah, Hamid~R Rabiee, and
  Nassir Navab.
\newblock Ra-gcn: Graph convolutional network for disease prediction problems
  with imbalanced data.
\newblock \emph{Medical Image Analysis}, 75:\penalty0 102272, 2022.

\bibitem[Hamilton et~al.(2017)Hamilton, Ying, and
  Leskovec]{hamilton2017inductive}
William~L. Hamilton, Zhitao Ying, and Jure Leskovec.
\newblock Inductive representation learning on large graphs.
\newblock In \emph{Advances in Neural Information Processing Systems 30: Annual
  Conference on Neural Information Processing Systems 2017, December 4-9, 2017,
  Long Beach, CA, {USA}}, pp.\  1024--1034, 2017.

\bibitem[Hard et~al.(2018)Hard, Rao, Mathews, Ramaswamy, Beaufays, Augenstein,
  Eichner, Kiddon, and Ramage]{hard2018federated}
Andrew Hard, Kanishka Rao, Rajiv Mathews, Swaroop Ramaswamy, Fran{\c{c}}oise
  Beaufays, Sean Augenstein, Hubert Eichner, Chlo{\'e} Kiddon, and Daniel
  Ramage.
\newblock Federated learning for mobile keyboard prediction.
\newblock \emph{arXiv preprint arXiv:1811.03604}, 2018.

\bibitem[Hu et~al.(2020)Hu, Fey, Zitnik, Dong, Ren, Liu, Catasta, and
  Leskovec]{hu2020open}
Weihua Hu, Matthias Fey, Marinka Zitnik, Yuxiao Dong, Hongyu Ren, Bowen Liu,
  Michele Catasta, and Jure Leskovec.
\newblock Open graph benchmark: Datasets for machine learning on graphs.
\newblock In \emph{Advances in Neural Information Processing Systems 33: Annual
  Conference on Neural Information Processing Systems 2020, NeurIPS 2020,
  December 6-12, 2020, virtual}, 2020.

\bibitem[Hu et~al.(2021)Hu, Fey, Ren, Nakata, Dong, and Leskovec]{hu2021ogb}
Weihua Hu, Matthias Fey, Hongyu Ren, Maho Nakata, Yuxiao Dong, and Jure
  Leskovec.
\newblock Ogb-lsc: A large-scale challenge for machine learning on graphs.
\newblock \emph{arXiv preprint arXiv:2103.09430}, 2021.

\bibitem[Ioffe \& Szegedy(2015)Ioffe and Szegedy]{ioffe2015batch}
Sergey Ioffe and Christian Szegedy.
\newblock Batch normalization: Accelerating deep network training by reducing
  internal covariate shift.
\newblock In \emph{Proceedings of the 32nd International Conference on Machine
  Learning, {ICML} 2015, Lille, France, 6-11 July 2015}, volume~37 of
  \emph{{JMLR} Workshop and Conference Proceedings}, pp.\  448--456. JMLR.org,
  2015.

\bibitem[Jiang \& Rumi(2021)Jiang and Rumi]{jiang2021communication}
Peng Jiang and Masuma~Akter Rumi.
\newblock Communication-efficient sampling for distributed training of graph
  convolutional networks.
\newblock \emph{arXiv preprint arXiv:2101.07706}, 2021.

\bibitem[Kipf \& Welling(2017)Kipf and Welling]{kipf2016semi}
Thomas~N. Kipf and Max Welling.
\newblock Semi-supervised classification with graph convolutional networks.
\newblock In \emph{5th International Conference on Learning Representations,
  {ICLR} 2017, Toulon, France, April 24-26, 2017, Conference Track
  Proceedings}. OpenReview.net, 2017.

\bibitem[Klicpera et~al.(2019)Klicpera, Bojchevski, and
  G{\"{u}}nnemann]{klicpera2018predict}
Johannes Klicpera, Aleksandar Bojchevski, and Stephan G{\"{u}}nnemann.
\newblock Predict then propagate: Graph neural networks meet personalized
  pagerank.
\newblock In \emph{7th International Conference on Learning Representations,
  {ICLR} 2019, New Orleans, LA, USA, May 6-9, 2019}. OpenReview.net, 2019.

\bibitem[Kone{\v{c}}n{\`y} et~al.(2018)Kone{\v{c}}n{\`y}, McMahan, Felix,
  Suresh, Bacon, and Richt{\'a}rik]{konevcny2018federated}
Jakub Kone{\v{c}}n{\`y}, H~Brendan McMahan, X~Yu Felix, Ananda~Theertha Suresh,
  Dave Bacon, and Peter Richt{\'a}rik.
\newblock Federated learning: Strategies for improving communication
  efficiency.
\newblock 2018.

\bibitem[Li et~al.(2019)Li, M{\"{u}}ller, Thabet, and Ghanem]{li2019deepgcns}
Guohao Li, Matthias M{\"{u}}ller, Ali~K. Thabet, and Bernard Ghanem.
\newblock Deepgcns: Can gcns go as deep as cnns?
\newblock In \emph{2019 {IEEE/CVF} International Conference on Computer Vision,
  {ICCV} 2019, Seoul, Korea (South), October 27 - November 2, 2019}, pp.\
  9266--9275. {IEEE}, 2019.
\newblock \doi{10.1109/ICCV.2019.00936}.

\bibitem[Li et~al.(2020{\natexlab{a}})Li, Zhao, Varma, Salpekar, Noordhuis, Li,
  Paszke, Smith, Vaughan, Damania, et~al.]{li2020pytorch}
Shen Li, Yanli Zhao, Rohan Varma, Omkar Salpekar, Pieter Noordhuis, Teng Li,
  Adam Paszke, Jeff Smith, Brian Vaughan, Pritam Damania, et~al.
\newblock Pytorch distributed: Experiences on accelerating data parallel
  training.
\newblock \emph{arXiv preprint arXiv:2006.15704}, 2020{\natexlab{a}}.

\bibitem[Li et~al.(2020{\natexlab{b}})Li, Huang, Yang, Wang, and
  Zhang]{li2019convergence}
Xiang Li, Kaixuan Huang, Wenhao Yang, Shusen Wang, and Zhihua Zhang.
\newblock On the convergence of fedavg on non-iid data.
\newblock In \emph{8th International Conference on Learning Representations,
  {ICLR} 2020, Addis Ababa, Ethiopia, April 26-30, 2020}. OpenReview.net,
  2020{\natexlab{b}}.

\bibitem[Lin et~al.(2021)Lin, He, Zeng, Wang, Huang, Soltanolkotabi, Ren, and
  Avestimehr]{lin2021fednlp}
Bill~Yuchen Lin, Chaoyang He, Zihang Zeng, Hulin Wang, Yufen Huang, Mahdi
  Soltanolkotabi, Xiang Ren, and Salman Avestimehr.
\newblock Fednlp: A research platform for federated learning in natural
  language processing.
\newblock \emph{arXiv preprint arXiv:2104.08815}, 2021.

\bibitem[Md et~al.(2021)Md, Misra, Ma, Mohanty, Georganas, Heinecke, Kalamkar,
  Ahmed, and Avancha]{md2021distgnn}
Vasimuddin Md, Sanchit Misra, Guixiang Ma, Ramanarayan Mohanty, Evangelos
  Georganas, Alexander Heinecke, Dhiraj Kalamkar, Nesreen~K Ahmed, and
  Sasikanth Avancha.
\newblock Distgnn: Scalable distributed training for large-scale graph neural
  networks.
\newblock \emph{arXiv preprint arXiv:2104.06700}, 2021.

\bibitem[Ramezani et~al.(2020)Ramezani, Cong, Mahdavi, Sivasubramaniam, and
  Kandemir]{ramezani2020gcn}
Morteza Ramezani, Weilin Cong, Mehrdad Mahdavi, Anand Sivasubramaniam, and
  Mahmut Kandemir.
\newblock Gcn meets gpu: Decoupling “when to sample” from “how to
  sample”.
\newblock \emph{Advances in Neural Information Processing Systems}, 33, 2020.

\bibitem[Scardapane et~al.(2020)Scardapane, Spinelli, and
  Di~Lorenzo]{scardapane2020distributed}
Simone Scardapane, Indro Spinelli, and Paolo Di~Lorenzo.
\newblock Distributed graph convolutional networks.
\newblock \emph{arXiv preprint arXiv:2007.06281}, 2020.

\bibitem[Shin et~al.(2018)Shin, Kim, Shin, and Xiao]{shin2018privacy}
Hyejin Shin, Sungwook Kim, Junbum Shin, and Xiaokui Xiao.
\newblock Privacy enhanced matrix factorization for recommendation with local
  differential privacy.
\newblock \emph{IEEE Transactions on Knowledge and Data Engineering},
  30\penalty0 (9):\penalty0 1770--1782, 2018.

\bibitem[Stich(2019)]{stich2018local}
Sebastian~U. Stich.
\newblock Local {SGD} converges fast and communicates little.
\newblock In \emph{7th International Conference on Learning Representations,
  {ICLR} 2019, New Orleans, LA, USA, May 6-9, 2019}. OpenReview.net, 2019.

\bibitem[Sun \& Wu(2020)Sun and Wu]{sun2020adaptive}
Chuxiong Sun and Guoshi Wu.
\newblock Adaptive graph diffusion networks with hop-wise attention.
\newblock \emph{arXiv preprint arXiv:2012.15024}, 2020.

\bibitem[Sun \& Wu(2021)Sun and Wu]{sun2021scalable}
Chuxiong Sun and Guoshi Wu.
\newblock Scalable and adaptive graph neural networks with self-label-enhanced
  training.
\newblock \emph{arXiv preprint arXiv:2104.09376}, 2021.

\bibitem[Tripathy et~al.(2020)Tripathy, Yelick, and
  Buluc]{tripathy2020reducing}
Alok Tripathy, Katherine Yelick, and Aydin Buluc.
\newblock Reducing communication in graph neural network training.
\newblock \emph{arXiv preprint arXiv:2005.03300}, 2020.

\bibitem[Velickovic et~al.(2018)Velickovic, Cucurull, Casanova, Romero,
  Li{\`{o}}, and Bengio]{velivckovic2017graph}
Petar Velickovic, Guillem Cucurull, Arantxa Casanova, Adriana Romero, Pietro
  Li{\`{o}}, and Yoshua Bengio.
\newblock Graph attention networks.
\newblock In \emph{6th International Conference on Learning Representations,
  {ICLR} 2018, Vancouver, BC, Canada, April 30 - May 3, 2018, Conference Track
  Proceedings}. OpenReview.net, 2018.

\bibitem[Wang et~al.(2018)Wang, Huang, Zhao, Zhang, Zhao, and
  Lee]{wang2018billion}
Jizhe Wang, Pipei Huang, Huan Zhao, Zhibo Zhang, Binqiang Zhao, and Dik~Lun
  Lee.
\newblock Billion-scale commodity embedding for e-commerce recommendation in
  alibaba.
\newblock In \emph{Proceedings of the 24th {ACM} {SIGKDD} International
  Conference on Knowledge Discovery {\&} Data Mining, {KDD} 2018, London, UK,
  August 19-23, 2018}, pp.\  839--848. {ACM}, 2018.
\newblock \doi{10.1145/3219819.3219869}.

\bibitem[Wu et~al.(2021)Wu, Wu, Cao, Huang, and Xie]{wu2021fedgnn}
Chuhan Wu, Fangzhao Wu, Yang Cao, Yongfeng Huang, and Xing Xie.
\newblock Fedgnn: Federated graph neural network for privacy-preserving
  recommendation.
\newblock \emph{arXiv preprint arXiv:2102.04925}, 2021.

\bibitem[Ying et~al.(2018)Ying, He, Chen, Eksombatchai, Hamilton, and
  Leskovec]{ying2018graph}
Rex Ying, Ruining He, Kaifeng Chen, Pong Eksombatchai, William~L. Hamilton, and
  Jure Leskovec.
\newblock Graph convolutional neural networks for web-scale recommender
  systems.
\newblock In \emph{Proceedings of the 24th {ACM} {SIGKDD} International
  Conference on Knowledge Discovery {\&} Data Mining, {KDD} 2018, London, UK,
  August 19-23, 2018}, pp.\  974--983. {ACM}, 2018.
\newblock \doi{10.1145/3219819.3219890}.

\bibitem[Yu et~al.(2019)Yu, Yang, and Zhu]{yu2019parallel}
Hao Yu, Sen Yang, and Shenghuo Zhu.
\newblock Parallel restarted sgd with faster convergence and less
  communication: Demystifying why model averaging works for deep learning.
\newblock In \emph{Proceedings of the AAAI Conference on Artificial
  Intelligence}, volume~33, pp.\  5693--5700, 2019.

\bibitem[Zeng et~al.(2020)Zeng, Zhou, Srivastava, Kannan, and
  Prasanna]{zeng2019graphsaint}
Hanqing Zeng, Hongkuan Zhou, Ajitesh Srivastava, Rajgopal Kannan, and Viktor~K.
  Prasanna.
\newblock Graphsaint: Graph sampling based inductive learning method.
\newblock In \emph{8th International Conference on Learning Representations,
  {ICLR} 2020, Addis Ababa, Ethiopia, April 26-30, 2020}. OpenReview.net, 2020.

\bibitem[Zhang et~al.(2021)Zhang, Wipf, Gan, and Song]{zhang2021biased}
Qingru Zhang, David Wipf, Quan Gan, and Le~Song.
\newblock A biased graph neural network sampler with near-optimal regret.
\newblock \emph{arXiv preprint arXiv:2103.01089}, 2021.

\bibitem[Zheng et~al.(2020)Zheng, Ma, Wang, Zhou, Su, Song, Gan, Zhang, and
  Karypis]{zheng2020distdgl}
Da~Zheng, Chao Ma, Minjie Wang, Jinjing Zhou, Qidong Su, Xiang Song, Quan Gan,
  Zheng Zhang, and George Karypis.
\newblock Distdgl: Distributed graph neural network training for billion-scale
  graphs.
\newblock \emph{arXiv preprint arXiv:2010.05337}, 2020.

\bibitem[Zou et~al.(2019)Zou, Hu, Wang, Jiang, Sun, and Gu]{zou2019layer}
Difan Zou, Ziniu Hu, Yewen Wang, Song Jiang, Yizhou Sun, and Quanquan Gu.
\newblock Layer-dependent importance sampling for training deep and large graph
  convolutional networks.
\newblock In \emph{Advances in Neural Information Processing Systems 32: Annual
  Conference on Neural Information Processing Systems 2019, NeurIPS 2019,
  December 8-14, 2019, Vancouver, BC, Canada}, pp.\  11247--11256, 2019.

\end{thebibliography}
\bibliographystyle{configs/iclr2022_conference}

\clearpage
\appendix
\makeappendixtitle
\addcontentsline{toc}{section}{Appendix} 
\part{Appendix} 
{
\hypersetup{linkcolor=black}
\parttoc 
}
\clearpage
\clearpage
\section{Additional experimental setup and studies}
\label{appendix:experiments}

We provide detailed experimental setup and additional evaluations in this section.
The link to the GitHub repository is available at \url{https://github.com/MortezaRamezani/llcg/}. 

\subsection{Hardware specification and environment}

For all our experiments we use a single server equipped with 4 \texttt{NVIDIA QUADRO RTX} 8000 GPUs with driver version 460.80, two \texttt{Intel Xeon} 6230 CPU and 768GB of main memory using \texttt{Ubuntu} 18.04 running kernel 5.4.0.
For fair comparison between all methods, we developed a unified framework for performing the experiments, using \texttt{Pytorch} 1.7 compiled for CUDA 11.2.
We also used Pytorch Geometric 1.7.0 and Pytorch Sparse 0.6.8 for various GNN and sparse operations.
Please refer to \texttt{README.md} for detailed instruction on how to install and run the code, and also dataset specific configurations.

\begin{table}[h]
\centering
\caption{Summary of datasets statistics. $\ddag$ indicates multi-labels dataset.}
\label{table:datasets-detailed}
\resizebox{1\linewidth}{!}{
\begin{tabular}{@{}lllccccc@{}} \toprule
\textbf{Dataset}        & \textbf{Nodes} & \textbf{Edges}   & \textbf{Feature}  & \textbf{Classes}  & \textbf{Train / Validation / Test}                        & \textbf{Base Arch.} \\ \midrule 
\texttt{Flickr}~\cite{zeng2019graphsaint}   & $89,250$       & $899,756$        & $500$             & $7$               & 50\%~/~25\%~/~25\%                    & BSBSBL            \\ 
\texttt{OGB-Proteins}~\cite{hu2020open}     & $132,534$      & $39,561,252$     & $8$               & $112^{\bm\ddag}$  & 65\%~/~16\%~/~19\%                    & SSS           \\ 
\texttt{OGB-Arxiv}~\cite{hu2020open}        & $169,343$      & $1,166,243$      & $12$              & $40$              & 54\%~/~17\%~/~28\%                    & GBGBG            \\ 
\texttt{Reddit}~\cite{hamilton2017inductive}& $232,965$      & $11,606,919$     & $602$             & $41$              & 66\%~/~10\%~/~24\%                    & SBSBS            \\ 
\texttt{Yelp}~\cite{zeng2019graphsaint}     & $716,847$      & $13,954,819$     & $300$             & $100^{\bm\ddag}$  & 75\%~/~15\%~/~10\%                    & BSBSBL           \\ 
\texttt{OGB Products}~\cite{hu2020open}     & $2,449,029$    & $61,859,140$     & $100$             & $47$              & \phantom{0}8\%~/~\phantom{1}2\%~/~90\%& GGG       \\ 
\bottomrule
\end{tabular}
}
\end{table}

\subsection{Dataset and models details} \label{section:dataset_and_model_details}

In all experiments, we used ADAM optimizer, and the learning rate, alongside the local machine mini-batch size ($B_L$) and server mini-batch size ($B_S$) are included in \texttt{config} files under \texttt{scripts/configs} folder.
We used three datasets \texttt{Flickr, Reddit, Yelp} from \cite{zeng2019graphsaint} and \texttt{OGB-Proteins, OGB-Arxiv, OGB-Products} from \cite{hu2020open}.
For each dataset, we choose the aggregation method that has been shown to achieve the highest accuracy, and report the results in Table~\ref{table:new-results}.
In addition, we also evaluate two frequently used architectures, GAT and APPNP to show the flexibility of our proposed method to other models.
Note that some of OGB datasets can use more complicated models \cite{sun2021scalable, sun2020adaptive}, however for the sake of simplicity, we only used popular GNN operators which show competitive results in most cases.
Next we formally introduce various layers and operators used in our experiments, followed by the base architecture for each dataset in Table~\ref{table:datasets-detailed}.

\begin{itemize}
    \item \textbf{GCN (G)}: Originally introduced by Kipf et. al~\cite{kipf2016semi}, the representation for node $v_i$ is calculated using:
    \begin{equation}
        \mathbf{h}_i^{(\ell)} = \sigma\Big( \sum_{j\in\mathcal{N}(v_i)} \alpha_{i,j} \mathbf{h}_j^{(\ell-1)} \mathbf{W}^{(\ell)}  \Big),
    \end{equation}
    where $\alpha_{i,j} = \frac{1}{\sqrt{\deg(v_i)\deg{(v_j)}}}$ for symmetric normalized Laplacian and $\alpha_{i,j} = \frac{1}{\deg(v_i)}$ for row normalized Laplacian.
    \item \textbf{SAGE (S)}: First introduce in~\cite{hamilton2017inductive}, SAGE learn different weights for the node and its neighbors and the representation is calculated by:
    \begin{equation}
        \mathbf{h}_i^{(\ell)} = \sigma\Big( \mathbf{h}_i^{(\ell-1)} \mathbf{W}_1^{(\ell)} + \frac{1}{|\mathcal{N}(v_i)|}\sum_{j\in\mathcal{N}(v_i)}  \mathbf{h}_j^{(\ell-1)} \mathbf{W}_2^{(\ell)} \Big)
    \end{equation}
    Note that different operators such as concatenation can be used instead of addition in the above formula. However in our experiments we used addition for all SAGE layers.
    \item \textbf{Linear (L)}: Applies a linear transformation to the node features. In other words, the graph structure is completely ignored in this layer and the representation is computed as follows.
    \begin{equation}
        \mathbf{h}_i^{(\ell)} = \mathbf{h}_i^{(\ell-1)} \mathbf{W}^{(\ell)}
    \end{equation}
    \item \textbf{BatchNorm (B)}: Applies batch normalization according to~\cite{ioffe2015batch} using the following where $\gamma$ and $\beta$ are learnable parameters and $\epsilon$ is added for numerical stability.
    \begin{equation}
        \mathbf{h}_i^{(\ell)} = \frac{\mathbf{h}_i^{(\ell-1)} - \mathbb{E}(\mathbf{h}_i^{(\ell-1)})}{\sqrt{\mathbf{Var}(\mathbf{h}_i^{(\ell-1)}) + \epsilon}} * \gamma + \beta
    \end{equation}
    \item \textbf{GAT}: The Graph Attention layer proposed by~\citep{velivckovic2017graph} as follows.
    \begin{equation}
        \mathbf{h}_i^{(\ell)} = \sigma\Big(\sum_{j\in\mathcal{N}(v_i)}  \alpha_{i,j}\mathbf{h}_j^{(\ell-1)} \mathbf{W}^{(\ell)} \Big)
    \end{equation}
    where $\alpha_{i,j}$ is the attention between node $i$ and $j$ and calculated by:
    \begin{equation}
        \alpha_{i,j} = \frac{exp(\text{LeakyRelu}(\bm{a}[\mathbf{h}_i^{(\ell-1)} \mathbf{W}^{(\ell)} || \mathbf{h}_j^{(\ell-1)} \mathbf{W}^{(\ell)}]))}{\sum_{k\in\mathcal{N}(v_i)} exp(\text{LeakyRelu}(\bm{a}[\mathbf{h}_i^{(\ell-1)} \mathbf{W}^{(\ell)} || \mathbf{h}_k^{(\ell-1)}\mathbf{W}^{(\ell)}]))}
    \end{equation}
    where $\bm{a}$ is the learnable weight vector and $||$ indicates concatenation.
    \item \textbf{APPNP}: This network is proposed by~\citep{klicpera2018predict}, which is a combination of graph-agnostic prediction and label propagation:
    \begin{equation}
        \mathbf{h}_i^{(\ell)} = \beta\mathbf{h}_i^{(0)} + (1-\beta) \sum_{j\in\mathcal{N}(v_i)}  \alpha_{i,j}\mathbf{h}_j^{(\ell-1)}
    \end{equation}
    where $\alpha_{i,j}$ is the same as defined in GCN, $\beta$ is the teleport probability and $\mathbf{h}^{(0)}$ is the output of the first linear layer.
\end{itemize}

\begin{figure}[h]
    \vspace{-10pt}
    \centering
    \begin{minipage}{0.40\textwidth}
        \centering
        \includegraphics[width=1\linewidth]{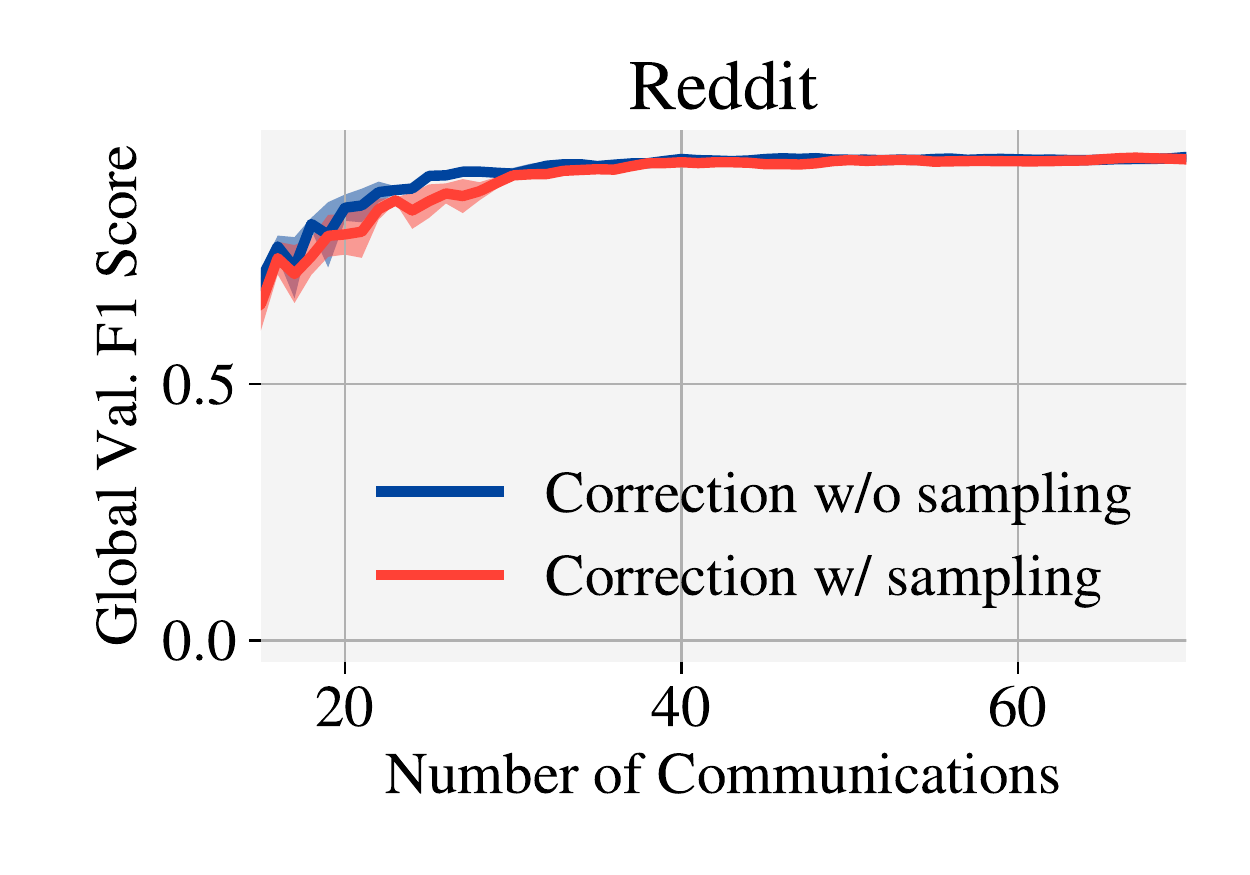}
        \vspace{-16pt}
        \caption{Impact of sampling in correction steps on the \texttt{Reddit} dataset.}
        \label{figure:correction-sampling}
    \end{minipage}
    \hfill
    \begin{minipage}{0.40\textwidth}
        \centering
        \includegraphics[width=1\linewidth]{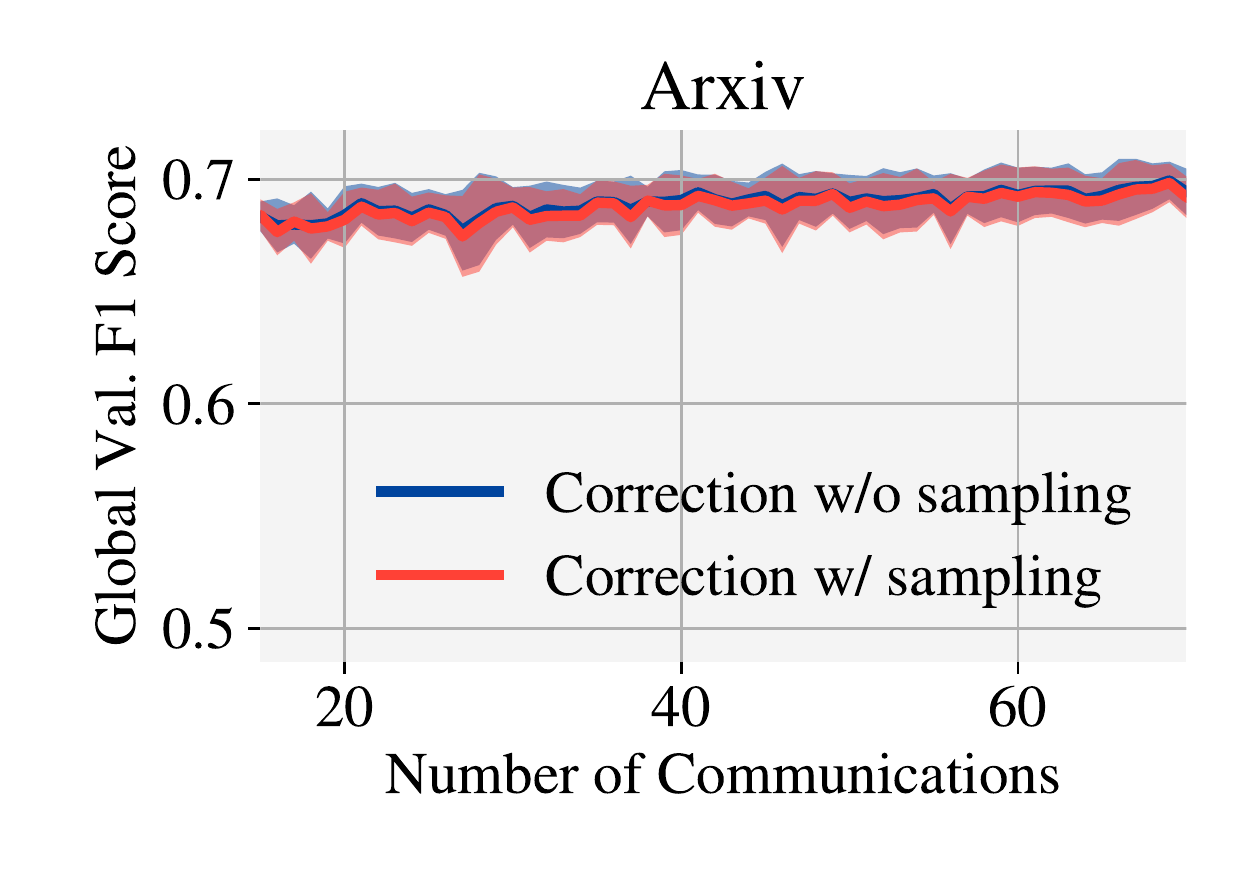}
        \vspace{-16pt}
        \caption{Impact of sampling in correction steps on the \texttt{Arxiv} dataset.}
        \label{figure:correction-sampling-arxiv}
    \end{minipage}
    \vspace{5pt}
\end{figure}

\noindent\textbf{Effect of sampling at correction.~}
Recall that \CG{} requires full-neighbors for the server correction step for the convergence analysis, however, we find that server correction with neighbor sampling also works well in practice. 
As shown in Figure~\ref{figure:correction-sampling} and Figure~\ref{figure:correction-sampling-arxiv}, although server correction with neighbor sampling can introduce some randomness at the beginning of the training phase, the final accuracy of training is very close to the server correction with full-neighbors. 

\subsection{Minibatch selection for correction step}
\begin{figure}[h]
    \vspace{-10pt}
    \centering
    \begin{minipage}{0.40\textwidth}
        \centering
        \includegraphics[width=1\linewidth]{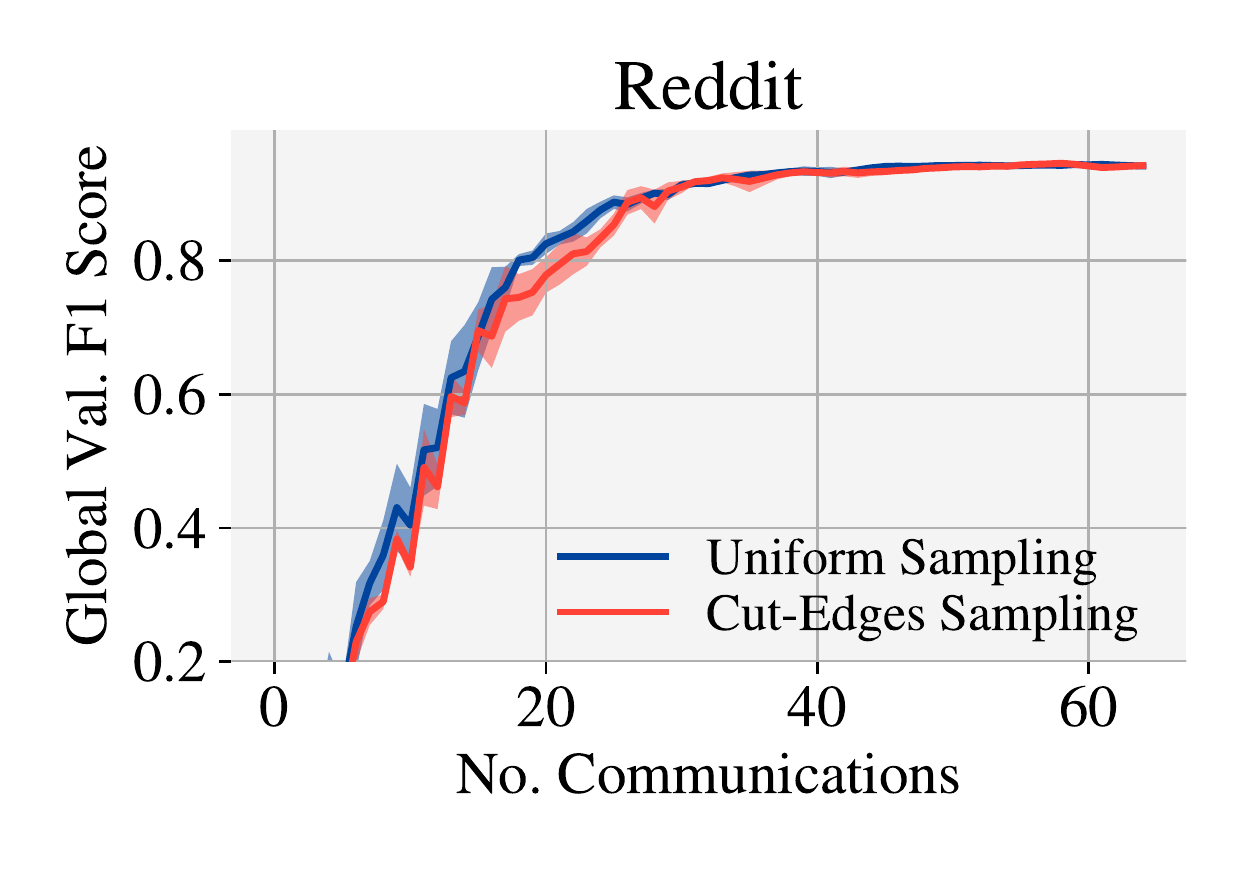}
        \vspace{-16pt}
    \end{minipage}
    \hfill
    \begin{minipage}{0.40\textwidth}
        \centering
        \includegraphics[width=1\linewidth]{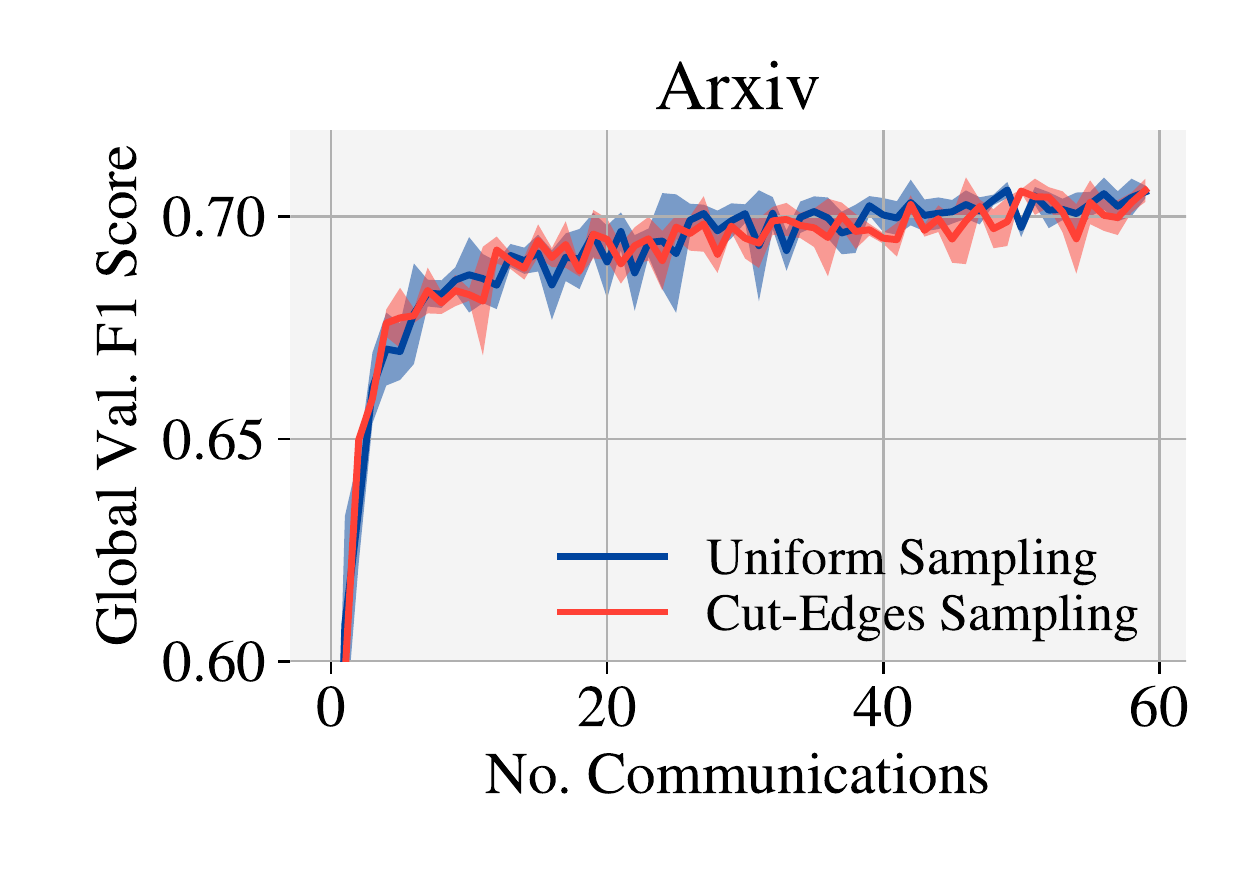}
        \vspace{-16pt}
    \end{minipage}
    \caption{Comparing the validation F1-score of uniform sampling and sampling more cut-edges (i.e., \textit{max. cut edges mini-batch}) on the \texttt{Reddit} and \texttt{Arxiv} dataset.}
    \label{figure:srv-mb}
    \vspace{5pt}
\end{figure}

Recall that in a server correction step, the mini-batch is selected by sampling from the entire graph, uniformly at random to estimate an unbiased stochastic gradient of the global loss function. 
However, one might suggest to include more cut-edges, which are missing from the local machines, in the correction minibatch instead of uniform random sampling, to improve the performance of the global model.
To this end, we conduct an experiments on two datasets (\texttt{Reddit} and \texttt{Arxiv}), where we compare the random minibatch (default setting in \CG{}) and minibatch with higher number of cut-edges, by selecting the nodes on the ends of cut-edges and building the minibatch from there.
As shown in Figure~\ref{figure:srv-mb}, including more cut-edges nodes in the mini-batch does not make significant improvement when comparing to selecting mini-batchs using uniform sampling. 
This potentially happens due to the fact that sampling more cut-edges make the gradient of server correction step biased, while we need an unbiased gradient of full graph for  correction steps at server.

\subsection{Effectiveness of correction}
\label{appendix:largedata}

\begin{figure}[h]
    \begin{subfigure}[t]{0.32\textwidth}
        \centering
        \includegraphics[trim=0 0pt 0 0, width=1\linewidth]{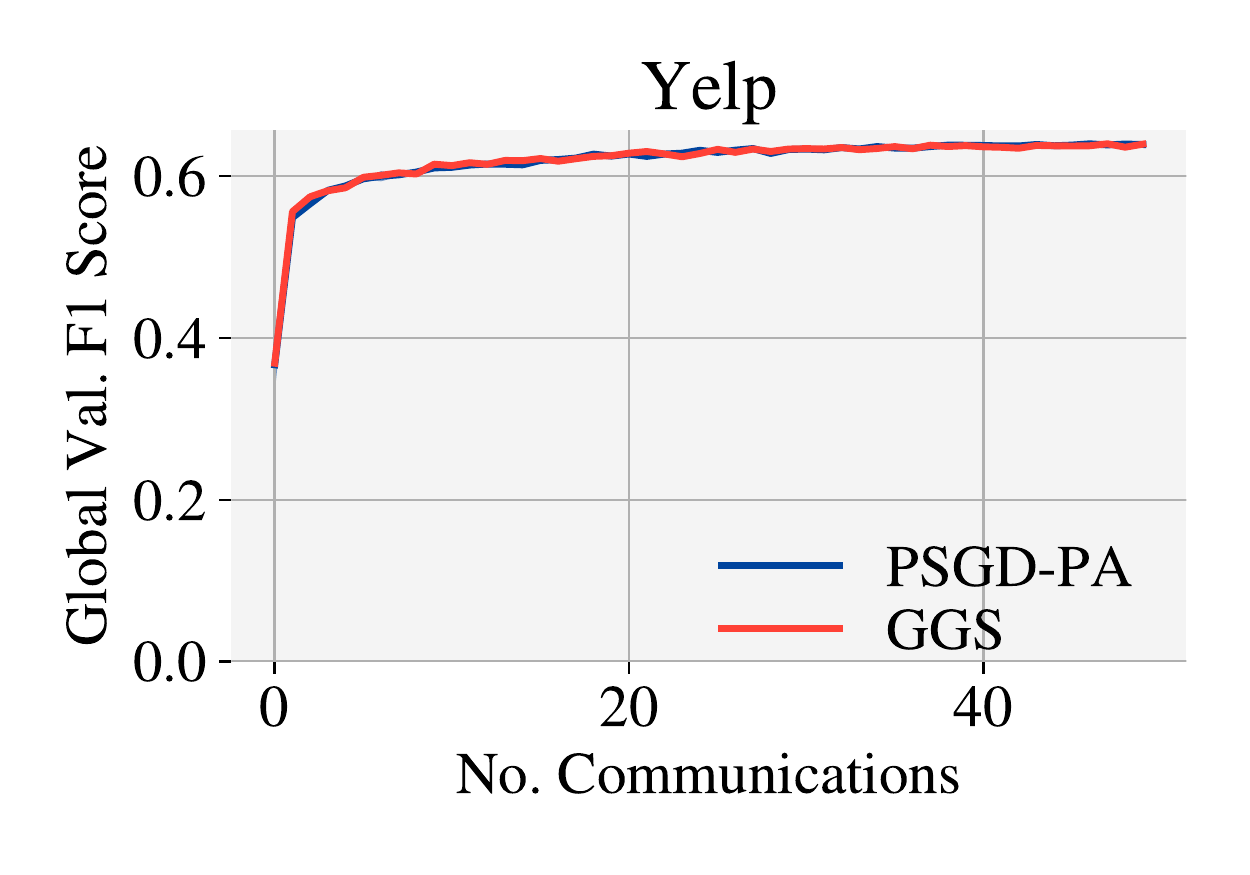}
        \caption{}
        \label{figure:acc-yelp}
    \end{subfigure}
    \hfill
    \begin{subfigure}[t]{0.32\textwidth}
        \centering
        \includegraphics[trim=0 0pt 0 0, width=1\linewidth]{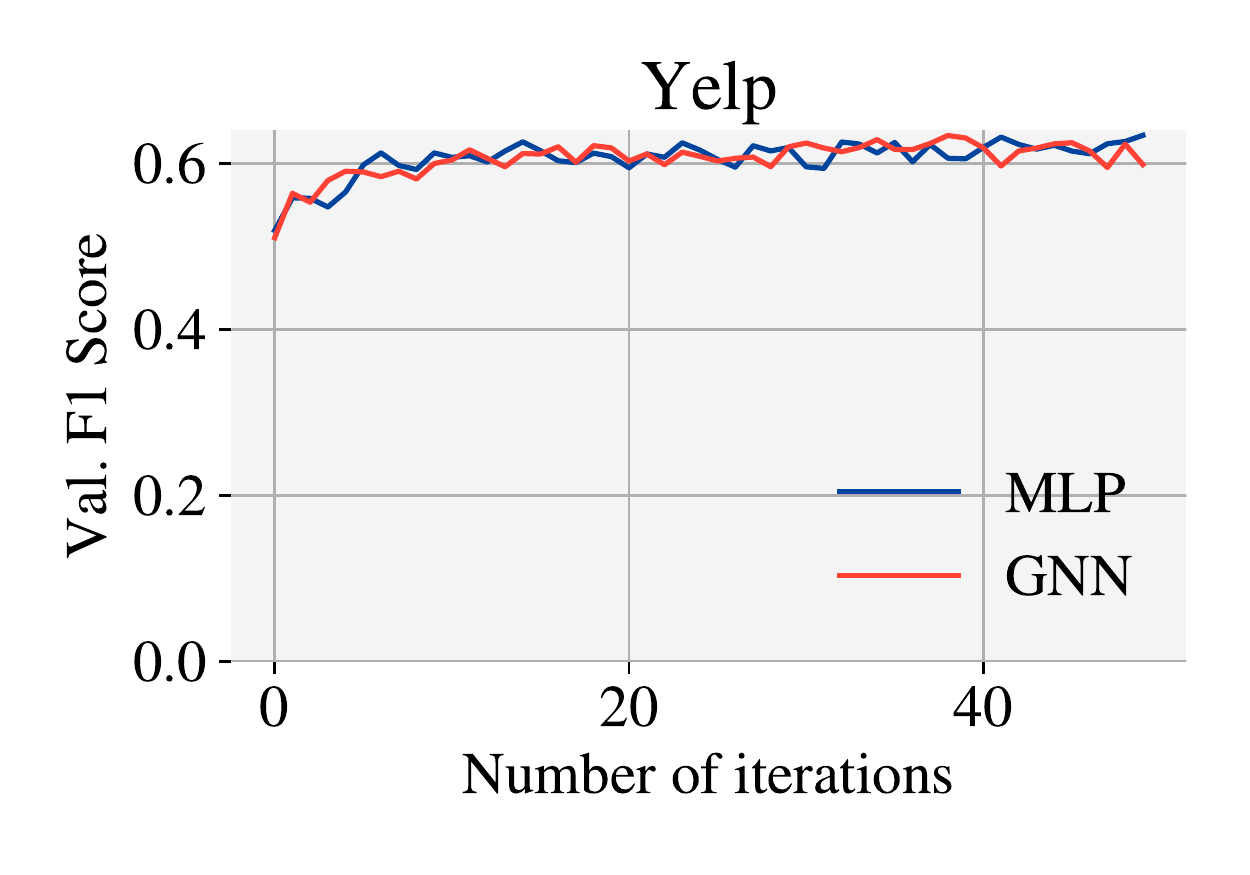}
        \caption{}
        \label{figure:bad-yelp}
    \end{subfigure}
    \hfill
    \begin{subfigure}[t]{0.32\textwidth}
        \centering
        \includegraphics[trim=0 0pt 0 0, width=1\linewidth]{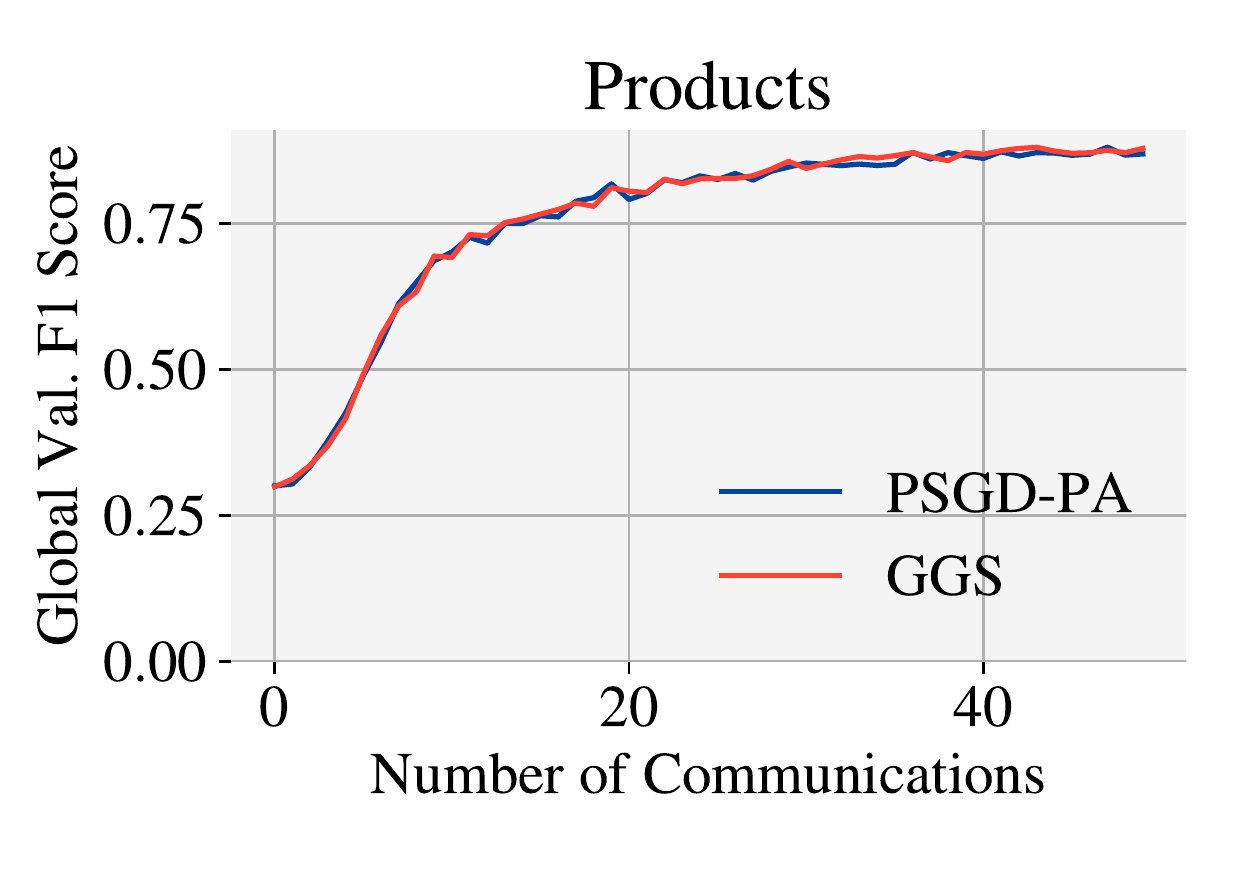}
        \caption{}
        \label{figure:products-bad}
    \end{subfigure}
    \caption{Comparing (a) PSGD-PA vs GGS and (b) MLP vs GNN on the \texttt{Yelp} dataset. (c) Comparing PSGD-PA and GGS for \texttt{OGB-Products}. We exclude the error bar from the figure for better visibility. }
    \label{figure:add-results}
\end{figure}

To illustrate the effectiveness  of correction steps and relate it to the benefits of global graph structure,  we conduct the same experiments on \texttt{Yelp} and \texttt{OGB-Products} with 700K and 2.4M nodes, respectively and report the results in Figure~\ref{figure:add-results}.

\noindent\textbf{Yelp dataset.~}
In the case of \texttt{Yelp} in Figure~\ref{figure:acc-yelp}, as we can see the PSGD-PA and GGS are performing quite similarly.
To further investigate this, we compared the validation accuracy of Yelp using GCN against MLP (i.e., without utilizing the graph structure in training), where the GCN layers replaced by Linear layer (i.e LLL instead of GGG architecture) and plot the validation accuracy per iterations in Figure~\ref{figure:bad-yelp}.
As it can be deduced from this figure, for this dataset MLP is performing as good as GCN, which means this dataset does not depend on the global structure of the graph, further explaining why there is no performance gap between PSGD-PA and GGS.
In other words, no server correction is necessary and we can use $S=0$ in this case for LLCG.

\noindent\textbf{OGB-Products.~}
Similarly, in the case of \texttt{OGB-Products} in Figure~\ref{figure:products-bad}, we cannot see any noticeable accuracy drop due to distributed training.
However, unlike \texttt{Yelp}, this is mainly due to very small $\kappa$ which is caused by two main factors:
(1) very small ratio of training nodes for this dataset, only $8\%$ of the nodes are used for training,
and (2) very small number of cut-edges (less than $7\%$) after applying \texttt{METIS} on this dataset.
It is also worth mentioning that due to the shallow structure of the model required for this dataset, we barely need multi-hop neighbors and consequently we cannot see any noticeable performance drop.

\subsection{Effectiveness in large-scale settings}

In this section, we further investigate the  effect of large number of local machines and large graphs on the performance of proposed algorithm, under the simulated environment with up to $16$ local machines on \texttt{OGB-Products} and \texttt{OGB-MAG240M}~\cite{hu2021ogb} datasets for node classification.
Besides, we also compare against subgraph approximation distributed GNN training algorithm~\cite{angerd2020distributed} using $10\%$ extra storage overhead and fully synchronous distributed GNN training.\footnote{Suppose there are $N_p$ nodes on the $p$-th local machine, then subgraph of size $10\% \times N_p$ is sampled and stored on the $p$th local machine. Notice that $10\%$ storage overhead is the maximum overhead recommended by~\cite{angerd2020distributed}, which is expected to bring the best accuracy performance.}

\paragraph{Node classification tasks.}

\begin{figure}[h]
    \begin{subfigure}[t]{0.49\textwidth}
        \centering
        \includegraphics[trim=0 0pt 0 0, width=1\linewidth]{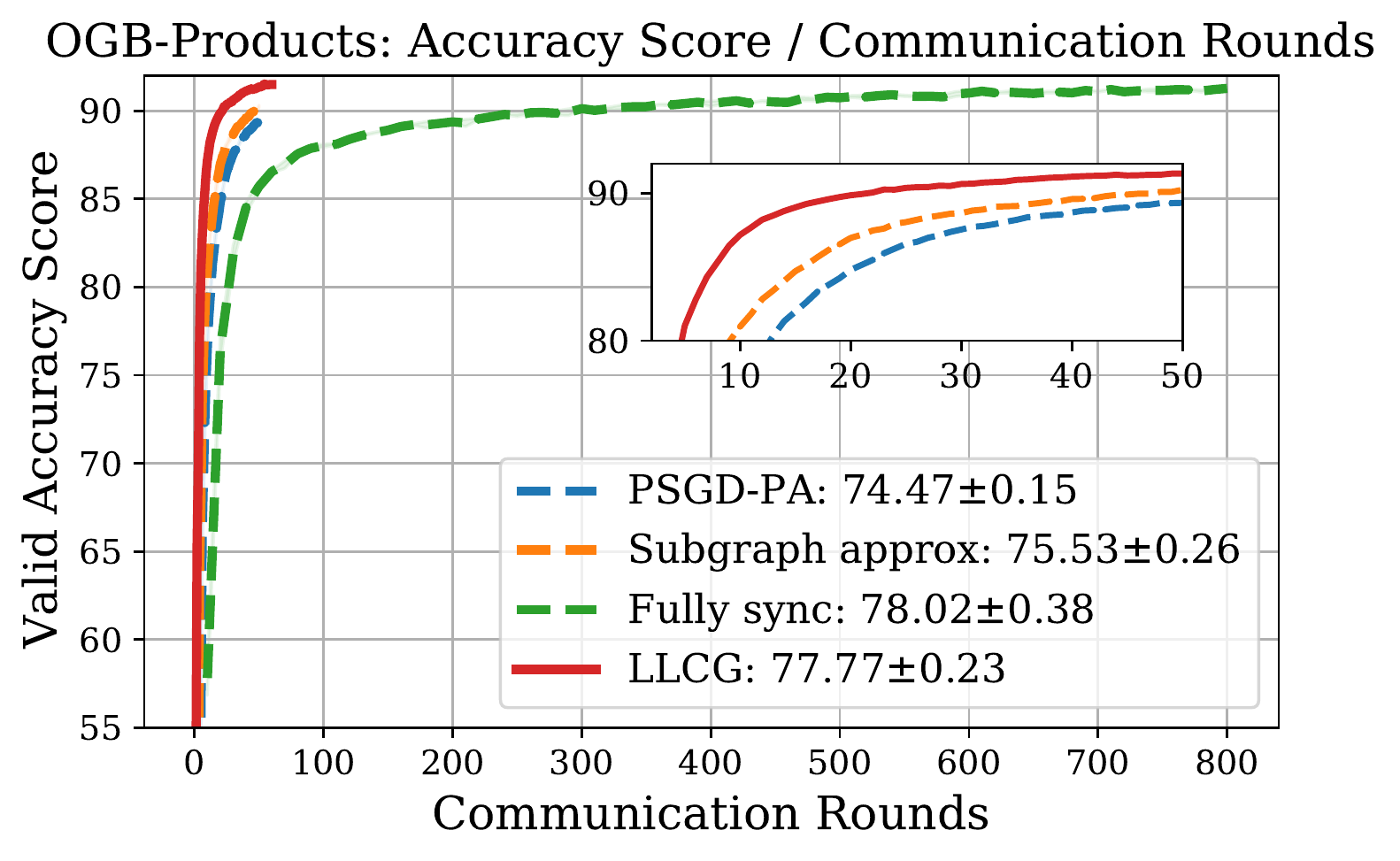}
        \caption{}
    \end{subfigure}
    \hfill
    \begin{subfigure}[t]{0.49
    \textwidth}
        \centering
        \includegraphics[trim=0 0pt 0 0, width=1\linewidth]{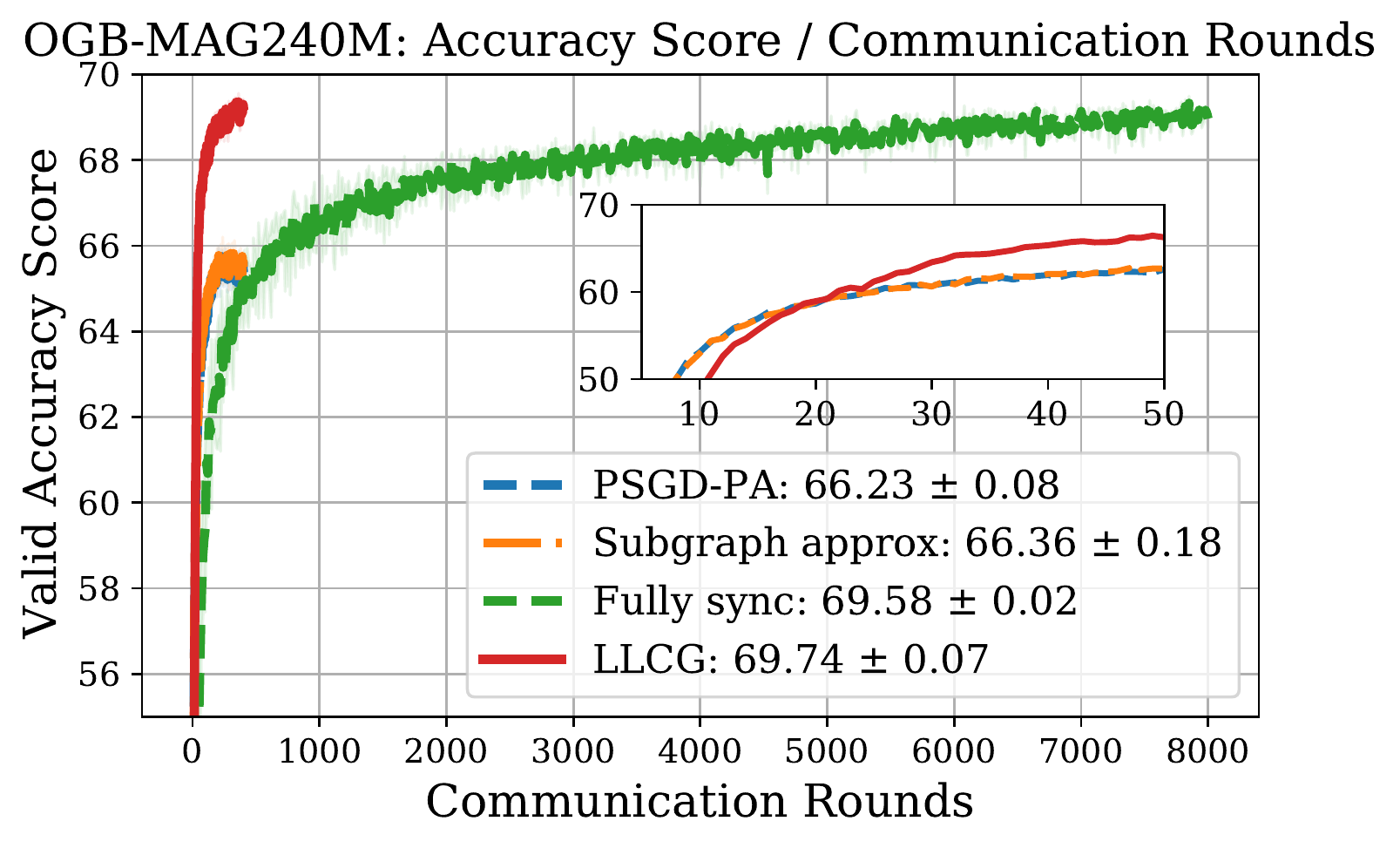}
        \caption{}
    \end{subfigure}
    \begin{subfigure}[t]{0.49\textwidth}
        \centering
        \includegraphics[trim=0 0pt 0 0, width=1\linewidth]{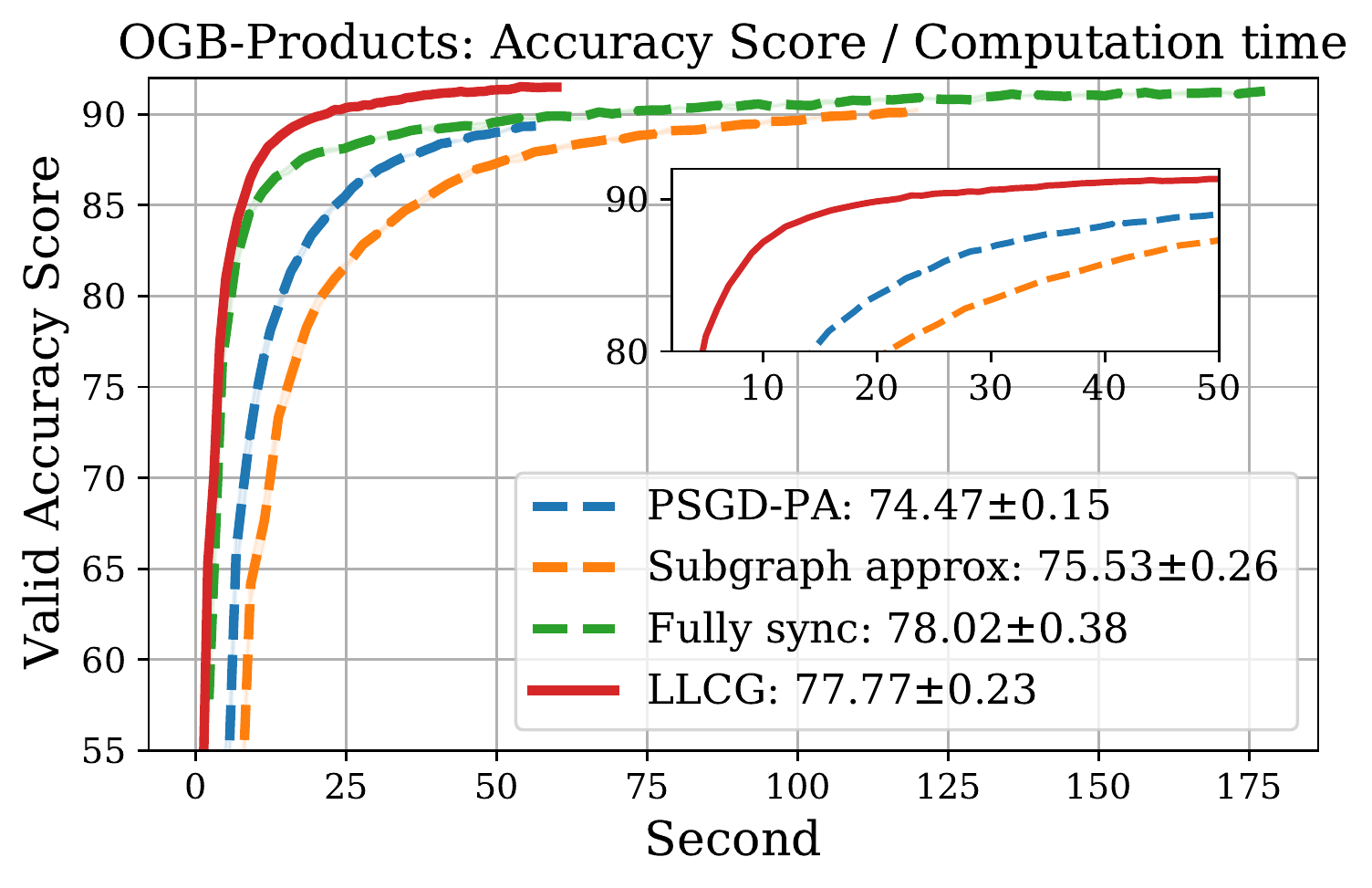}
        \caption{}
    \end{subfigure}
    \hfill
    \begin{subfigure}[t]{0.49
    \textwidth}
        \centering
        \includegraphics[trim=0 0pt 0 0, width=1\linewidth]{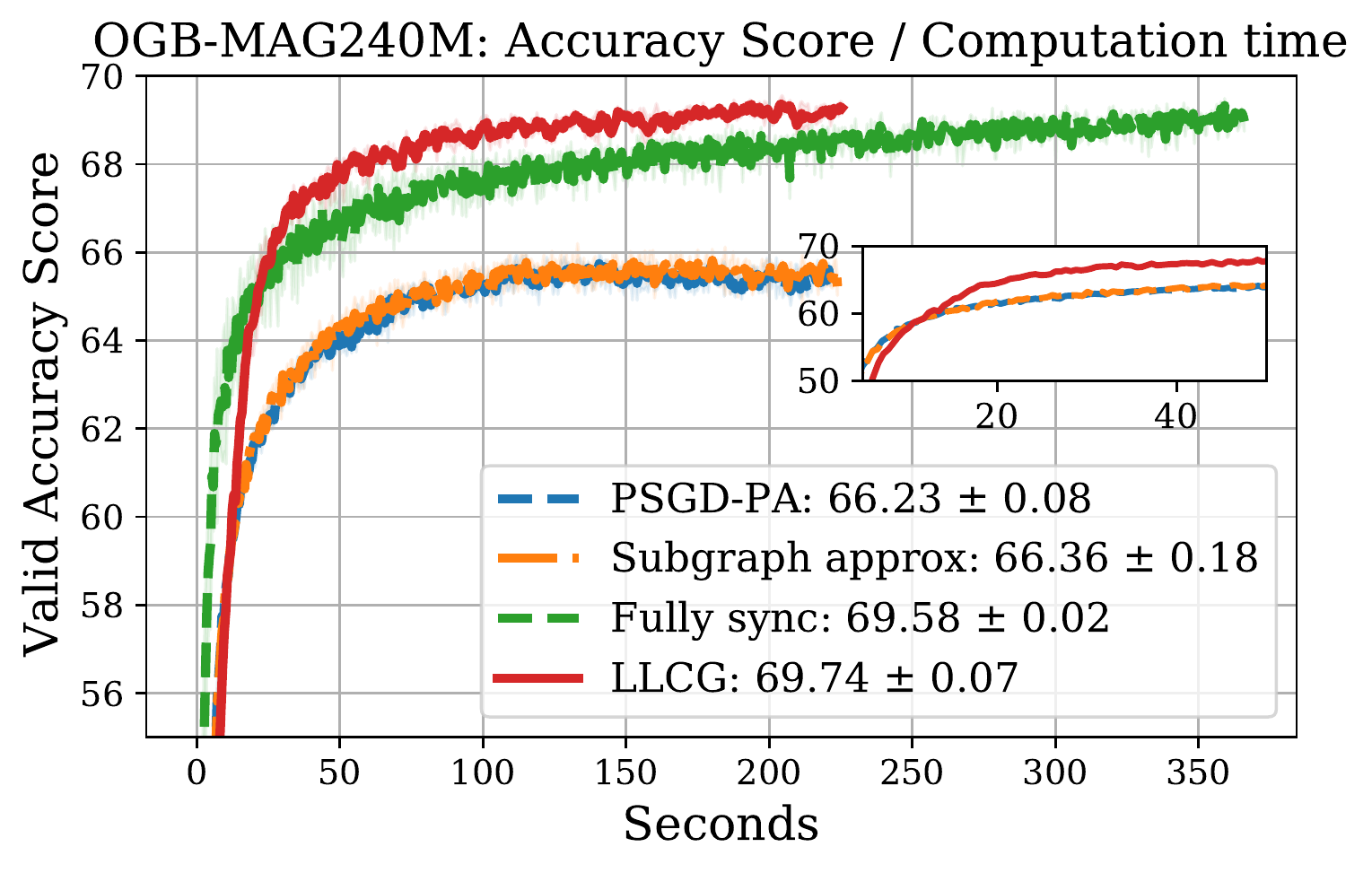}
        \caption{}
    \end{subfigure}
    \caption{Comparing PSGD-PA, periodic model averaging with subgraph approximation, full sync distributed GNN training, and \CG{} on \texttt{OGB-Products} and \texttt{OGB-MAG240M} datasets on a simulated environment with $16$ local machines.}
    \label{figure:ogb-large-scale}
\end{figure}

For the \texttt{OGB-Products} dataset, we train a $3$-layer GraphSAGE model with learning rate $0.003$ with $50$ rounds of communications ,  which has the same hyper-parameter configuration and model architecture as \texttt{OGB-Products}'s \href{https://github.com/snap-stanford/ogb/blob/master/examples/nodeproppred/products/gnn.py}{leaderboard implementation }.
For the \texttt{OGB-MAG240M} dataset, we train a $2$-layer skip-connected GraphSAGE model with learning rate $0.001$ with $400$ rounds of communications, which has the same hyper-parameter selection and model architecture as \texttt{OGB-MAG240M}'s \href{https://github.com/snap-stanford/ogb/blob/master/examples/lsc/mag240m/gnn.py}{leaderboard implementation}. 
As shown in Figure~\ref{figure:ogb-large-scale}, we can observe that 
\begin{itemize}
    \item PSGD-PA suffers from performance degeneration and has a large gap to the full sync training accuracy. Subgraph approximation might alleviate the issue to some extent, but requires significant storage overhead. Our proposal \CG{} can bridge the gap between PSGD-PA and full sync training.
    \item The pure computation time\footnote{We record the time using Python's \texttt{time.time()} function. } is negligible when ignoring the communication time but just consider the computation time. Besides, the server correction step does not introduce significant computation overhead.
\end{itemize}
It is worth mentioning that we do not observe the performance gap between PSGD-PA and fully sync distributed training in Figure~\ref{figure:products-bad} because of different number of layers (2-layers) and limited number of local machine (8 local machines) are used in the previous experiments. 
However, as the number of layers and number of local machines increase, the effect of ignoring cut-edges and data heterogeneity is becoming more serious, which is an interesting observation and worth exploring as a potential future direction.

\section{Proof of Theorem~\ref{theorem:distgnn_param_avg}}

In this section, we first introduce the useful lemmas in Section~\ref{section:thm1_useful_lemmas}, then process our proof of theorem in Section~\ref{section:thm1_main}.
In particular, we show that solely averaging the local machine models and ignoring the global graph structure will suffer from an irreducible residual error.

\subsection{Useful lemmas} \label{section:thm1_useful_lemmas}
The following lemma gives the upper bound for the norm of the stochastic gradient on each local machine as the norm of the expectation of local gradient and the stochastic gradient variance scaled by the number of local machines $P$.

\begin{lemma} \label{lemma:gradient_variance}
Let $\tilde{\nabla} \mathcal{L}_p^\text{local}(\boldsymbol{\theta}_p; \xi_p)$ be stochastic gradients such that 
\begin{equation}
    \begin{aligned}
    \mathbb{E}\left[ \left\| \tilde{\nabla} \mathcal{L}_p^\text{local}(\boldsymbol{\theta}_p; \xi_p) - \mathbb{E}[\tilde{\nabla} \mathcal{L}_p^\text{local}(\boldsymbol{\theta}_p; \xi_p)] \right\|^2 \right] &\leq \sigma^2_\text{var},~ \\
    \end{aligned}
\end{equation}
Then, we have
\begin{equation}
    \mathbb{E}\Big[ \Big\| \frac{1}{P} \sum_{p=1}^P \tilde{\nabla} \mathcal{L}_p^\text{local}(\boldsymbol{\theta}_p; \xi_p) \Big\|^2  \Big] \leq 
    \frac{\sigma^2_\text{var}}{P} + \mathbb{E}\Big[ \Big\| \frac{1}{P} \sum_{p=1}^P \mathbb{E} \left[ \tilde{\nabla} \mathcal{L}_p^\text{local}(\boldsymbol{\theta}_p, \xi_p^t)  \right]\Big\|^2  \Big]
\end{equation}

\end{lemma}

The following lemma provides an upper bound on the difference of each local gradient $\nabla \mathcal{L}_p^\text{local}(\boldsymbol{\theta}_p)$ compared to $\smash{\frac{1}{P}\sum_{q=1}^P \nabla \mathcal{L}_q^\text{local}(\boldsymbol{\theta}_q)}$, which is a function of:
\begin{itemize} 
    \item The deviation of each local model to the virtual averaged model, i.e., $\frac{1}{P}\sum_{i=1}^P \| \boldsymbol{\theta}_p - \bar{\boldsymbol{\theta}} \|^2$.
    \item The difference of gradient computed with and without having access to the full graph information, i.e., $\kappa^2$.
\end{itemize}

\begin{lemma} \label{lemma:local_diff}
For any $P$ machines $\boldsymbol{\theta}_p,~p\in\{1,\ldots,P\}$, if we define $\bar{\boldsymbol{\theta}} = \frac{1}{P}\sum_{i=1}^P \boldsymbol{\theta}_p$ and let $\kappa>0$ such that $\smash{\| \nabla \mathcal{L}_p^\text{local}(\boldsymbol{\theta}) - \nabla \mathcal{L}(\boldsymbol{\theta}) \|^2 \leq \kappa^2}$, we have
\begin{equation}
    \frac{1}{P}\sum_{p=1}^P \Big\| \nabla \mathcal{L}_p^\text{local}(\boldsymbol{\theta}_p) - \frac{1}{P}\sum_{q=1}^P \nabla \mathcal{L}_q^\text{local}(\boldsymbol{\theta}_q) \Big\|^2 \leq \frac{8L^2}{P}\sum_{i=1}^P \| \boldsymbol{\theta}_p - \bar{\boldsymbol{\theta}} \|^2 + 8 \kappa^2.
\end{equation}
\end{lemma}

The following lemma provides an upper bound on the deviation of each local model to the virtual averaged model, which plays an important role in the previous lemma.

\begin{lemma} \label{lemma:global_local_diff}
For all $P$ machines with parameters $\boldsymbol{\theta}_p,~p\in\{1,\ldots,P\}$, if we define $\bar{\boldsymbol{\theta}} = \frac{1}{P}\sum_{i=1}^P \boldsymbol{\theta}_p$ and let $\kappa>0$ such that $\| \mathcal{L}_p^\text{local}(\boldsymbol{\theta}) - \nabla \mathcal{L}(\boldsymbol{\theta}) \|^2 \leq \kappa^2$, we have
\begin{equation}
    \sum_{t=0}^{T-1} \frac{1}{P}\sum_{p=1}^P  \mathbb{E}[ \| \bar{\boldsymbol{\theta}}^t - \boldsymbol{\theta}_p^t \|^2 ] \leq \frac{4 \eta^2 K \sigma^2_\text{var}}{1- 16 L^2 \eta^2 K^2} T + \frac{4 \eta^2 K^2 \sigma^2_\text{bias} + 16 \eta^2 K^2 \kappa^2}{1- 16 L^2 \eta^2 K^2} T.
\end{equation}
\end{lemma}

\subsection{Main proof of Theorem~\ref{theorem:distgnn_param_avg}} \label{section:thm1_main}

Equipped with the above lemmas, we are now ready to present the  proof of Theorem~\ref{theorem:distgnn_param_avg}. 
From the smoothness assumption, we have
\begin{equation} \label{eq:thm1_proof_1}
    \mathbb{E}[\mathcal{L}(\bar{\boldsymbol{\theta}}^{t+1})] \leq \mathbb{E}[\mathcal{L}(\bar{\boldsymbol{\theta}}^t)] + \mathbb{E}[ \langle \nabla \mathcal{L}(\bar{\boldsymbol{\theta}}^t), \bar{\boldsymbol{\theta}}^{t+1} - \bar{\boldsymbol{\theta}}^t  \rangle ] + \frac{L}{2} \mathbb{E}[\| \bar{\boldsymbol{\theta}}^{t+1} - \bar{\boldsymbol{\theta}}^t \|^2].
\end{equation}

Recall that $\bar{\boldsymbol{\theta}}^t$ is defined as $\bar{\boldsymbol{\theta}}^t = \frac{1}{P}\sum_{i=1}^P \boldsymbol{\theta}_p^t$ for any $t$.
Then, according to the update rule
\begin{equation}
    \bar{\boldsymbol{\theta}}^{t+1} = \bar{\boldsymbol{\theta}}^t -  \frac{\eta}{P}\sum_{p=1}^P \tilde{\nabla} \mathcal{L}_p^\text{local}(\boldsymbol{\theta}_p^t; \xi_p^t),~
\end{equation} 
we have
\begin{equation} \label{eq:thm1_proof_2}
    \begin{aligned}
    \mathbb{E}[ \langle \nabla \mathcal{L}(\bar{\boldsymbol{\theta}}^t), \bar{\boldsymbol{\theta}}^{t+1} - \bar{\boldsymbol{\theta}}^t  \rangle ] 
    &= - \eta \mathbb{E} \Big[ \Big\langle \nabla \mathcal{L}(\bar{\boldsymbol{\theta}}^t), \frac{1}{P}\sum_{p=1}^P \tilde{\nabla} \mathcal{L}_p^\text{local}(\boldsymbol{\theta}_p^t; \xi_p^t)  \Big\rangle \Big] \\
    &= - \eta \Big\langle \nabla \mathcal{L}(\bar{\boldsymbol{\theta}}^t), \frac{1}{P}\sum_{p=1}^P \mathbb{E}\left[ \tilde{\nabla} \mathcal{L}_p^\text{local}(\boldsymbol{\theta}_p^t; \xi_p^t) \right] \Big\rangle.
    \end{aligned}
\end{equation}

We can upper bound the right hand side of Eq.~\ref{eq:thm1_proof_2} by
\begin{equation} \label{eq:thm1_proof_3}
    \begin{aligned}
    &-\eta \Big\langle \nabla \mathcal{L}(\bar{\boldsymbol{\theta}}^t), \frac{1}{P}\sum_{p=1}^P \mathbb{E}[\tilde{\nabla} \mathcal{L}_p^\text{local}(\boldsymbol{\theta}_p^t;\xi_p^t)] \Big\rangle \\
    &\underset{(a)}{=} - \frac{\eta}{2} \Big( \| \nabla \mathcal{L}(\bar{\boldsymbol{\theta}}^t) \|^2 + \Big\| \frac{1}{P}\sum_{p=1}^P \mathbb{E}[ \tilde{\nabla} \mathcal{L}_p^\text{local}(\boldsymbol{\theta}_p^t;\xi_p^t)] \Big\|^2 - \Big\| \nabla \mathcal{L}(\bar{\boldsymbol{\theta}}^t) - \frac{1}{P}\sum_{p=1}^P \mathbb{E}[ \tilde{\nabla} \mathcal{L}_p^\text{local}(\boldsymbol{\theta}_p^t;\xi_p^t)] \Big\|^2 \Big) \\
    &\underset{(b)}{\leq} - \frac{\eta}{2} \Big( \| \nabla \mathcal{L}(\bar{\boldsymbol{\theta}}^t) \|^2 + \Big\| \frac{1}{P}\sum_{p=1}^P \mathbb{E}[ \tilde{\nabla} \mathcal{L}_p^\text{local}(\boldsymbol{\theta}_p^t;\xi_p^t)] \Big\|^2 - 2\kappa^2- 4\sigma_\text{bias}^2 - \frac{4L^2}{P} \sum_{p=1}^P \| \bar{\boldsymbol{\theta}}^t - \boldsymbol{\theta}_p^t \|^2 \Big) ,
    \end{aligned}
\end{equation}
where $(a)$ is due to $2 \langle \mathbf{x}, \mathbf{y} \rangle = \| \mathbf{x} \|^2 + \| \mathbf{y} \|^2 - \| \mathbf{x} - \mathbf{y} \|^2$ and $(b)$ is due to 

\begin{equation} \label{eq:thm1_proof_4}
    \begin{aligned}
    &\Big\| \nabla \mathcal{L}(\bar{\boldsymbol{\theta}}^t) - \frac{1}{P}\sum_{p=1}^P \mathbb{E}[ \tilde{\nabla} \mathcal{L}_p^\text{local}(\boldsymbol{\theta}_p^t;\xi_p^t)] \Big\|^2 \\
    &= \left\| \frac{1}{P}\sum_{p=1}^P \left( \nabla \mathcal{L}(\bar{\boldsymbol{\theta}}^t) - \nabla \mathcal{L}_p^\text{local}(\bar{\boldsymbol{\theta}}^t) + \nabla \mathcal{L}_p^\text{local}(\bar{\boldsymbol{\theta}}^t)  - \mathbb{E}[ \tilde{\nabla} \mathcal{L}_p^\text{local}(\boldsymbol{\theta}_p^t;\xi_p^t)] \right) \right\|^2 \\
    &\leq \frac{2}{P}\sum_{p=1}^P \| \nabla \mathcal{L}(\bar{\boldsymbol{\theta}}^t) - \nabla \mathcal{L}_p^\text{local}(\bar{\boldsymbol{\theta}}^t) \|^2 + \frac{4}{P}\sum_{p=1}^P \| \nabla \mathcal{L}_p^\text{local}(\bar{\boldsymbol{\theta}}^t) - \nabla \mathcal{L}_p^\text{local}(\boldsymbol{\theta}_p^t) \|^2 \\
    &\quad + \frac{4}{P}\sum_{p=1}^P \left\| \mathbb{E}[ \tilde{\nabla} \mathcal{L}_p^\text{local}(\bar{\boldsymbol{\theta}}^t;\xi_p^t)] - \nabla \mathcal{L}_p^\text{local}(\boldsymbol{\theta}_p^t) \right\|^2 \\
    &\underset{(c)}{\leq} 2\kappa^2 + 4\sigma_\text{bias}^2 + \frac{4 L^2}{P}\sum_{p=1}^P \| \bar{\boldsymbol{\theta}}^t - \boldsymbol{\theta}_p^t\|^2,
    \end{aligned}
\end{equation}

where $(c)$ follows from the smoothness assumption and the definition of $\kappa^2, \sigma_\text{bias}^2$ in Theorem~\ref{theorem:distgnn_param_avg}.

Combining Eq.~\ref{eq:thm1_proof_2} and Eq.~\ref{eq:thm1_proof_3} gives us
\begin{equation} \label{eq:thm1_proof_5}
    \begin{aligned}
    &\mathbb{E}[ \langle \nabla \mathcal{L}(\bar{\boldsymbol{\theta}}^t), \bar{\boldsymbol{\theta}}^{t+1} - \bar{\boldsymbol{\theta}}^t  \rangle ] \\
    &\leq - \frac{\eta}{2}  \mathbb{E}[ \| \nabla \mathcal{L}(\bar{\boldsymbol{\theta}}^t) \|^2 ] - \frac{\eta}{2} \mathbb{E}\Big[ \Big\| \frac{1}{P}\sum_{p=1}^P \mathbb{E}\left[ \tilde{\nabla} \mathcal{L}_p^\text{local}(\boldsymbol{\theta}_p^t,\xi_p^t) \right] \Big\|^2 \Big] \\
    &\quad + \eta (\kappa^2 + 2 \sigma_\text{bias}^2) + \frac{2 \eta L^2}{ P} \sum_{p=1}^P \mathbb{E}[ \| \bar{\boldsymbol{\theta}}^t - \boldsymbol{\theta}_p^t \|^2 ].
    \end{aligned}
\end{equation}

According to the update rule
\begin{equation}
    \bar{\boldsymbol{\theta}}^{t+1} = \bar{\boldsymbol{\theta}}^t -  \frac{\eta}{P}\sum_{p=1}^P \tilde{\nabla} \mathcal{L}_p^\text{local}(\boldsymbol{\theta}_p^t; \xi_p^t),
\end{equation}
we have
\begin{equation} \label{eq:thm1_proof_6}
    \mathbb{E}[ \| \bar{\boldsymbol{\theta}}^{t+1} - \bar{\boldsymbol{\theta}}^t \|^2] =  \eta^2 \mathbb{E}\Big[ \Big\| \frac{1}{P}\sum_{p=1}^P \tilde{\nabla} \mathcal{L}_p^\text{local}(\boldsymbol{\theta}_p^t; \xi_p^t) \Big\|^2 \Big] .
\end{equation}

\clearpage
Substituting Eq.~\ref{eq:thm1_proof_5} and Eq.~\ref{eq:thm1_proof_6} to Eq.~\ref{eq:thm1_proof_1}, we have

\begin{equation} \label{eq:thm1_proof_7}
    \begin{aligned}
    \mathbb{E}[\mathcal{L}(\bar{\boldsymbol{\theta}}^{t+1})] &\leq \mathbb{E}[\mathcal{L}(\bar{\boldsymbol{\theta}}^t)]  - \frac{\eta}{2}  \mathbb{E}[ \| \nabla \mathcal{L}(\bar{\boldsymbol{\theta}}^t) \|^2 ] - \frac{\eta}{2} \mathbb{E}\Big[ \Big\| \frac{1}{P}\sum_{p=1}^P \mathbb{E}\left[ \tilde{\nabla} \mathcal{L}_p^\text{local}(\boldsymbol{\theta}_p^t, \xi_p^t) \right] \Big\|^2 \Big] \\
    &\quad + \eta (\kappa^2 + 2 \sigma_\text{bias}^2) + \frac{2 \eta L^2}{ P} \sum_{p=1}^P \mathbb{E}[ \| \bar{\boldsymbol{\theta}}^t - \boldsymbol{\theta}_p^t \|^2 ] \\
    &\quad + \frac{\eta^2 L}{2} \mathbb{E}\Big[ \Big\| \frac{1}{P}\sum_{p=1}^P \tilde{\nabla} \mathcal{L}_p^\text{local}(\boldsymbol{\theta}_p^t; \xi_p^t) \Big\|^2 \Big] .
    \end{aligned}
\end{equation}

Dividing both sides by $\frac{\eta}{2}$ and rearranging the  terms yields
\begin{equation} 
    \begin{aligned}
       \mathbb{E}[ \| \nabla \mathcal{L}(\bar{\boldsymbol{\theta}}^t) \|^2 ] &\leq \frac{2}{\eta} \Big( \mathbb{E}[\mathcal{L}(\bar{\boldsymbol{\theta}}^t)]  - \mathbb{E}[\mathcal{L}(\bar{\boldsymbol{\theta}}^{t+1})] \Big) - \mathbb{E}\Big[ \Big\| \frac{1}{P}\sum_{p=1}^P \mathbb{E} \left[ \tilde{\nabla} \mathcal{L}_p^\text{local}(\boldsymbol{\theta}_p^t,\xi_p^t) \right] \Big\|^2 \Big] \\
    &\quad + 2(\kappa^2 + 2\sigma_\text{bias}^2) + \frac{ 4 L^2}{ P} \sum_{p=1}^P \mathbb{E}[ \| \bar{\boldsymbol{\theta}}^t - \boldsymbol{\theta}_p^t \|^2 ] + \eta L \mathbb{E}\Big[ \Big\| \frac{1}{P}\sum_{p=1}^P \tilde{\nabla} \mathcal{L}_p^\text{local}(\boldsymbol{\theta}_p^t; \xi_p^t) \Big\|^2 \Big] \\
    &\underset{(a)}{\leq} \frac{2}{\eta} \Big( \mathbb{E}[\mathcal{L}(\bar{\boldsymbol{\theta}}^t)]  - \mathbb{E}[\mathcal{L}(\bar{\boldsymbol{\theta}}^{t+1})] \Big) - \mathbb{E}\Big[ \Big\| \frac{1}{P}\sum_{p=1}^P \mathbb{E}\left[ \tilde{\nabla} \mathcal{L}_p^\text{local}(\boldsymbol{\theta}_p^t, \xi_t^p) \right] \Big\|^2 \Big] \\
    &\quad + 2(\kappa^2 + 2\sigma_\text{bias}^2) + \frac{ 4 L^2}{ P} \sum_{p=1}^P \mathbb{E}[ \| \bar{\boldsymbol{\theta}}^t - \boldsymbol{\theta}_p^t \|^2 ] \\
    &\quad  + \eta L \Big( \mathbb{E}\Big[ \Big\| \frac{1}{P}\sum_{p=1}^P \mathbb{E}\left[ \tilde{\nabla} \mathcal{L}_p^\text{local}(\boldsymbol{\theta}_p^t, \xi_p^t) \right] \Big\|^2  \Big] + \frac{\sigma^2_\text{var}}{P} \Big) \\
    &= \frac{2}{\eta} \Big( \mathbb{E}[\mathcal{L}(\bar{\boldsymbol{\theta}}^t)]  - \mathbb{E}[\mathcal{L}(\bar{\boldsymbol{\theta}}^{t+1})] \Big) +(\eta L - 1 ) \mathbb{E}\Big[ \Big\| \frac{1}{P}\sum_{p=1}^P \mathbb{E}\left[ \tilde{\nabla} \mathcal{L}_p^\text{local}(\boldsymbol{\theta}_p^t, \xi_p^t) \right] \Big\|^2 \Big] \\
    &\quad + 2(\kappa^2 + 2\sigma_\text{bias}^2) + \frac{ 4 L^2}{ P} \sum_{p=1}^P \mathbb{E}[ \| \bar{\boldsymbol{\theta}}^t - \boldsymbol{\theta}_p^t \|^2 ] + \frac{ \eta L\sigma^2_\text{var}}{P},
    \end{aligned}
\end{equation}
where $(a)$ is due to Lemma~\ref{lemma:gradient_variance}.

\clearpage
Summing over $t \in \{ 0, \ldots, T-1\}$ and dividing both side by $T$, we get
\begin{equation} \label{eq:sdlfnkashfdouq}
    \begin{aligned}
    & \frac{1}{T}\sum_{t=0}^{T-1} \mathbb{E}[ \| \nabla \mathcal{L}(\bar{\boldsymbol{\theta}}^t) \|^2 ] \\
    &\leq \frac{2}{\eta T} \Big( \mathcal{L}(\bar{\boldsymbol{\theta}}^0)  - \mathcal{L}(\boldsymbol{\theta}^\star) \Big) + \frac{\eta L - 1}{T} \sum_{t=0}^{T-1}  \mathbb{E}\Big[ \Big\| \frac{1}{P}\sum_{p=1}^P \mathbb{E} \left[ \tilde{\nabla} \mathcal{L}_p^\text{local}(\boldsymbol{\theta}_p^t, \xi_p^t)  \right] \Big\|^2 \Big] \\
    &\quad + 2(\kappa^2 + 2\sigma_\text{bias}^2) + \frac{4 L^2}{T} \sum_{t=0}^{T-1}  \frac{ 1}{ P} \sum_{p=1}^P \mathbb{E}[ \| \bar{\boldsymbol{\theta}}^t - \boldsymbol{\theta}_p^t \|^2 ] + \frac{ \eta L\sigma^2_\text{var}}{P } \\
    &\underset{(a)}{\leq} \frac{2}{\eta T} \Big( \mathcal{L}(\bar{\boldsymbol{\theta}}^0)  - \mathcal{L}(\boldsymbol{\theta}^\star) \Big) + \frac{\eta L - 1}{T} \sum_{t=0}^{T-1}  \mathbb{E}\Big[ \Big\| \frac{1}{P}\sum_{p=1}^P \mathbb{E} \left[ \tilde{\nabla} \mathcal{L}_p^\text{local}(\boldsymbol{\theta}_p^t, \xi_p^t)  \right] \Big\|^2 \Big] + \frac{ \eta L\sigma^2_\text{var}}{P }  \\
    &\quad + 2(\kappa^2 +  2\sigma_\text{bias}^2) + L^2 \frac{16\eta^2 K \sigma_\text{var}^2 }{(1 - 16 \eta^2 L^2K^2)} + L^2 \frac{16 \eta^2 K^2 \sigma_\text{bias}^2 + 64 \eta^2 K^2 \kappa^2 }{(1 - 16 \eta^2 L^2K^2)} \\
    &\underset{(b)}{\leq} \frac{2}{\eta T} \Big( \mathcal{L}(\bar{\boldsymbol{\theta}}^0)  - \mathcal{L}(\boldsymbol{\theta}^\star) \Big) + 2(\kappa^2 +  2\sigma_\text{bias}^2) + \frac{ \eta L\sigma^2_\text{var}}{P }  \\
    &\quad + L^2 \frac{16\eta^2 K \sigma^2_\text{var} }{(1 - 16 \eta^2 L^2K^2)} + L^2 \frac{16 \eta^2 K^2 \sigma_\text{bias}^2 + 64 \eta^2 K^2 \kappa^2 }{(1 - 16 \eta^2 L^2K^2)} \\
    &\underset{(c)}{\leq} \frac{2}{\eta T} \Big( \mathcal{L}(\bar{\boldsymbol{\theta}}^0)  - \mathcal{L}(\boldsymbol{\theta}^\star) \Big)  + 2(\kappa^2+  2\sigma_\text{bias}^2 )+ \frac{ \eta L\sigma^2_\text{var}}{P }   \\
    &\quad + 32 L^2 \eta^2 K \sigma^2_\text{var}  + 128 L^2  \eta^2 K^2 \kappa^2 +  32 L^2  \eta^2 K^2 \sigma_\text{bias}^2,
    \end{aligned}
\end{equation}
where $(a)$ is due to Lemma~\ref{lemma:global_local_diff}, $(b)$ is due to $0 <\eta < \frac{1}{L}$, and $(c)$ is due to $K \leq \frac{\sqrt{2}}{8 L\eta}$ is selected to satisfy $1 - 16 \eta^2 L^2K^2 \geq \frac{1}{2}$.

If we choose $\eta = \frac{\sqrt{P}}{\sqrt{T}}$ and $K \leq  \frac{\sqrt{2}T^{1/4}}{8L P^{3/4}} = \mathcal{O}(\frac{T^{1/4}}{P^{3/4}})$, then for any $T \geq L^2 P$ we have
\begin{equation} \label{eq:thm1_concludion_appendix}
    \frac{1}{T}\sum_{t=0}^{T-1} \mathbb{E}[ \| \nabla \mathcal{L}(\bar{\boldsymbol{\theta}}^t) \|^2 ]  = \mathcal{O}\left(\frac{1}{\sqrt{PT}}\right) + \mathcal{O}(\kappa^2 + \sigma^2_\text{bias}) ,
\end{equation}
with $R = \mathcal{O}(P^{3/4} T^{3/4})$ rounds of communication.

\subsection{Discussion on the irreducible error}

To better understand why the error in the upper bound might be irreducible (i.e., the second term in RHS of Theorem 1 is independent of T),
let's first recall the key sources that cause  gradient diversity $
\Big\| \nabla \mathcal{L}(\bar{\boldsymbol{\theta}}^t) - \frac{1}{P}\sum_{p=1}^P \mathbb{E}[ \tilde{\nabla} \mathcal{L}_p^\text{local}(\boldsymbol{\theta}_p^t;\xi_p^t)] \Big\|^2 $ (in Eq. 21), in our setting:
\begin{itemize}
    \item The $\kappa^2 $ term, which is the upper bound of $ \frac{1}{P}\sum_{p=1}^P \| \nabla \mathcal{L}(\bar{\boldsymbol{\theta}}^t) - \nabla \mathcal{L}_p^\text{local}(\bar{\boldsymbol{\theta}}^t) \|^2$ and is caused by ignoring the cut-edges for local gradient computation;
    \item The $\sigma_\text{bias}^2$ term, which is the upper bound of $\frac{1}{P}\sum_{p=1}^P \left\| \mathbb{E}[ \tilde{\nabla} \mathcal{L}_p^\text{local}(\bar{\boldsymbol{\theta}}^t;\xi_p^t)] - \nabla \mathcal{L}_p^\text{local}(\boldsymbol{\theta}_p^t) \right\|^2$ and is caused by using neighbor sampling;
    \item The model divergence term $\frac{1}{P}\sum_{p=1}^P \| \bar{\boldsymbol{\theta}}^t - \boldsymbol{\theta}_p^t\|^2$ which is caused by the difference between the model parameters and the virtual average model due to periodic averaging. Notice that the model divergence term also exists in distributed learning regardless of whether the dataset is graph or not and caused by infrequent synchronization.
\end{itemize}

Now, let's have a closer look at the model divergence term. 
As shown in Lemma 3 the model divergence term $\frac{1}{P}\sum_{p=1}^P \| \bar{\boldsymbol{\theta}}^t - \boldsymbol{\theta}_p^t\|^2$ is further caused by three factors:
\begin{itemize}
    \item The $\sigma_\text{var}^2$ term (in Eq. 64),  which is the upper bound of mini-batch sampling variance $\frac{1}{P}\sum_{p=1}^P \mathbb{E} \Big[ \Big\|  \sum_{\tau=t_0}^{t-1} \left( \mathbb{E}[\tilde{\nabla} \mathcal{L}_p^\text{local}(\boldsymbol{\theta}_p^{\tau}; \xi_p^{\tau})] - \nabla \mathcal{L}_p^\text{local}(\boldsymbol{\theta}_p^{\tau})  \right) \Big\|^2 \Big]$ 
    \item The $\kappa^2 $ term, which is the upper bound of $ \frac{1}{P}\sum_{p=1}^P \| \nabla \mathcal{L}(\bar{\boldsymbol{\theta}}^t) - \nabla \mathcal{L}_p^\text{local}(\bar{\boldsymbol{\theta}}^t) \|^2$ and is caused by ignoring the cut-edges for local gradient computation;
    \item The $\sigma_\text{bias}^2$ term, which is the upper bound of $\frac{1}{P}\sum_{p=1}^P \left\| \mathbb{E}[ \tilde{\nabla} \mathcal{L}_p^\text{local}(\bar{\boldsymbol{\theta}}^t;\xi_p^t)] - \nabla \mathcal{L}_p^\text{local}(\boldsymbol{\theta}_p^t) \right\|^2$ and is caused by using neighbor sampling.
\end{itemize}

\begin{figure}
    \centering
    \includegraphics[width=1.0\textwidth]{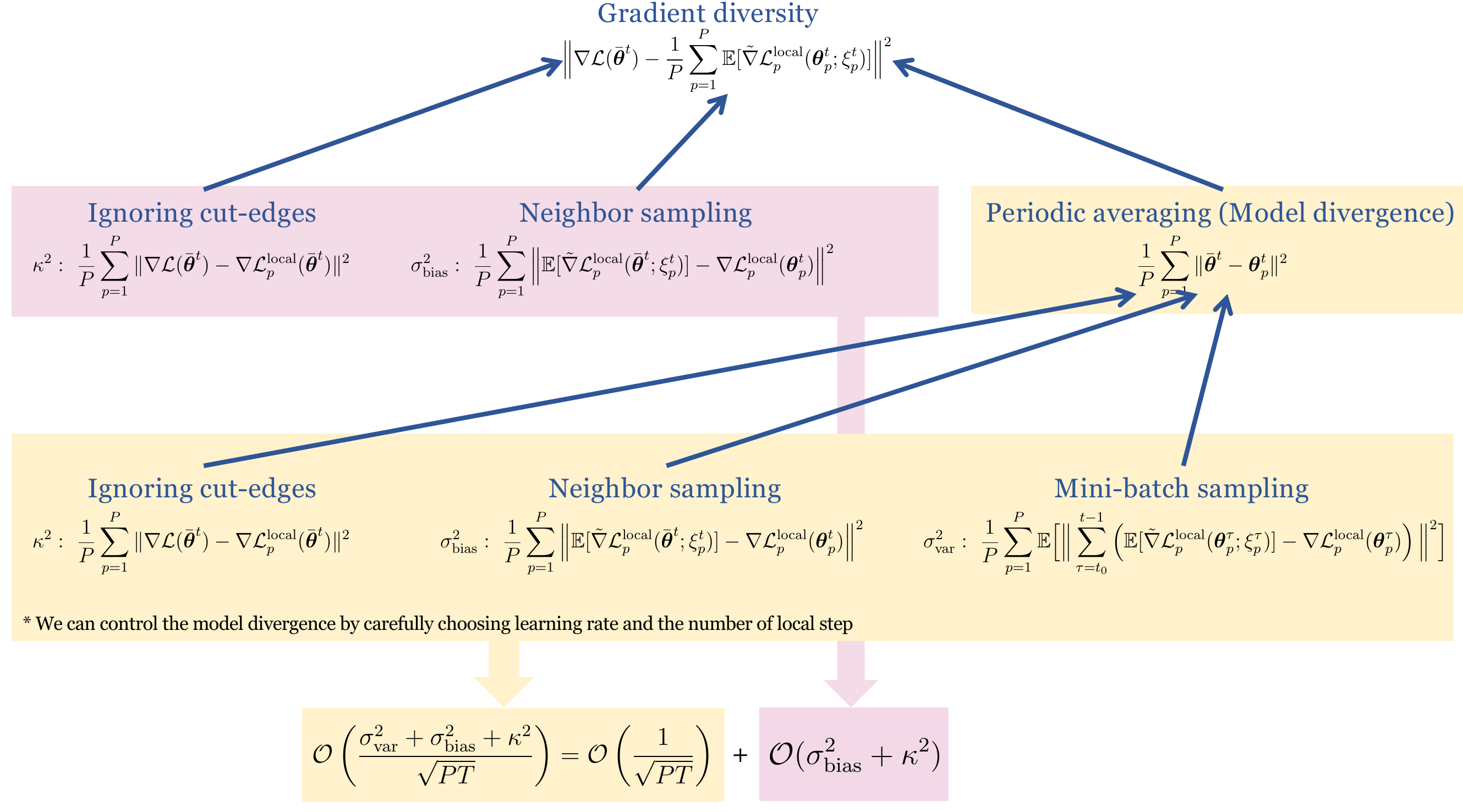}
    \caption{An overview on the existence of irreducible error from the theoretical point of view. We use A $\rightarrow$ B to denote A is causing B.}
    \label{fig:irreducible_error}
\end{figure}

Fortunately, the model divergence term can be controlled by the number of local gradient update steps and learning rate, which is reducing with respect to the number of total training steps $T$ and the number of local machine $P$, and it leads to the first term in our upper bound in Theorem 1, i.e., $\mathcal{O}(\frac{\sigma_\text{var}^2 + \sigma_\text{bias}^2 + \kappa^2}{\sqrt{PT}}) = \mathcal{O}(\frac{1}{\sqrt{PT}})$.  
However, unfortunately, $\kappa^2$ and $\sigma^2_\text{bias}$ in gradient diversity are isolated from the model diversity part, therefore are irreducible and results in $\mathcal{O}(\sigma_\text{bias}^2 + \kappa^2)$ in the second term in our upper bound in Theorem 1. Please refer to Figure~\ref{fig:irreducible_error} for an illustration.

Intuitively, these theoretical results make sense. During training, we are minimizing the loss without cut-edges $\frac{1}{P}\sum_{p=1}^P \mathcal{L}^\text{local}(\boldsymbol{\theta})$, which is a different objective to the loss defined on the full-graph by taking the cut-edges into consideration $\mathcal{L}(\boldsymbol{\theta})$. Therefore, solely by adding the number of training iterations, we cannot guarantee a small gradient of $\mathcal{L}(\boldsymbol{\theta})$ by minimizing $\frac{1}{P}\sum_{p=1}^P \mathcal{L}^\text{local}(\boldsymbol{\theta})$.

\clearpage
\subsection{Proof of Lemma~\ref{lemma:gradient_variance}}

By the definition of $\mathcal{L}_p^\text{local}(\boldsymbol{\theta}_p; \xi_p)$, we have

\begin{equation}
    \begin{aligned}
    &\mathbb{E}\Big[ \Big\| \frac{1}{P} \sum_{p=1}^P \tilde{\nabla} \mathcal{L}_p^\text{local}(\boldsymbol{\theta}_p; \xi_p) \Big\|^2  \Big] \\
    &= \mathbb{E}\left[ \Big\| \frac{1}{P} \sum_{p=1}^P \left( \tilde{\nabla} \mathcal{L}_p^\text{local}(\boldsymbol{\theta}_p; \xi_p) - \mathbb{E} \left[ \tilde{\nabla} \mathcal{L}_p^\text{local}(\boldsymbol{\theta}_p, \xi_p^t)  \right]\right) \Big\|^2  \right] + \mathbb{E}\Big[ \Big\| \frac{1}{P} \sum_{p=1}^P \mathbb{E} \left[ \tilde{\nabla} \mathcal{L}_p^\text{local}(\boldsymbol{\theta}_p, \xi_p^t)  \right]\Big\|^2  \Big] \\
    &\underset{(a)}{\leq} \frac{1}{P^2} \sum_{p=1}^P \mathbb{E}\left[ \Big\| \tilde{\nabla} \mathcal{L}_p^\text{local}(\boldsymbol{\theta}_p; \xi_p) - \mathbb{E}\left[ \tilde{\nabla} \mathcal{L}_p^\text{local}(\boldsymbol{\theta}_p; \xi_p) \right]  \Big\|^2  \right] + \mathbb{E}\left[ \left\| \frac{1}{P} \sum_{p=1}^P \mathbb{E} \left[ \tilde{\nabla} \mathcal{L}_p^\text{local}(\boldsymbol{\theta}_p, \xi_p^t)  \right]\right\|^2  \right] \\
    &\underset{(b)}{\leq} \frac{\sigma^2_\text{var}}{P} + \mathbb{E}\Big[ \Big\| \frac{1}{P} \sum_{p=1}^P \mathbb{E} \left[ \tilde{\nabla} \mathcal{L}_p^\text{local}(\boldsymbol{\theta}_p, \xi_p^t)  \right]\Big\|^2  \Big]
    \end{aligned}
\end{equation}
where $(a)$ is due to fact that each $\tilde{\nabla} \mathcal{L}_p^\text{local}(\boldsymbol{\theta}_p; \xi_p) - \mathbb{E}[\tilde{\nabla} \mathcal{L}_p^\text{local}(\boldsymbol{\theta}_p; \xi_p)]$ is independent random vectors with zero mean and $(b)$ is from Assumption~\ref{assumption:bound_local_variance}.

\subsection{Proof of Lemma~\ref{lemma:local_diff}}
By the definition of local gradient $\nabla \mathcal{L}_p^\text{local}(\boldsymbol{\theta})$, we have
\begin{equation}
    \begin{aligned}
    & \frac{1}{P} \sum_{p=1}^P \Big\| \nabla \mathcal{L}_p^\text{local}(\boldsymbol{\theta}_p) - \frac{1}{P}\sum_{q=1}^P \nabla \mathcal{L}_q^\text{local}(\boldsymbol{\theta}_q) \Big\|^2 \\
    &= \frac{1}{P} \sum_{p=1}^P \Big\| \nabla \mathcal{L}_p^\text{local}(\boldsymbol{\theta}_p) - \nabla \mathcal{L}_p^\text{local}(\bar{\boldsymbol{\theta}}) + \nabla \mathcal{L}_p^\text{local}(\bar{\boldsymbol{\theta}}) - \nabla \mathcal{L}(\bar{\boldsymbol{\theta}}) \\
    &\quad + \nabla \mathcal{L}(\bar{\boldsymbol{\theta}}) - \frac{1}{P}\sum_{q=1}^P \nabla \mathcal{L}(\boldsymbol{\theta}_q) + \frac{1}{P}\sum_{q=1}^P \nabla \mathcal{L}(\boldsymbol{\theta}_q) - \frac{1}{P}\sum_{q=1}^P \nabla \mathcal{L}_q^\text{local}(\boldsymbol{\theta}_q) \Big\|^2 \\
    &\underset{(a)}{\leq} \frac{1}{P} \sum_{p=1}^P 4 \|  \nabla \mathcal{L}_p^\text{local}(\boldsymbol{\theta}_p) - \nabla \mathcal{L}_p^\text{local}(\bar{\boldsymbol{\theta}}) \|^2 + \frac{1}{P} \sum_{p=1}^P 4 \| \nabla \mathcal{L}_p^\text{local}(\bar{\boldsymbol{\theta}}) - \nabla \mathcal{L}(\bar{\boldsymbol{\theta}})  \|^2 \\
    &\quad + \frac{1}{P} \sum_{p=1}^P 4 \Big\| \nabla \mathcal{L}(\bar{\boldsymbol{\theta}}) - \frac{1}{P}\sum_{q=1}^P \nabla \mathcal{L}(\boldsymbol{\theta}_q) \Big\|^2 + \frac{1}{P} \sum_{p=1}^P 4 \Big\| \frac{1}{P}\sum_{q=1}^P \nabla \mathcal{L}(\boldsymbol{\theta}_q) - \frac{1}{P}\sum_{q=1}^P \nabla \mathcal{L}_q^\text{local}(\boldsymbol{\theta}_q) \Big\|^2 \\
    &\underset{(b)}{\leq} \frac{8 L}{P} \sum_{p=1}^P \| \boldsymbol{\theta}_p - \bar{\boldsymbol{\theta}} \|^2 + 8\kappa^2,
    \end{aligned}
\end{equation}
where $(a)$ is due to $\| \sum_{i=1}^n \mathbf{x}_i\|^2 \leq \sum_{i=1}^n n\| \mathbf{x}_i \|$ and $(b)$ is due to the definition of $\kappa$.

\subsection{Proof of Lemma~\ref{lemma:global_local_diff}}
When $(t~\text{mod}~K) = 0$ we have $\bar{\boldsymbol{\theta}}^t = \boldsymbol{\theta}_p^t $. When $(t~\text{mod}~K) \neq 0$ and $t \geq 1$, let $t_0 < t$ be the largest iteration index that $t_0~\text{mod}~K=0$. 
For any $\tau \in \{ t_0 + 1, \ldots, t\}$, we have
\begin{equation}
    \boldsymbol{\theta}_p^\tau - \boldsymbol{\theta}_p^{\tau-1} = -\eta \tilde{\nabla} \mathcal{L}_p^\text{local}(\boldsymbol{\theta}_p^{\tau-1}; \xi_p^{\tau - 1}).
\end{equation}

Summing over $\tau \in \{ t_0 + 1, \ldots, t\}$, we have
\begin{equation} \label{eq:lemma_local_global_eq_1_thm1}
    \begin{aligned}
    \boldsymbol{\theta}_p^t  &= \boldsymbol{\theta}_p^{t_0} -\eta \sum_{\tau=t_0}^{t-1} \tilde{\nabla} \mathcal{L}_p^\text{local}(\boldsymbol{\theta}_p^{\tau}; \xi_p^{\tau}) \\
    & \underset{(a)}{=} \bar{\boldsymbol{\theta}}^{t_0} -\eta \sum_{\tau=t_0}^{t-1} \tilde{\nabla} \mathcal{L}_p^\text{local}(\boldsymbol{\theta}_p^{\tau}; \xi_p^{\tau}),
    \end{aligned}
\end{equation}
where $(a)$ is due to $\bar{\boldsymbol{\theta}}^t = \boldsymbol{\theta}_p^t $ when $(t~\text{mod}~K) = 0$.

Similarly, we have
\begin{equation} \label{eq:lemma_local_global_eq_2_thm1}
    \bar{\boldsymbol{\theta}}^{t} = \bar{\boldsymbol{\theta}}^{t_0} -\eta \sum_{\tau=t_0}^{t-1} \frac{1}{P}\sum_{p=1}^P \tilde{\nabla} \mathcal{L}_p^\text{local}(\boldsymbol{\theta}_p^{\tau}; \xi_p^{\tau}).
\end{equation}

Combining Eq.~\ref{eq:lemma_local_global_eq_1_thm1} and Eq.~\ref{eq:lemma_local_global_eq_2_thm1}, we have
\begin{equation} \label{eq:lemma_local_global_eq_3} 
    \begin{aligned}
    &\frac{1}{P}\sum_{p=1}^P \mathbb{E}[\| \bar{\boldsymbol{\theta}}^{t} - \boldsymbol{\theta}^{t}_p \|^2] \\
    &= \frac{\eta^2}{P}\sum_{p=1}^P \mathbb{E} \Big[ \Big\| \sum_{\tau=t_0}^{t-1} \Big( \tilde{\nabla} \mathcal{L}_p^\text{local}(\boldsymbol{\theta}_p^{\tau}; \xi_p^{\tau}) - \frac{1}{P}\sum_{q=1}^P \tilde{\nabla} \mathcal{L}_q^\text{local}(\boldsymbol{\theta}_q^{\tau}; \xi_q^{\tau}) \Big) \Big\|^2 \Big] \\
    &\underset{(a)}{\leq} 2 \eta^2 \underbrace{\frac{1}{P}\sum_{p=1}^P \mathbb{E} \Big[ \Big\| \sum_{\tau=t_0}^{t-1} \Big( \big( \tilde{\nabla} \mathcal{L}_p^\text{local}(\boldsymbol{\theta}_p^{\tau}; \xi_p^{\tau}) - \nabla \mathcal{L}_p^\text{local}(\boldsymbol{\theta}_p^{\tau}) \big) - \frac{1}{P}\sum_{q=1}^P \big( \tilde{\nabla} \mathcal{L}_q^\text{local}(\boldsymbol{\theta}_q^{\tau}; \xi_q^{\tau}) - \nabla \mathcal{L}_q^\text{local}(\boldsymbol{\theta}_q^{\tau}) \big) \Big) \Big\|^2 \Big]}_{(A)} \\
    &\quad + 2 \eta^2 \underbrace{\frac{1}{P}\sum_{p=1}^P \mathbb{E} \Big[ \Big\| \sum_{\tau=t_0}^{t-1} \Big( \nabla \mathcal{L}_p^\text{local}(\boldsymbol{\theta}_p^{\tau}) - \frac{1}{P}\sum_{q=1}^P \nabla \mathcal{L}_q^\text{local}(\boldsymbol{\theta}_q^{\tau}) \Big) \Big\|^2 \Big]}_{(B)},
    \end{aligned}
\end{equation}
where $(a)$ is due to addition and subtraction of $\tilde{\nabla} \mathcal{L}_p^\text{local}(\boldsymbol{\theta}_p^{\tau}; \xi_p^{\tau})$ and $\nabla \mathcal{L}_p^\text{local}(\boldsymbol{\theta}_p^{\tau})$ and using $\|\mathbf{x} + \mathbf{y}\|^2 \leq \| \mathbf{x}\|^2 + \| \mathbf{y}\|^2$.

We can upper bound $(A)$ in Eq.~\ref{eq:lemma_local_global_eq_3} by
\begin{equation} \label{eq:lemma_local_global_eq_4}
    \begin{aligned}
    (A) &\underset{(a)}{\leq} \frac{1}{P}\sum_{p=1}^P \mathbb{E} \Big[ \Big\| \sum_{\tau=t_0}^{t-1} \big( \tilde{\nabla} \mathcal{L}_p^\text{local}(\boldsymbol{\theta}_p^{\tau}; \xi_p^{\tau}) - \nabla \mathcal{L}_p^\text{local}(\boldsymbol{\theta}_p^{\tau}) \big)  \Big\|^2 \Big] \\
    &\leq \frac{2}{P}\sum_{p=1}^P  \mathbb{E} \Big[ \Big\|  \sum_{\tau=t_0}^{t-1} \Big( \tilde{\nabla} \mathcal{L}_p^\text{local}(\boldsymbol{\theta}_p^{\tau}; \xi_p^{\tau}) - \mathbb{E}[\tilde{\nabla} \mathcal{L}_p^\text{local}(\boldsymbol{\theta}_p^{\tau}; \xi_p^{\tau})] \Big) \Big\|^2 \Big] \\
    &\quad + \frac{2}{P}\sum_{p=1}^P  \mathbb{E} \Big[ \Big\|  \sum_{\tau=t_0}^{t-1} \left( \mathbb{E}[\tilde{\nabla} \mathcal{L}_p^\text{local}(\boldsymbol{\theta}_p^{\tau}; \xi_p^{\tau})] - \nabla \mathcal{L}_p^\text{local}(\boldsymbol{\theta}_p^{\tau})  \right) \Big\|^2 \Big] \\
    &\leq 2K\sigma^2_\text{var} + 2K^2\sigma^2_\text{bias},
    \end{aligned}
\end{equation}
where $(a)$ is due to 
\begin{equation} 
    \frac{1}{n}\sum_{i=1}^n \Big\| \mathbf{x}_i - \frac{1}{n}\sum_{j=1}^n \mathbf{x}_j \Big\|^2 = \frac{1}{n}\sum_{i=1}^n \| \mathbf{x}_i \|^2 - \Big\| \frac{1}{n}\sum_{i=1}^n \mathbf{x}_i \Big\|^2 \leq \frac{1}{n}\sum_{i=1}^n \| \mathbf{x}_i \|^2.
\end{equation}

We can further bound $(B)$ in Eq.~\ref{eq:lemma_local_global_eq_3} by
\begin{equation} \label{eq:lemma_local_global_eq_5}
    \begin{aligned}
    (B) &\leq \frac{1}{P}\sum_{p=1}^P (t-t_0) \sum_{\tau=t_0}^{t-1} \mathbb{E} \Big[ \Big\| \nabla \mathcal{L}_p^\text{local}(\boldsymbol{\theta}_p^{\tau}) - \frac{1}{P}\sum_{q=1}^P \nabla \mathcal{L}_q^\text{local}(\boldsymbol{\theta}_q^{\tau}) \Big\|^2 \Big] \\
    &\leq K \sum_{\tau=t_0}^{t-1} \frac{1}{P} \sum_{p=1}^P \mathbb{E} \Big[ \Big\| \nabla \mathcal{L}_p^\text{local}(\boldsymbol{\theta}_p^{\tau}) - \frac{1}{P}\sum_{q=1}^P \nabla \mathcal{L}_q^\text{local}(\boldsymbol{\theta}_q^{\tau}) \Big\|^2 \Big] \\
    &\underset{(a)}{\leq} K \sum_{\tau=t_0}^{t-1} \Big( \frac{8L^2}{P}\sum_{i=1}^P \| \boldsymbol{\theta}_p^\tau - \bar{\boldsymbol{\theta}}^\tau \|^2 + 8 \kappa^2  \Big) \\
    &\leq \frac{8 L^2K}{P} \sum_{\tau=t_0}^{t-1} \sum_{i=1}^P \| \boldsymbol{\theta}_p^\tau - \bar{\boldsymbol{\theta}}^\tau \|^2 + 8 K^2 \kappa^2,
    \end{aligned}
\end{equation}
where $(a)$ is due to Lemma~\ref{lemma:local_diff} and the definition of $\kappa$.

By plugging Eq.~\ref{eq:lemma_local_global_eq_4} and Eq.~\ref{eq:lemma_local_global_eq_5} into Eq.~\ref{eq:lemma_local_global_eq_3}, we have
\begin{equation}\label{eq:lemma_local_global_eq_6}
    \frac{1}{P}\sum_{p=1}^P \mathbb{E}[\| \bar{\boldsymbol{\theta}}^{t} - \boldsymbol{\theta}^{t}_p \|^2] 
    \leq 4 \eta^2 K \sigma^2_\text{var} + 4 \eta^2 K^2 \sigma^2_\text{bias} +  16 \eta^2 L^2K \sum_{\tau=t_0}^{t-1} \frac{1}{P} \sum_{i=1}^P \| \boldsymbol{\theta}_p^\tau - \bar{\boldsymbol{\theta}}^\tau \|^2 + 16 \eta^2 K^2 \kappa^2 .
\end{equation}

By summing over $t \in \{ 0, \ldots, T\}$, we have
\begin{equation}
    \begin{aligned}
    \underbrace{\sum_{t=0}^{T-1} \frac{1}{P}\sum_{p=1}^P \mathbb{E}[\| \bar{\boldsymbol{\theta}}^{t} - \boldsymbol{\theta}^{t}_p \|^2]}_{(A)} 
    &\leq 4\eta^2 K \sigma^2_\text{var} T + 4\eta^2 K^2 \sigma^2_\text{bias} T \\
    &\quad + 16 \eta^2 L^2K^2 \underbrace{\sum_{t=0}^{T-1} \frac{1}{P} \sum_{i=1}^P \| \boldsymbol{\theta}_p^t - \bar{\boldsymbol{\theta}}^t \|^2}_{(A)} + 16 \eta^2 K^2 \kappa^2 T.
    \end{aligned}
\end{equation}

Collecting common terms and dividing both sides by $(1 - 16 \eta^2 L^2K^2)$ gives
\begin{equation}
    (A) \leq \frac{4\eta^2 K \sigma^2 T}{(1 - 16 \eta^2 L^2K^2)} + \frac{4\eta^2 K^2 \sigma^2_\text{bias} T + 16\eta^2 K^2 \kappa^2 T}{(1 - 16 \eta^2 L^2K^2)}.
\end{equation}


\clearpage
\section{Proof of Theorem~\ref{theorem:distgnn_global_correction}}

In the following,
we first introduce the useful lemmas in Section~\ref{section:thm2_useful_lemmas}, then process our proof of the theorem in Section~\ref{section:thm2_main}.
In particular, we show that this residual error (which is caused by ignoring the cut-edges) can be eliminated by running server correction steps after the parameter averaging on the server.

\subsection{Useful lemma} \label{section:thm2_useful_lemmas}

The following lemma provides an upper bound on the deviation of each local model to virtual averaged model, which is important to upper bound on the difference of each local gradient $\nabla \mathcal{L}_p^\text{local}(\bm{\theta}_p)$ compared to $\frac{1}{P}\sum_{p=1}^P \nabla \mathcal{L}_p (\bm{\theta})$.

\begin{lemma} \label{lemma:global_local_diff_thm2}
For all $P$ machines with $\bm{\theta}_p,~p\in\{1,\ldots,P\}$, if we define $\bar{\bm{\theta}} = \frac{1}{P}\sum_{p=1}^P \bm{\theta}_p$ and let $\kappa>0$ such that $\| \mathcal{L}_p^\text{local}(\bm{\theta}) - \nabla \mathcal{L}(\bm{\theta}) \|^2 \leq \kappa^2$, we have
\begin{equation}
    \sum_{t\in \mathcal{T}_\text{local}(r)}\frac{1}{P}\sum_{p=1}^P \mathbb{E}[\| \bar{\bm{\theta}}^{t} - \bm{\theta}^{t}_p \|^2] \leq \frac{4 \eta^2 (\sigma^2_\text{bias} + \sigma^2_\text{var}) K^2 \rho^{2r}  + 16 \eta^2 \kappa^2  K^2 \rho^{2r}}{1-16 \eta^2 L^2 K^2 \rho^{2r}}.
\end{equation}
\end{lemma}

\subsection{Main proof of Theorem~\ref{theorem:distgnn_global_correction}} \label{section:thm2_main}
Equipped with the above lemma and results from Appendix~\ref{section:thm1_useful_lemmas} , we are now ready to present the proof of Theorem~\ref{theorem:distgnn_global_correction}.

Let $\mathcal{T}_\text{local}(r)$ and $\mathcal{T}_\text{server}(r)$ as the iteration indices that a local machine and server run on, which is defined as
\begin{equation}\label{eq:T_local_server} 
    \begin{aligned}
    \mathcal{T}_\text{local}(r) &= \left\{ k + \left(\sum_{j=1}^{r-1} K \rho^j \right) 
    + S(r-1): k = 1,\ldots, K\rho^r \right\} \\
    \mathcal{T}_\text{server}(r) &= \left\{ s + 
    \left(\sum_{j=1}^{r} K \rho^j \right) + S(r-1): s = 1,\ldots, S \right\},
    \end{aligned}
\end{equation}
and let define $\mathcal{T}_\text{local} = \mathcal{T}_\text{local}(1) \cup \ldots \cup \mathcal{T}_\text{local} (R)$ and $\mathcal{T}_\text{global} = \mathcal{T}_\text{global}(1) \cup \ldots \cup \mathcal{T}_\text{global} (R)$.

By the smoothness assumption, we have
\begin{equation} \label{eq:thm2_proof_1}
    \mathbb{E}[\mathcal{L}(\bar{\bm{\theta}}^{t+1})] \leq \mathbb{E}[\mathcal{L}(\bar{\bm{\theta}}^t)] + \mathbb{E}[ \langle \nabla \mathcal{L}(\bar{\bm{\theta}}^t), \bar{\bm{\theta}}^{t+1} - \bar{\bm{\theta}}^t  \rangle ] + \frac{L}{2} \mathbb{E}[\| \bar{\bm{\theta}}^{t+1} - \bar{\bm{\theta}}^t \|^2].
\end{equation}

Let first consider $t \in \mathcal{T}_\text{server}(r),$ with the following update 
\begin{equation} \label{eq:thm2_proof_4_add_new_1}
    \begin{aligned}
    \bar{\bm{\theta}}^{t+1} 
    &= \bar{\bm{\theta}}^t - \gamma \tilde{\nabla} \mathcal{L}(\bar{\bm{\theta}}^t; \xi^t)\\
    &= \bar{\bm{\theta}}^t -  \frac{\gamma}{P}\sum_{p=1}^P \tilde{\nabla} \mathcal{L}_p^\text{full}(\bar{\bm{\theta}}^t; \xi_p^t).
    \end{aligned}
\end{equation}

Therefore, we have
\begin{equation} \label{eq:thm2_proof_4}
    \begin{aligned}
    \mathbb{E}[ \langle \nabla \mathcal{L}(\bar{\bm{\theta}}^t), \bar{\bm{\theta}}^{t+1} - \bar{\bm{\theta}}^t  \rangle ] 
    &= - \gamma \mathbb{E} \Big[ \Big\langle \nabla \mathcal{L}(\bar{\bm{\theta}}^t), \frac{1}{P}\sum_{p=1}^P \tilde{\nabla} \mathcal{L}_p^\text{full}(\bar{\bm{\theta}}^t; \xi_p^t)  \Big\rangle \Big] \\
    &\underset{(a)}{=} - \gamma \mathbb{E}[ \| \nabla \mathcal{L}(\bar{\bm{\theta}}^t) \|^2],
    \end{aligned}
\end{equation}
where the equality is due to $\frac{1}{P}\sum_{p=1}^P \mathbb{E}[ \tilde{\nabla} \mathcal{L}_p^\text{full}(\bar{\bm{\theta}}^t; \xi_p^t)] = \nabla \mathcal{L}(\bar{\bm{\theta}}^t) $ since all neighbors are used for the server correction steps.

Besides, by taking the norm on the both side of Eq.~\ref{eq:thm2_proof_4_add_new_1}, we  have the following equality
\begin{equation} \label{eq:thm2_proof_5}
    \begin{aligned}
    \mathbb{E}[ \| \bar{\bm{\theta}}^{t+1} - \bar{\bm{\theta}}^t \|^2] 
    &=  \gamma^2 \mathbb{E}\Big[ \Big\| \frac{1}{P}\sum_{p=1}^P \tilde{\nabla} \mathcal{L}_p^\text{full}(\bar{\bm{\theta}}^t; \xi_p^t) \Big\|^2 \Big] \\
    &\underset{(a)}{\leq} \gamma^2 \mathbb{E}\Big[ \Big\| \frac{1}{P}\sum_{p=1}^P \nabla \mathcal{L}_p^\text{full}(\bar{\bm{\theta}}^t) \Big\|^2 \Big] + \gamma^2 \frac{\sigma_\text{global}^2}{P} \\
    &= \gamma^2 \mathbb{E}[\| \nabla \mathcal{L}(\bar{\bm{\theta}}^t )\|^2 ] + \gamma^2 \frac{\sigma_\text{global}^2}{P} .
    \end{aligned}
\end{equation}
where inequality $(b)$ holds due to Lemma~\ref{lemma:gradient_variance}. Notice that the result in Lemma~\ref{lemma:gradient_variance} holds for both the ``local'' and the ``global'' setting since $\frac{1}{P}\sum_{p=1}^P \mathbb{E}[ \tilde{\nabla} \mathcal{L}_p^\text{full}(\bar{\bm{\theta}}^t; \xi_p^t)] = \nabla \mathcal{L}(\bar{\bm{\theta}}^t) $.

Substituting Eq.~\ref{eq:thm2_proof_4},~\ref{eq:thm2_proof_5} into Eq.~\ref{eq:thm2_proof_1}, we know that for $t\in\mathcal{T}_\text{server}(r),$ we have 

\begin{equation} \label{eq:thm2_proof_7}
    \begin{aligned}
    \mathbb{E}[\mathcal{L}(\bar{\bm{\theta}}^{t+1})] &\leq \mathbb{E}[\mathcal{L}(\bar{\bm{\theta}}^t)]  + \Big( \frac{\gamma^2 L}{2} - \gamma \Big) \mathbb{E}[ \| \nabla \mathcal{L}(\bar{\bm{\theta}}^t) \|^2] + \frac{\gamma^2 L}{2}\frac{\sigma_\text{global}^2}{P} \\
    &=\mathbb{E}[\mathcal{L}(\bar{\bm{\theta}}^t)]  + \frac{\gamma}{2} \Big( \gamma L - 1 \Big) \mathbb{E}[ \| \nabla \mathcal{L}(\bar{\bm{\theta}}^t) \|^2] + \frac{\gamma^2 L}{2}\frac{\sigma_\text{global}^2}{P} - \frac{\gamma}{2} \mathbb{E}[ \| \nabla \mathcal{L}(\bar{\bm{\theta}}^t) \|^2].
    \end{aligned}
\end{equation}

Dividing both sides by $\frac{\gamma}{2}$ and rearranging terms yields
\begin{equation} \label{eq:thm2_proof_8}
    \mathbb{E}[ \| \nabla \mathcal{L}(\bar{\bm{\theta}}^t) \|^2] \leq \frac{2}{\gamma} \Big( \mathbb{E}[\mathcal{L}(\bar{\bm{\theta}}^t)] - \mathbb{E}[\mathcal{L}(\bar{\bm{\theta}}^{t+1})] \Big) + (\gamma L - 1) \mathbb{E}[ \| \nabla \mathcal{L}(\bar{\bm{\theta}}^t) \|^2] + \gamma L \frac{\sigma_\text{global}^2}{P}.
\end{equation}

Then, let us consider the local update steps where $t\in\mathcal{T}_\text{local}(r)$.
According to Eq.~\ref{eq:thm1_proof_7} in proof of Theorem~\ref{theorem:distgnn_param_avg}, we have
\begin{equation} \label{eq:thm2_proof_9}
    \begin{aligned}
    \mathbb{E}[ \| \nabla \mathcal{L}(\bar{\bm{\theta}}^t) \|^2 ] &\leq \frac{2}{\eta} \Big( \mathbb{E}[\mathcal{L}(\bar{\bm{\theta}}^t)]  - \mathbb{E}[\mathcal{L}(\bar{\bm{\theta}}^{t+1})] \Big) +(\eta L - 1 ) \mathbb{E}\Big[ \Big\| \frac{1}{P}\sum_{p=1}^P \nabla \mathcal{L}_p^\text{local}(\bm{\theta}_p^t) \Big\|^2 \Big] \\
    &\quad + 2(\kappa^2 + 2\sigma_\text{bias}^2) + \frac{ 4 L^2}{ P} \sum_{p=1}^P \mathbb{E}[ \| \bar{\bm{\theta}}^t - \bm{\theta}_p^t \|^2 ] + \frac{ \eta L\sigma_\text{var}^2}{P}.
    \end{aligned}
\end{equation}

Let $T = 
\left(\sum_{r=1}^{R} K \rho^r \right)
+ SR$ and summing over $t \in \{ 1, \ldots, T\}$, combining Eq.~\ref{eq:thm2_proof_8} and Eq.~\ref{eq:thm2_proof_9} we have

\begin{equation}
    \begin{aligned}
    &\sum_{t=1}^T \mathbb{E}[ \| \nabla \mathcal{L}(\bar{\bm{\theta}}^t) \|^2 ] \\
    &= \sum_{r=1}^R \sum_{t\in \mathcal{T}_\text{local}(r)} \mathbb{E}[ \| \nabla \mathcal{L}(\bar{\bm{\theta}}^t) \|^2 ] + \sum_{r=1}^R \sum_{t\in \mathcal{T}_\text{global}(r)} \mathbb{E}[ \| \nabla \mathcal{L}(\bar{\bm{\theta}}^t) \|^2 ] \\
    &\leq \sum_{r=1}^R \sum_{t\in \mathcal{T}_\text{global}(r)} \Big[ \frac{2}{\gamma} \Big( \mathbb{E}[\mathcal{L}(\bar{\bm{\theta}}^t)] - \mathbb{E}[\mathcal{L}(\bar{\bm{\theta}}^{t+1})] \Big) + (\gamma L - 1) \mathbb{E}[ \| \nabla \mathcal{L}(\bar{\bm{\theta}}^t) \|^2] + \gamma L \frac{\sigma_\text{global}^2}{P} \Big] \\
    &\quad + \sum_{r=1}^R \sum_{t\in \mathcal{T}_\text{local}(r)} \Big[ \frac{2}{\eta} \Big( \mathbb{E}[\mathcal{L}(\bar{\bm{\theta}}^t)]  - \mathbb{E}[\mathcal{L}(\bar{\bm{\theta}}^{t+1})] \Big) +(\eta L - 1 ) \mathbb{E}\Big[ \Big\| \frac{1}{P}\sum_{p=1}^P \nabla \mathcal{L}_p^\text{local}(\bm{\theta}_p^t) \Big\|^2 \Big] \\
    &\qquad + 2(\kappa^2 + \sigma_\text{bias}^2) + \frac{ 4 L^2}{ P} \sum_{p=1}^P \mathbb{E}[ \| \bar{\bm{\theta}}^t - \bm{\theta}_p^t \|^2 ] + \frac{ \eta L\sigma^2_\text{var}}{P} \Big] .
    \end{aligned}
\end{equation}

Rearrange the above equation, we have
\begin{equation}
    \begin{aligned}
    &\sum_{t=1}^T \mathbb{E}[ \| \nabla \mathcal{L}(\bar{\bm{\theta}}^t) \|^2 ] \\
    &\leq \sum_{r=1}^R \left[ \sum_{t\in \mathcal{T}_\text{global}(r)} \frac{2}{\gamma} \Big( \mathbb{E}[\mathcal{L}(\bar{\bm{\theta}}^t)] - \mathbb{E}[\mathcal{L}(\bar{\bm{\theta}}^{t+1})] \Big) + \sum_{t\in \mathcal{T}_\text{local}(r)} \frac{2}{\eta} \Big( \mathbb{E}[\mathcal{L}(\bar{\bm{\theta}}^t)]  - \mathbb{E}[\mathcal{L}(\bar{\bm{\theta}}^{t+1})] \Big) \right] \\
    &\quad + \sum_{r=1}^R \left[ \sum_{t\in \mathcal{T}_\text{global}(r)} (\gamma L-1) \mathbb{E}[\| \nabla \mathcal{L}(\bar{\bm{\theta}}^t)\|^2] + \sum_{t\in \mathcal{T}_\text{local}(r)} (\eta L - 1) \mathbb{E}\Big[ \Big\| \frac{1}{P}\sum_{p=1}^P \nabla \mathcal{L}_p^\text{local}(\bm{\theta}_p^t) \Big\|^2 \right] \\
    &\quad + \sum_{r=1}^R \left[ \sum_{t\in \mathcal{T}_\text{global}(r)} \gamma L \frac{\sigma_\text{global}^2}{P} + \sum_{t\in \mathcal{T}_\text{local}(r)} \frac{\eta L \sigma_\text{var}^2}{P} \right] + 
    \left(\sum_{r=1}^{R} K \rho^r \right)
    (\kappa^2 + 2\sigma_\text{var}^2)\\
    &\quad + \sum_{r=1}^R \sum_{t\in \mathcal{T}_\text{local}(r)} \left( \frac{ 4 L^2}{ P} \sum_{p=1}^P \mathbb{E}[ \| \bar{\bm{\theta}}^t - \bm{\theta}_p^t \|^2 ] \right).
    \end{aligned}
\end{equation}

Let define 
\begin{equation}
    G_\text{global}^r = \min_{t \in \mathcal{T}_\text{global}(r)} \mathbb{E}[ \| \nabla \mathcal{L}(\bar{\bm{\theta}}^t) \|^2],~
    G_\text{local}^r = \min_{t \in \mathcal{T}_\text{local}(r)} \mathbb{E}\Big[ \Big\| \frac{1}{P}\sum_{p=1}^P \nabla \mathcal{L}_p^\text{local}(\bm{\theta}_p^t) \Big\|^2.
\end{equation}
Our goal is to select the size of $\mathcal{T}_\text{local}(r)$ and $\mathcal{T}_\text{global}(r)$ such that the following inequality holds
\begin{equation}
    \begin{aligned}
    \left( K \rho^r \right)
    (\kappa^2 + 2\sigma_\text{bias}^2 ) 
    &\leq (1-\gamma L) \sum_{t\in \mathcal{T}_\text{global}(r)} \mathbb{E}[\| \nabla \mathcal{L}(\bar{\bm{\theta}}^t)\|^2] + (1-\eta L) \sum_{t\in \mathcal{T}_\text{local}(r)} \mathbb{E}\Big[ \Big\| \frac{1}{P}\sum_{p=1}^P \nabla \mathcal{L}_p^\text{local}(\bm{\theta}_p^t) \Big\|^2 \\
    &\leq (1-\gamma L) S G_\text{global}^r + (1-\eta L) K\rho^r G_\text{local}^r
    \end{aligned}
\end{equation}

After rearranging it, we will have
\begin{equation}\label{eq:proof_of_thm2_eq2}
    S \geq 
    \frac{K \rho^r }{G^r_\text{global} (1-\gamma L)} 
    \left(\kappa^2 + 2\sigma_\text{bias}^2 - (1-\eta L) G_\text{local}^r \right).
\end{equation}

Suppose Eq.~\ref{eq:proof_of_thm2_eq2} holds, we have
\begin{equation}
    \begin{aligned}
    &\sum_{t=1}^T \mathbb{E}[ \| \nabla \mathcal{L}(\bar{\bm{\theta}}^t) \|^2 ] \\
    &\leq \sum_{r=1}^R \left[ \sum_{t\in \mathcal{T}_\text{global}(r)} \frac{2}{\gamma} \Big( \mathbb{E}[\mathcal{L}(\bar{\bm{\theta}}^t)] - \mathbb{E}[\mathcal{L}(\bar{\bm{\theta}}^{t+1})] \Big) + \sum_{t\in \mathcal{T}_\text{local}(r)} \frac{2}{\eta} \Big( \mathbb{E}[\mathcal{L}(\bar{\bm{\theta}}^t)]  - \mathbb{E}[\mathcal{L}(\bar{\bm{\theta}}^{t+1})] \Big) \right] \\
    &\quad + \sum_{r=1}^R \left[ \sum_{t\in \mathcal{T}_\text{global}(r)} \gamma L \frac{\sigma_\text{global}^2}{P} + \sum_{t\in \mathcal{T}_\text{local}(r)} \frac{\eta L \sigma_\text{var}^2}{P} \right] + \sum_{r=1}^R \sum_{t\in \mathcal{T}_\text{local}(r)} \left( \frac{ 4 L^2}{ P} \sum_{p=1}^P \mathbb{E}[ \| \bar{\bm{\theta}}^t - \bm{\theta}_p^t \|^2 ] \right) \\
    &\underset{(a)}{\leq} \sum_{r=1}^R \left[ \sum_{t\in \mathcal{T}_\text{global}(r)} \frac{2}{\gamma} \Big( \mathbb{E}[\mathcal{L}(\bar{\bm{\theta}}^t)] - \mathbb{E}[\mathcal{L}(\bar{\bm{\theta}}^{t+1})] \Big) + \sum_{t\in \mathcal{T}_\text{local}(r)} \frac{2}{\eta} \Big( \mathbb{E}[\mathcal{L}(\bar{\bm{\theta}}^t)]  - \mathbb{E}[\mathcal{L}(\bar{\bm{\theta}}^{t+1})] \Big) \right] \\
    &\quad + \sum_{r=1}^R \left[ \sum_{t\in \mathcal{T}_\text{global}(r)} \gamma L \frac{\sigma_\text{global}^2}{P} + \sum_{t\in \mathcal{T}_\text{local}(r)} \frac{\eta L \sigma_\text{var}^2}{P} \right]  \\
    &\quad + 4L^2 \sum_{r=1}^R \left( \frac{4\eta^2 (\sigma^2_\text{bias} + \sigma^2_\text{var}) K^2 \rho^{2r}  }{1-16 \eta^2 L^2 K^2 \rho^{2r}} + \frac{ 16 \eta^2 \kappa^2  K^2 \rho^{2r}}{1-16 \eta^2 L^2 K^2 \rho^{2r}}\right),
    \end{aligned}
\end{equation}
where $(a)$ is due to Lemma~\ref{lemma:global_local_diff_thm2}.

By selecting $K\rho^r$ such that $1-16\eta^2 L^2 K^2 \rho^{2r} \geq \frac{1}{2}$, we have $K \rho^r \leq \frac{\sqrt{2}}{8 L \eta}$ and
\begin{equation}
    \begin{aligned}
    &\frac{1}{T} \sum_{t=1}^T \mathbb{E}[ \| \nabla \mathcal{L}(\bar{\bm{\theta}}^t) \|^2 ] \\
    &=\frac{1}{T} \sum_{r=1}^R \left[ \sum_{t\in \mathcal{T}_\text{global}(r)} \frac{2}{\gamma} \Big( \mathbb{E}[\mathcal{L}(\bar{\bm{\theta}}^t)] - \mathbb{E}[\mathcal{L}(\bar{\bm{\theta}}^{t+1})] \Big) +  \sum_{t\in \mathcal{T}_\text{global}(r)} \gamma L \frac{\sigma_\text{global}^2}{P} \right] \\
    &\quad + \frac{1}{T} \sum_{r=1}^R \left[ \sum_{t\in \mathcal{T}_\text{local}(r)} \frac{2}{\eta} \Big( \mathbb{E}[\mathcal{L}(\bar{\bm{\theta}}^t)]  - \mathbb{E}[\mathcal{L}(\bar{\bm{\theta}}^{t+1})] \Big) + \sum_{t\in \mathcal{T}_\text{local}(r)} \frac{\eta L \sigma_\text{var}^2}{P} \right]  \\
    &\quad + \frac{4 \eta^2 L^2 \left(8 (\sigma_\text{var}^2 + \sigma_\text{bias}^2)  + 32 \kappa^2 \right) }{T} \sum_{r=1}^R K^2 \rho^{2r} .
    \end{aligned}
\end{equation}

Notice that the condition $K \rho^r \leq \frac{\sqrt{2}}{8 L \eta}$ implies 
\begin{equation}
    \sum_{r=1}^R K^2 \rho^{2r} = K^2 \frac{1-\rho^{2r}}{1-\rho^2} \leq \frac{R}{32 L^2 \eta^2}.
\end{equation}

To this end, by selecting $\eta =\gamma = \frac{\sqrt{P}}{\sqrt{T}}$, $\sum_{r=1}^R K^2 \rho^{2r} \leq \frac{R T^{1/2}}{32 L^2 P^{3/2}}$, and $\sigma^2 = \max\{ \sigma_\text{var}^2, \sigma^2_\text{global}\}$, we have
\begin{equation}
    \frac{1}{T}\sum_{t=1}^T \mathbb{E}[ \| \nabla \mathcal{L}(\bar{\bm{\theta}}^t) \|^2 ] 
    \leq \frac{2}{\sqrt{PT}} \Big( \mathcal{L}(\bar{\bm{\theta}}^0) - \mathcal{L}(\bar{\bm{\theta}}^\star) \Big) + \frac{L\sigma^2}{\sqrt{PT}} + \frac{4L^2}{\sqrt{PT}} \sum_{r=1}^R (8(\sigma^2_\text{bias} + \sigma^2_\text{var}) + 32\kappa^2) ,
\end{equation}
and 
\begin{equation} 
    S \geq \left( \frac{ \kappa^2 + 2\sigma_\text{bias}^2 }{1- L (\sqrt{P/T})} - G_\text{local}^r \right) 
    \frac{ K \rho^r}{ G_\text{global}^r}.
\end{equation}

\subsection{Proof of Lemma~\ref{lemma:global_local_diff_thm2}}
Recall that when $t \in \mathcal{T}_\text{global}(r)$ we have $\bar{\bm{\theta}}^t = \bm{\theta}_p^t $. 
For any $t \in \mathcal{T}_\text{local}(r)$, let $t_0+1$ denote as the first indices of $\mathcal{T}_\text{local}(r)$ and $|\mathcal{T}_\text{local}(r)| = K\rho^r$.
Then, for any $\tau \in \{ t_0 + 1, \ldots, t\}$, we have
\begin{equation}
    \bm{\theta}_p^\tau - \bm{\theta}_p^{\tau-1} = -\eta \tilde{\nabla} \mathcal{L}_p^\text{local}(\bm{\theta}_p^{\tau-1}; \xi_p^{\tau - 1})
\end{equation}

Summing over $\tau \in \{ t_0 + 1, \ldots, t\}$, we have
\begin{equation} \label{eq:lemma_local_global_eq_1}
    \begin{aligned}
    \bm{\theta}_p^t  &= \bm{\theta}_p^{t_0} -\eta \sum_{\tau=t_0}^{t-1} \tilde{\nabla} \mathcal{L}_p^\text{local}(\bm{\theta}_p^{\tau}; \xi_p^{\tau}) \\
    & \underset{(a)}{=} \bar{\bm{\theta}}^{t_0} -\eta \sum_{\tau=t_0}^{t-1} \tilde{\nabla} \mathcal{L}_p^\text{local}(\bm{\theta}_p^{\tau}; \xi_p^{\tau}),
    \end{aligned}
\end{equation}
where $(a)$ is due to $\bar{\bm{\theta}}^t = \bm{\theta}_p^t $ when $t \in \mathcal{T}_\text{global}(r)$.

Similarly, we have
\begin{equation} \label{eq:lemma_local_global_eq_2}
    \bar{\bm{\theta}}^{t} = \bar{\bm{\theta}}^{t_0} -\eta \sum_{\tau=t_0}^{t-1} \frac{1}{P}\sum_{p=1}^P \tilde{\nabla} \mathcal{L}_p^\text{local}(\bm{\theta}_p^{\tau}; \xi_p^{\tau}).
\end{equation}

\clearpage

By combining Eq.~\ref{eq:lemma_local_global_eq_1} and Eq.~\ref{eq:lemma_local_global_eq_2}, we have
\begin{equation} \label{eq:lemma_local_global_eq_3} 
    \begin{aligned}
    &\frac{1}{P}\sum_{p=1}^P \mathbb{E}[\| \bar{\bm{\theta}}^{t} - \bm{\theta}^{t}_p \|^2] \\
    &= \frac{\eta^2}{P}\sum_{p=1}^P \mathbb{E} \Big[ \Big\| \sum_{\tau=t_0}^{t-1} \Big( \tilde{\nabla} \mathcal{L}_p^\text{local}(\bm{\theta}_p^{\tau}; \xi_p^{\tau}) - \frac{1}{P}\sum_{q=1}^P \tilde{\nabla} \mathcal{L}_q^\text{local}(\bm{\theta}_q^{\tau}; \xi_q^{\tau}) \Big) \Big\|^2 \Big] \\
    &\underset{(a)}{\leq} 2 \eta^2 \underbrace{\frac{1}{P}\sum_{p=1}^P \mathbb{E} \Big[ \Big\| \sum_{\tau=t_0}^{t-1} \Big( \big( \tilde{\nabla} \mathcal{L}_p^\text{local}(\bm{\theta}_p^{\tau}; \xi_p^{\tau}) - \nabla \mathcal{L}_p^\text{local}(\bm{\theta}_p^{\tau}) \big) - \frac{1}{P}\sum_{q=1}^P \big( \tilde{\nabla} \mathcal{L}_q^\text{local}(\bm{\theta}_q^{\tau}; \xi_q^{\tau}) - \nabla \mathcal{L}_q^\text{local}(\bm{\theta}_q^{\tau}) \big) \Big) \Big\|^2 \Big]}_{(A)} \\
    &\quad + 2 \eta^2 \underbrace{\frac{1}{P}\sum_{p=1}^P \mathbb{E} \Big[ \Big\| \sum_{\tau=t_0}^{t-1} \Big( \nabla \mathcal{L}_p^\text{local}(\bm{\theta}_p^{\tau}) - \frac{1}{P}\sum_{q=1}^P \nabla \mathcal{L}_q^\text{local}(\bm{\theta}_q^{\tau}) \Big) \Big\|^2 \Big]}_{(B)},
    \end{aligned}
\end{equation}
where $(a)$ follows by adding and subtracting $\tilde{\nabla} \mathcal{L}_p^\text{local}(\bm{\theta}_p^{\tau}; \xi_p^{\tau}), \nabla \mathcal{L}_p^\text{local}(\bm{\theta}_p^{\tau})$ and using $\|\mathbf{x} + \mathbf{y}\|^2 \leq \| \mathbf{x}\|^2 + \| \mathbf{y}\|^2$.

We can upper bound $(A)$ in Eq.~\ref{eq:lemma_local_global_eq_3} by
\begin{equation} \label{eq:lemma_local_global_eq_4}
    \begin{aligned}
    (A) &\underset{(a)}{\leq} \frac{1}{P}\sum_{p=1}^P \mathbb{E} \Big[ \Big\| \sum_{\tau=t_0}^{t-1} \big( \tilde{\nabla} \mathcal{L}_p^\text{local}(\bm{\theta}_p^{\tau}; \xi_p^{\tau}) - \nabla \mathcal{L}_p^\text{local}(\bm{\theta}_p^{\tau}) \big)  \Big\|^2 \Big] \\
    &\leq \frac{2}{P}\sum_{p=1}^P \sum_{\tau=t_0}^{t-1} \mathbb{E} \Big[ \Big\|  \tilde{\nabla} \mathcal{L}_p^\text{local}(\bm{\theta}_p^{\tau}; \xi_p^{\tau}) - \mathbb{E}[\tilde{\nabla} \mathcal{L}_p^\text{local}(\bm{\theta}_p^{\tau}; \xi_p^{\tau})]  \Big\|^2 \Big] \\
    &\quad + \frac{2}{P}\sum_{p=1}^P \sum_{\tau=t_0}^{t-1} \mathbb{E} \Big[ \Big\|  \mathbb{E}[\tilde{\nabla} \mathcal{L}_p^\text{local}(\bm{\theta}_p^{\tau}; \xi_p^{\tau})] - \nabla \mathcal{L}_p^\text{local}(\bm{\theta}_p^{\tau})  \Big\|^2 \Big] \\
    &\leq 2K\rho^r ( \sigma^2_\text{var} + \sigma^2_\text{bias} ),
    \end{aligned}
\end{equation}
where $(a)$ is due to 
\begin{equation} 
    \frac{1}{n}\sum_{i=1}^n \Big\| \mathbf{x}_i - \frac{1}{n}\sum_{j=1}^n \mathbf{x}_j \Big\|^2 = \frac{1}{n}\sum_{i=1}^n \| \mathbf{x}_i \|^2 - \Big\| \frac{1}{n}\sum_{i=1}^n \mathbf{x}_i \Big\|^2 \leq \frac{1}{n}\sum_{i=1}^n \| \mathbf{x}_i \|^2.
\end{equation}

We can further bound $(B)$ in Eq.~\ref{eq:lemma_local_global_eq_3} by
\begin{equation} \label{eq:lemma_local_global_eq_5}
    \begin{aligned}
    (B) &\leq \frac{1}{P}\sum_{p=1}^P (t-t_0) \sum_{\tau=t_0}^{t-1} \mathbb{E} \Big[ \Big\| \nabla \mathcal{L}_p^\text{local}(\bm{\theta}_p^{\tau}) - \frac{1}{P}\sum_{q=1}^P \nabla \mathcal{L}_q^\text{local}(\bm{\theta}_q^{\tau}) \Big\|^2 \Big] \\
    &\leq K \rho^r \sum_{\tau=t_0}^{t-1} \frac{1}{P} \sum_{p=1}^P \mathbb{E} \Big[ \Big\| \nabla \mathcal{L}_p^\text{local}(\bm{\theta}_p^{\tau}) - \frac{1}{P}\sum_{q=1}^P \nabla \mathcal{L}_q^\text{local}(\bm{\theta}_q^{\tau}) \Big\|^2 \Big] \\
    &\underset{(a)}{\leq} K\rho^r \sum_{\tau=t_0}^{t-1} \Big( \frac{8L^2}{P}\sum_{p=1}^P \| \bm{\theta}_p^\tau - \bar{\bm{\theta}}^\tau \|^2 + 8 \kappa^2  \Big) \\
    &\leq \frac{8 L^2K \rho^r}{P} \sum_{\tau=t_0}^{t-1} \sum_{p=1}^P \| \bm{\theta}_p^\tau - \bar{\bm{\theta}}^\tau \|^2 + 8 K^2 \rho^{2r} \kappa^2,
    \end{aligned}
\end{equation}
where $(a)$ is due to Lemma~\ref{lemma:local_diff} and the definition of $\kappa$.

By plugging Eq.~\ref{eq:lemma_local_global_eq_4} and Eq.~\ref{eq:lemma_local_global_eq_5} into Eq.~\ref{eq:lemma_local_global_eq_3}, we have
\begin{equation}\label{eq:lemma_local_global_eq_6}
    \frac{1}{P}\sum_{p=1}^P \mathbb{E}[\| \bar{\bm{\theta}}^{t} - \bm{\theta}^{t}_p \|^2] 
    \leq 4\eta^2 K \rho^r (\sigma^2_\text{var} + \sigma_\text{bias}^2) +  16 \eta^2 L^2 K\rho^r \sum_{\tau=t_0}^{t-1} \frac{1}{P} \sum_{p=1}^P \| \bm{\theta}_p^\tau - \bar{\bm{\theta}}^\tau \|^2 + 16 \eta^2 K^2 \rho^{2r} \kappa^2 .
\end{equation}

Let us define $\mathcal{T}_\text{local} = \mathcal{T}_{local}(1) \cup \ldots \cup \mathcal{T}_{local}(R)$.
By summing over $t \in \mathcal{T}_\text{local}$, we have
\begin{equation}
    \begin{aligned}
    & \underbrace{\sum_{r=1}^{R} \sum_{t\in \mathcal{T}_\text{local}(r)}\frac{1}{P}\sum_{p=1}^P \mathbb{E}[\| \bar{\bm{\theta}}^{t} - \bm{\theta}^{t}_p \|^2]}_{(A)} \\
    &\leq \sum_{r=1}^{R} \left( 4\eta^2 (\sigma^2_\text{bias} + \sigma^2_\text{var}) K^2 \rho^{2r}  + 16 \eta^2 \kappa^2  K^2 \rho^{2r} \right) \\
    &\quad + \sum_{r=1}^{R} 16 \eta^2 L^2 K^2 \rho^{2r} \underbrace{\sum_{t\in\mathcal{T}_\text{local}(r)} \frac{1}{P} \sum_{p=1}^P \| \bm{\theta}_p^t - \bar{\bm{\theta}}^t \|^2}_{(A)} .
    \end{aligned}
\end{equation}

Rearranging the terms gives 
\begin{equation}
    \sum_{r=1}^{R} (1-16 \eta^2 L^2 K^2 \rho^{2r}) \underbrace{\sum_{t\in \mathcal{T}_\text{local}(r)}\frac{1}{P}\sum_{p=1}^P \mathbb{E}[\| \bar{\bm{\theta}}^{t} - \bm{\theta}^{t}_p \|^2] }_{(A)}
    \leq \sum_{r=1}^{R} \left( 4\eta^2 ((\sigma^2_\text{bias} + \sigma^2_\text{var})) K^2 \rho^{2r}  + 16 \eta^2 \kappa^2  K^2 \rho^{2r} \right).
\end{equation}

Therefore, we conclude that
\begin{equation}
    \sum_{t\in \mathcal{T}_\text{local}(r)}\frac{1}{P}\sum_{p=1}^P \mathbb{E}[\| \bar{\bm{\theta}}^{t} - \bm{\theta}^{t}_p \|^2] \leq \frac{4\eta^2 ((\sigma^2_\text{bias} + \sigma^2_\text{var})) K^2 \rho^{2r}  + 16 \eta^2 \kappa^2  K^2 \rho^{2r}}{1-16 \eta^2 L^2 K^2 \rho^{2r}}.
\end{equation}
which completes the proof. 

\end{document}